\documentclass[10pt,twocolumn,letterpaper]{article}

\usepackage{cvpr}              
\usepackage{graphicx}
\usepackage{booktabs}
\usepackage{tabularx}
\usepackage{adjustbox,diagbox,threeparttable,tabularray}
\usepackage{enumitem}
\usepackage{capt-of}
\usepackage{xspace}
\usepackage{xcolor}
\usepackage{minitoc}
\usepackage{amsmath,amsthm,amssymb,amsfonts,dsfont,pifont,bm,bbm,mathrsfs,mathtools,nicefrac,extarrows,relsize}
\usepackage{algorithm,algpseudocode,listings}
\usepackage[accsupp]{axessibility}
\usepackage{booktabs,multirow,adjustbox,diagbox,threeparttable,tabularray,setspace}

\definecolor{cvprblue}{rgb}{0.21,0.49,0.74}
\usepackage[pagebackref,breaklinks,colorlinks,citecolor=cvprblue]{hyperref}
\newcommand{\floor}[1]{\lfloor #1 \rfloor}

\newcommand{\drawguess}{\textit{Draw \& Guess~}}
\newcommand{\methodname}{MagicQuill}
\newcommand{\method}{\texttt{\methodname}\xspace}

\def\paperID{}
\def\confName{CVPR}
\def\confYear{2025}

\title{\methodname: An Intelligent Interactive Image Editing System}

\vspace{-0.2cm}
\author{Zichen Liu$^{\heartsuit,1,2}$, Yue Yu$^{\heartsuit,1,2}$, Hao Ouyang$^{2}$, Qiuyu Wang$^{2}$, \\  
Ka Leong Cheng$^{1,2}$, Wen Wang$^{3,2}$, Zhiheng Liu$^{4}$, Qifeng Chen$^{\dagger,1}$, Yujun Shen$^{\dagger,2}$ \\[0.2cm]
$^1$HKUST, $^2$Ant Group, $^3$ZJU, $^4$HKU
}

\begin{document}

\twocolumn[{
\renewcommand\twocolumn[1][]{#1}
\maketitle
\begin{center}
    \vspace{-10pt}
    \includegraphics[width=0.9\linewidth]{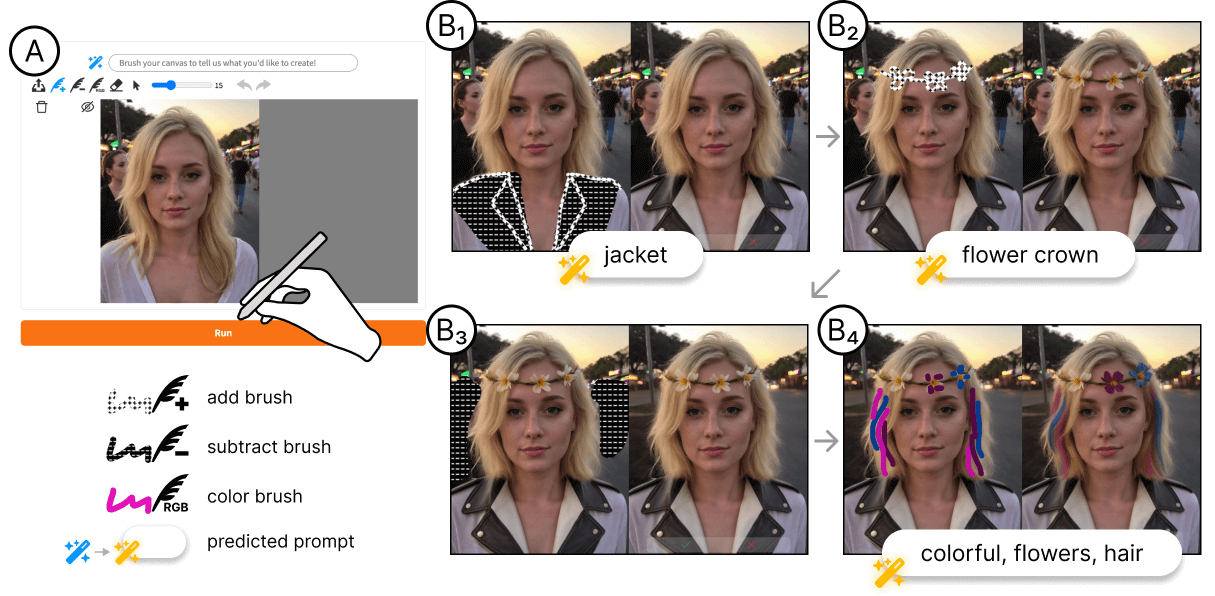}
    \vspace{-10pt}
    \captionsetup{type=figure}
    \caption{\method is an intelligent and interactive image editing system built upon diffusion models. Users seamlessly edit images using three intuitive brushstrokes: add, subtract, and color (A). A MLLM dynamically predicts user intentions from their brush strokes and suggests contextual prompts (B1-B4). The examples demonstrate diverse editing operations: to generate a jacket from clothing contour (B1), add a flower crown from head sketches (B2), remove background (B3), and apply color changes to the hair and flowers (B4).}
    \label{fig:teaser}
\end{center}
}]

\maketitle
\let\thefootnote\relax\footnotetext{\noindent$^\heartsuit$Equal contribution. $^\dagger$Corresponding author.}

\begin{abstract}
\vspace{-0.2cm}
As a highly practical application, image editing encounters a variety of user demands and thus prioritizes excellent ease of use.
In this paper, we unveil \method, an integrated image editing system designed to support users in swiftly actualizing their creativity.
Our system starts with a streamlined yet functionally robust interface, enabling users to articulate their ideas (\textit{e.g.}, inserting elements, erasing objects, altering color, \textit{etc.}) with just a few strokes.
These interactions are then monitored by a multimodal large language model (MLLM) to anticipate user intentions in real time, bypassing the need for prompt entry.
Finally, we apply the powerful diffusion prior, enhanced by a carefully learned two-branch plug-in module, to process the editing request with precise control.
%
%
Please visit the \href{https://magic-quill.github.io}{project page} to try out our system.

\end{abstract}
    
\section{Introduction}
\label{sec:intro}
Performing precise and efficient edits on digital photographs remains a significant challenge, especially when aiming for nuanced modifications. As shown in Fig.~\ref{fig:teaser}, consider the task of editing a portrait of a lady where specific alterations are desired: converting a shirt to a custom-designed jacket, adding a flower crown at an exact position with a well-designed shape, dyeing portions of her hair in particular colors, and removing certain parts of the background to refine her appearance. Despite the rapid advancements in diffusion models~\cite{Survey_DiffusionImageEditing,emuedit,instructdiffusion,instructpix2pix,phd,self-guidance,imagebrush,pixel-diffusion,sde-drag,draggan,dragondiffusion} and recent attempts to enhance control~\cite{smartedit, smartmask,powerpaint,brushnet}, achieving such fine-grained and precise edits continues to pose difficulties, typically due to a lack of intuitive interfaces and models for fine-grained control.

The challenges highlight the critical need for interactive editing systems that facilitate precise and efficient modifications. An ideal solution would empower users to specify \textit{\textbf{what}} they want to edit, \textit{\textbf{where}} to apply the changes, and \textit{\textbf{how}} the modifications should appear, all within a \textit{\textbf{user-friendly}} interface that streamlines the editing process.

We aim to develop the first robust, \textit{\textbf{open-source}}, interactive precise image editing system to make image editing easy and efficient. Our system seamlessly integrates three core modules: the \textbf{Editing Processor}, the \textbf{Painting Assistor}, and the \textbf{Idea Collector}. The Editing Processor ensures a high-quality, controllable generation of edits, accurately reflecting users' editing intentions in color and edge adjustments. The Painting Assistor enhances the ability of the system to predict and interpret the users' editing intent. The Idea Collector serves as an intuitive interface, allowing users to input their ideas quickly and effortlessly, significantly boosting the editing efficiency.

The Editing Processor implements two kinds of brushstroke-based guidance mechanisms: scribble guidance for structural modifications (e.g., adding, detailing, or removing elements) and color guidance for modification of color attributes. Inspired by ControlNet~\cite{controlnet} and BrushNet~\cite{brushnet}, our control architecture ensures precise adherence to user guidance while preserving unmodified regions.
Our Painting Assistor reduces the repetitive process of typing text prompts, which disrupts the editing workflow and creates a cumbersome transition between prompt input and image manipulation. It employs an MLLM to interpret brushstrokes and automatically predicts prompts based on image context. We call this novel task Draw\&Guess. We construct a dataset simulating real editing scenarios for fine-tuning to ensure the effectiveness of the MLLM in understanding user intentions. This enables a continuous editing workflow, allowing users to iteratively edit images without manual prompt input. 
The Idea Collector provides an intuitive interface compatible with various platforms including Gradio and ComfyUI, allowing users to draw with different brushes, manipulate strokes, and perform continuous editing with ease.

We present a comprehensive evaluation of our interactive editing framework. Through qualitative and quantitative analyses, we demonstrate that our system significantly improves both the precision and efficiency of performing detailed image edits compared to existing methods. 
Our Editing Processor achieves superior edge alignment and color fidelity compared to baselines like SmartEdit~\cite{smartedit} and BrushNet~\cite{brushnet}. 
The Painting Assistor exhibits superior user intent interpretation capabilities compared to state-of-the-art MLLMs, including LLaVA-1.5~\cite{llava}, LLaVA-Next~\cite{llava-next}, and GPT-4o~\cite{gpt-4o}. 
User studies indicate that the Idea Collector significantly outperforms baseline interfaces in all aspects of system usability.

By leveraging advanced generative models and a user-centric design, our interactive editing framework significantly reduces the time and expertise required to perform detailed image edits. By addressing the limitations of current image editing tools and providing an innovative solution that enhances both precision and efficiency, our work advances the field of digital image manipulation. Our framework opens possibilities for users to engage creatively with image editing, achieving their goals easily and effectively.


\section{Related Works}
\subsection{Image Editing}
Image editing involves modifying the visual appearance, structure, or elements of an existing image~\cite{Survey_DiffusionImageEditing}. Recent breakthroughs in diffusion models~\cite{DDPM, DDIM, LDM} have significantly advanced visual generation tasks, outperforming GAN-based models~\cite{GAN} in terms of image editing capabilities. 
To enable control and guidance in image editing, a variety of approaches have emerged, leveraging different modalities such as textual instructions~\cite{emuedit, instructdiffusion, instructpix2pix, magicbrush, dit4edit, omnigen, seedx, ledits++, infedit, pnp, imagic, agap, edicho}, masks~\cite{smartedit, smartmask, powerpaint, brushnet, inpainting_OU}, layouts~\cite{phd, self-guidance,liu2023cones2}, segmentation maps~\cite{imagebrush, pixel-diffusion}, strokes~\cite{unipaint, sdedit}, references~\cite{ cones, manganinja, imprint, anydoor}, and point-dragging interfaces~\cite{sde-drag, draggan, dragondiffusion}. Despite these advances, these methods often fall short when precise modifications at the regional level are required, such as alterations to object shape, color, and other details.
Among the various methods, sketch-based editing approaches~\cite{sketchedit, sketchffusion, plasticsurgery, customsketching, sc-fegan, reference_sketch, faceshop} offer users a more intuitive and precise means of interaction. 
However, the current methods remain limited by the accuracy of the text signals input alongside the sketches, making it challenging to precisely control the information of the editing areas, such as color.
To achieve precise control, we introduce two types of local guidance based on brushstrokes: scribble and color, thereby enabling fine-grained control over shape and color at the regional level.

\begin{figure*}[t]
    \centering
    \begin{subfigure}[t]{\linewidth}
        \centering
        \includegraphics[width=\linewidth]{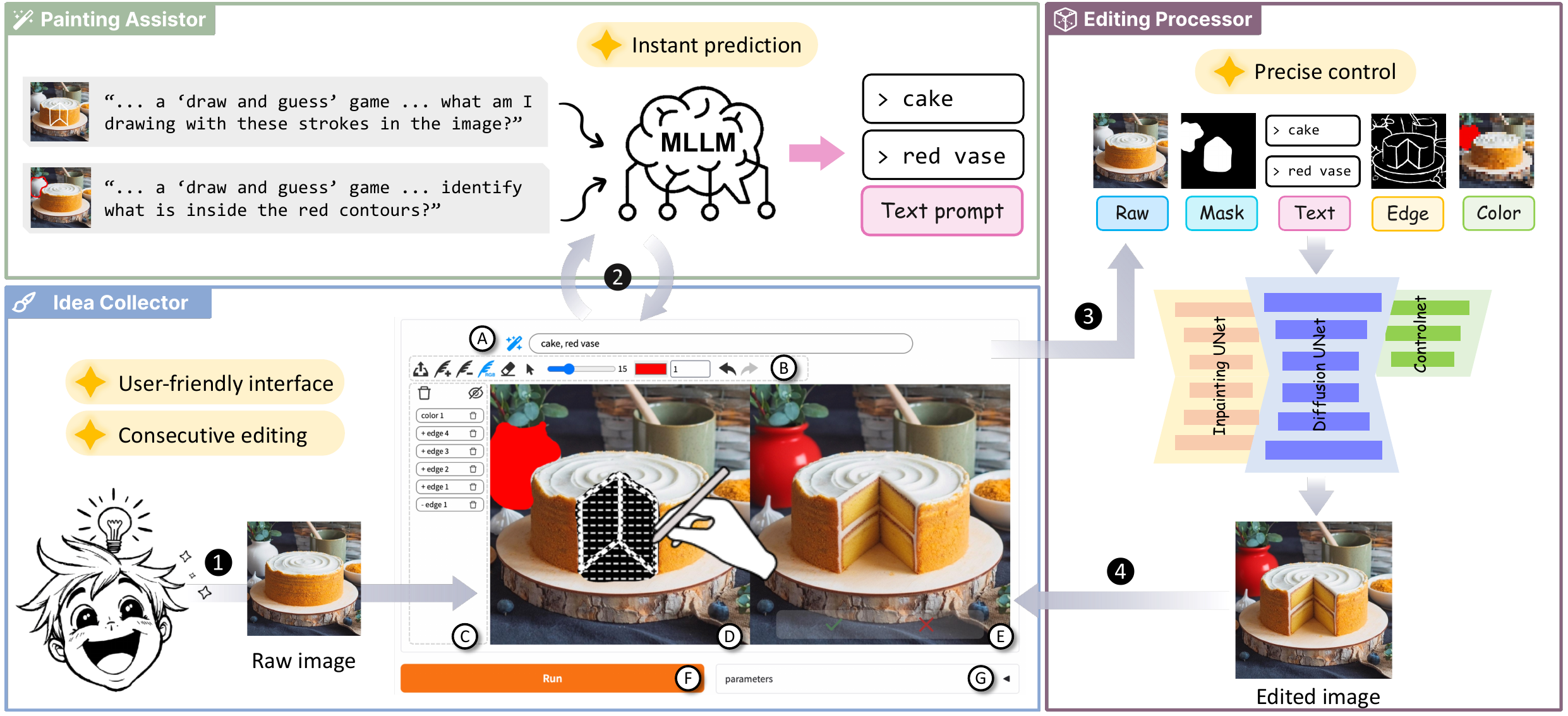}
    \end{subfigure}
    \vspace{-13pt}
    \caption{System framework consisting of three integrated components: an \textbf{Editing Processor} with dual-branch architecture for controllable image inpainting, a \textbf{Painting Assistor} for real-time intent prediction, and an \textbf{Idea Collector} offering versatile brush tools. This design enables intuitive and precise image editing through brushstroke-based interactions.}
    \vspace{-0.2cm}
    \label{fig:design}
\end{figure*}

\subsection{MLLMs for Image Editing}
Multi-modal large language models (MLLMs) extend LLMs to process both text and image content~\cite{survey_mllm_generation}, enabling text-to-image generation~\cite{emu, emu2, gill, dreamllm}, prompt-refinement~\cite{RPG, idea2img}, and image quality evaluation~\cite{dreamsync}.

In the area of image editing, MLLMs have demonstrated significant potential. MGIE~\cite{MGIE} enhances instruction-based image editing by using MLLMs to generate more expressive, detailed instructions. SmartEdit~\cite{smartedit} leverages MLLM for better understanding and reasoning towards complex instruction. FlexEdit~\cite{flexedit} integrates MLLM to understand image content, masks, and textual instructions. GenArtist~\cite{genartist} uses an MLLM agent to decompose complex tasks, guide tool selection, and enable systematic image generation, editing, and self-correction with step-by-step verification.
Our system extends this line of research by introducing a more intuitive approach, utilizing MLLM to simplify the editing process. 
Specifically, it directly integrates the image context with user-input strokes to infer and translate the editing intentions, thereby automatically generating the necessary prompts without requiring repeated user input. 
This innovative task, which we term Draw\&Guess, facilitates a continuous editing workflow, enabling users to iteratively refine images with minimal manual intervention.

\subsection{Interactive Support for Image Generation}
Interactive support enhances the performance and usability of generative models through human-in-the-loop collaboration~\cite{Ko2023LargeScale}. Recent works have focused on making prompt engineering more user-friendly through techniques like image clustering~\cite{promptify, PromptMagician} and attention visualization~\cite{promptcharm}.

Despite advancements in interactive support, a key challenge remains in bridging the gap between verbal prompts and visual output. While systems like PromptCharm~\cite{promptcharm} and DesignPrompt~\cite{DesignPrompt} use inpainting for interactive image editing, these tools typically offer only coarse-grained control over element addition and removal, requiring users to brush over areas before generating objects within those regions. Furthermore, users must manually input prompts to specify the objects they wish to generate.
Our approach addresses these limitations by introducing fine-grained image editing through the use of brushstrokes. 
Additionally, we incorporate a multimodal large language model (MLLM) that provides on-the-fly assistance by interpreting user intentions and suggesting prompts in real-time, thereby reducing cognitive load and enhancing overall usability.

\section{System Design}
Our system is structured around three key aspects: \textbf{Editing Processor} with strong generative prior, \textbf{Painting Assistor} with instant intent prediction, and \textbf{Idea Collector} with a user-friendly interface. An overview of our system design is presented in Fig.~\ref{fig:design}. 

\begin{figure}[t]
    \centering
    \includegraphics[width=\linewidth]{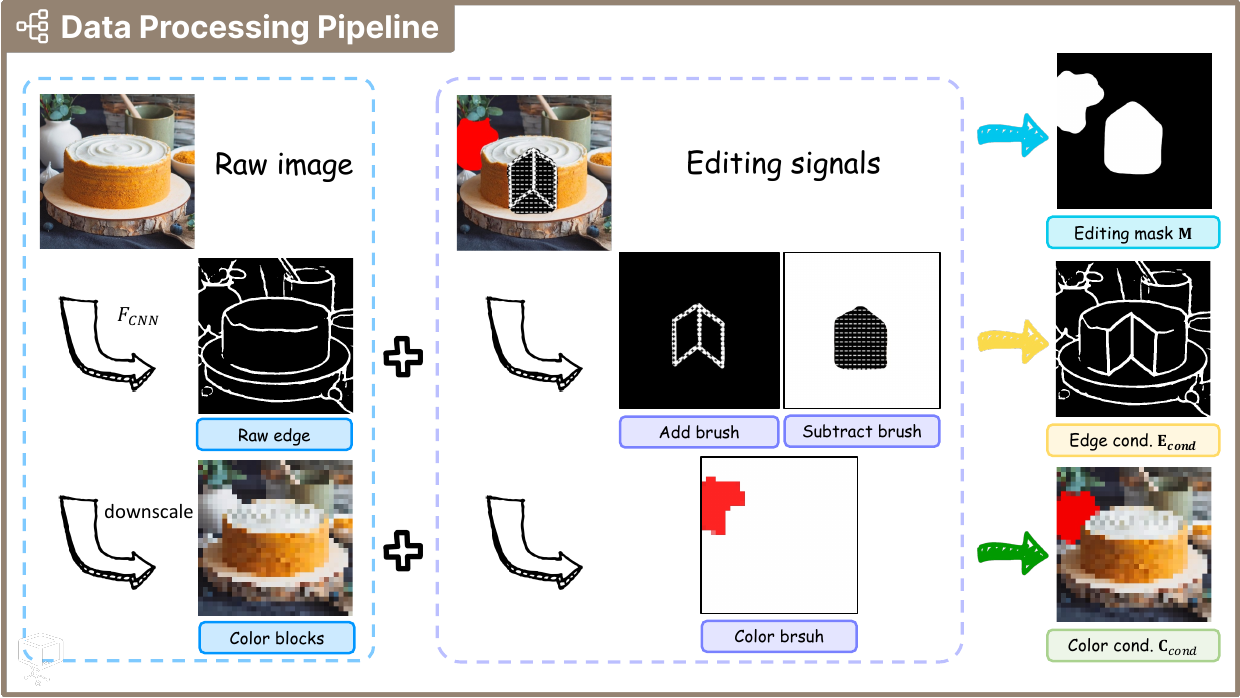}
    \caption{Data processing pipeline. The input image undergoes edge extraction via CNN and color simplification through downscaling. Three editing conditions are then generated based on brush signals: editing mask, edge condition, and color condition, which together provide control for image editing.}
    \label{fig:data}
    \vspace{-0.4cm}
\end{figure}

\begin{figure}[t]
    \centering
    \includegraphics[width=\linewidth]{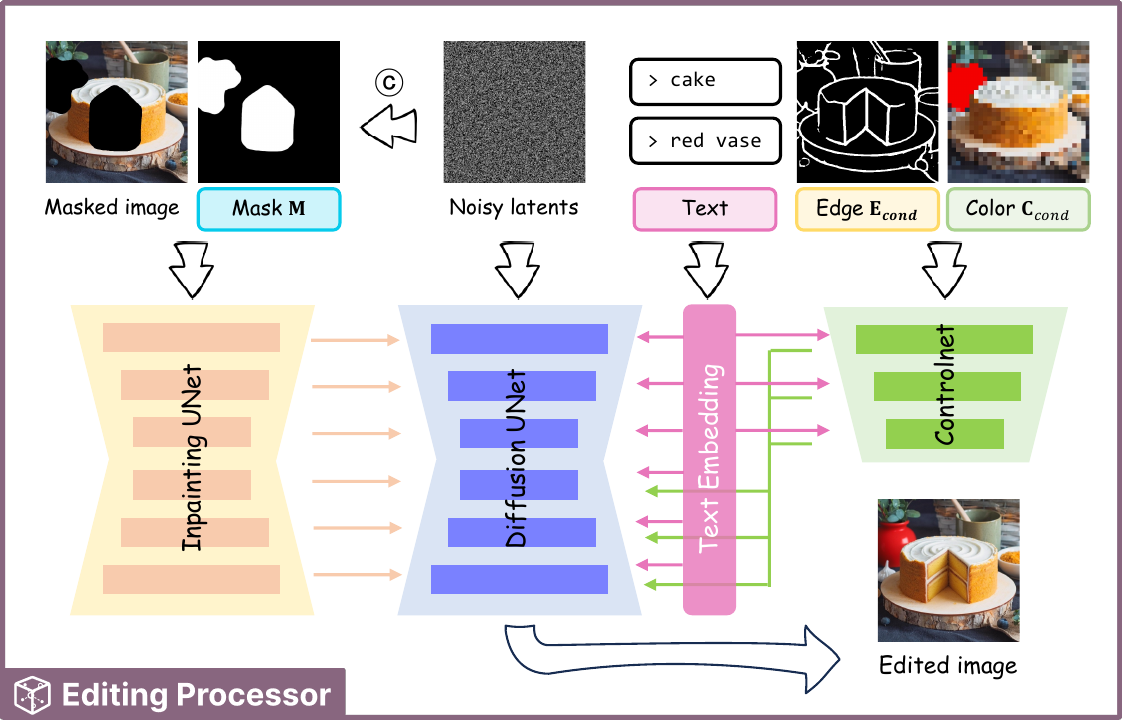}
    \caption{Overview of our Editing Processor. The proposed architecture extends the latent diffusion UNet with two specialized branches: an inpainting branch for content-aware per-pixel inpainting guidance and a control branch for structural guidance, enabling precise brush-based image editing.}
    \vspace{-0.4cm}
    \label{fig:arch}
\end{figure}

Our system introduces brushstroke-based control signals to give intuitive and precise control. These signals allow users to express their editing intentions by simply drawing what they envision. We designed two types of brushes, scribble and color, to accurately manipulate the edited image. 
The scribble brushes, \textbf{add brush} and \textbf{subtract brush}, aim to provide precise structural control by operating on the edge map of the original image.  
The \textbf{color brush} works with downsampled color blocks to enable fine-grained color manipulation of specific regions.
Fig.~\ref{fig:data} illustrates the workflow to convert the user hand-drawn input signal into control condition for faithfully inpainting the target editing area. 
Inspired by \citet{brushnet, controlnet}, we employ two additional branches to the latent diffusion framework~\cite{LDM}, with the inpainting branch giving content-aware per-pixel guidance for the re-generation of the editing area, and the control branch providing structural guidance. 
The model architecture is illustrated in Fig.~\ref{fig:arch}. Further details will be discussed in Sec.~\ref{Editing_Processor}.

To reduce the cognitive load for users to input appropriate prompts at every stage of editing, our system integrates a MLLM~\cite{improved_instruction} as the Painting Assistor. This component analyzes user brushstrokes to deduce the editing intention based on the image context, thereby automatically suggesting contextually relevant prompts for editing. 
We have named this innovative task Draw\&Guess. 
To effectively prepare the MLLM for Draw\&Guess, we designed a dataset construction method to simulate user hand-drawn editing scenarios and acquire ground truth for Draw\&Guess.
We fine-tuned a dedicated LLaVA~\cite{llava} model, achieving instant prompt guessing with satisfactory accuracy.
More specifics will be covered in Sec.~\ref{draw_guess}.

Additionally, to provide users with a streamlined, intuitive interface that empowers them to express their ideas for complex image editing tasks with ease, we designed an Idea Collector with a user-friendly interface. The key features of the interface will be outlined in Sec.~\ref{idea_collector}.

\subsection{Editing Processor}
\noindent\textbf{Control Condition from Brushstroke Signal:}
\label{Editing_Processor}
Let $\mathbf{M}_{add}$ and $\mathbf{M}_{sub}$ denote the binary masks corresponding to add and subtract brush respectively. These masks share the same dimensions as the original image $\mathbf{I}$, where values are set to $1$ in regions corresponding to user brush strokes and $0$ elsewhere.
The subtract brush masks out the edges from the edge map $\mathbf{E}$, which is initially extracted from the original image using a pre-trained CNN $f_{CNN}$. Conversely, the add brush introduces new edges by setting designated regions to white in the edge map.
The resulting modified edge map $\mathbf{E}_{cond}$ serves as the control condition for manipulating geometric structure in the editing processor. This can be formally expressed as
{\small
\begin{equation}
\begin{split}
&\mathbf{E} = f_{CNN}(\mathbf{I}), \\ 
&\mathbf{E}_{sub} = \mathbf{E} \odot (1 - \mathbf{M}_{sub}), \\
&\mathbf{E}_{cond} = \mathbf{E}_{sub} + \mathbf{M}_{add} \odot (1 - \mathbf{E}_{sub}).
\end{split}
\end{equation}
}

For precise region-specific colorization, we represent each color brush stroke as a tuple $(\mathbf{M}_{color}, \mathbf{c}, \alpha)$, where $\mathbf{M}_{color}$ denotes a binary mask indicating the user-defined stroke region, $\mathbf{c}$ specifies the stroke color, and $\alpha \in [0,1]$ represents the stroke opacity. The colorization operation can be formally expressed as
{\small
\begin{equation}
\mathbf{I}_{c} = (1 - \alpha\cdot\mathbf{M}_{color}) \odot \mathbf{I} + \alpha\cdot \mathbf{M}_{color} \cdot \mathbf{c},
\end{equation}
}

\noindent where the color \(\mathbf{c}\) with an alpha blending factor \(\alpha\) is applied over a specific region of the image \(\mathbf{I}\) defined by the binary mask \(\mathbf{M}_{color}\).
 
To generate the color condition $\mathbf{C}_{cond}$, we first downscale the image $\mathbf{I}_{c}$ by a factor of 16 using cubic interpolation, followed by upscaling to the original resolution using nearest-neighbor interpolation. This process generated a color block preserving the global color structure while simplifying local details.

The edge condition $\mathbf{E}_{cond}$ and color condition $\mathbf{C}_{cond}$ jointly guide the inpainting process for precise editing control. The editing region, represented by mask $\mathbf{M}$, is obtained by dilating the union of brush regions by $p$ pixels. The masked image $\mathbf{I}_{masked}$ can then be formulated as
{\small
\begin{equation}
\begin{split}
&\mathbf{M}=Grow_p(\mathbf{M}_{add} \cup \mathbf{M}_{sub} \cup \mathbf{M}_{color}), \\
&\mathbf{I}_{masked} = \mathbf{I} \odot (1 - \mathbf{M}).
\end{split}
\end{equation}
}

This expansion accounts for the fact that editing can affect areas surrounding the mask, such as shadows or other adjacent details. By growing the mask, we ensure that these peripheral regions are properly generated, resulting in a more seamless and realistic edit.

\noindent\textbf{Controllable Image Inpainting:} The inpainting branch adopts the UNet~\cite{brushnet,UNet} architecture, incorporating the masked image feature into the pre-trained diffusion network. 
This branch inputs the concatenated noisy latent at $t$-th step $z_t$, masked image latent $z_{masked}$ extracted using VAE~\cite{vae} from $\mathbf{I}_{masked}$, and downsampled mask $\mathbf{m}$ by cubic interpolation from $\mathbf{M}$. 
The inpainting branch processes these features, utilizing a trainable clone of the diffusion model, stripped of cross-attention layers to focus solely on the image feature. 
The extracted features carrying pixel-level information are inserted into each layer of the frozen diffusion model through zero-convolution layers $\mathcal{Z}$~\cite{controlnet}. 
Given text condition $\tau$, timestep $t$, let $F(z_t, \tau, t; \Theta)_i$ represents the feature of the $i$-th layer in the total $n$ layers of the diffusion UNet with parameter $\Theta$. Similarly, let $F^I([z_t, z_{masked}, \mathbf{m}], t; \Theta^I)_i$ represents the output of the $i$-th layer in the inpainting UNet, where $[\cdot]$ denotes the concatenation operation. This feature insertion can be represented by
{\small
\begin{equation}
\begin{split}
F(z_t, \tau, t; \Theta)_i \, +\!= \, w_I \cdot \mathcal{Z} (F^I([z_t, z_{masked}, \mathbf{m}], t; \Theta^I)_i), \\ 
\end{split}
\end{equation}}

\noindent where $w_I$ is an adjustable hyperparameter that determines the inpainting strength. Equipped with the inpainting branch, the diffusion UNet can fill the masked area in a content-aware manner based on the text prompt. 

The control branch aims to introduce conditional generation ability to the diffusion UNet based on condition $\mathcal{C}=\{\mathbf{E}_{cond}, \mathbf{C}_{cond}\}$. 
We adopt ControlNet~\cite{controlnet} to insert conditional control into the middle and decoder blocks of the diffusion UNet.
Let $F^C(z_t, \mathcal{C}, t; \Theta^C)_i$ represent the output of the $i$-th layer in the ControlNet, the control feature insertion can be formulated as
{\small
\begin{equation}
\begin{split}
F(z_t, \tau, t; \Theta)_{\floor{\frac{n}{2}}+i} \, +\!= \, w_C \cdot \mathcal{Z}(F^C(z_t, \mathcal{C}, t; \Theta^C)_i), \\ 
\end{split}
\end{equation}}

\noindent where $w_C$ is an adjustable hyperparameter that determines the control strength. Both the inpainting and control branches don't alter the weights of the pre-trained diffusion models, enabling it to be a plug-and-play component applicable to any community fine-tuned diffusion models.
The control branch is trained using the denoising score matching objective, which can be written as
\begin{equation}
\mathcal{L}=\mathbb{E}_{z_t, t, \epsilon \sim \mathcal{N}(0, \mathbf{I})}\left[\left\|\epsilon-\epsilon^{c}\left(z_t, \mathcal{C}, t; \{ \Theta , \Theta^C \} \right)\right\|^2\right],
\label{eq:contorlnet}
\end{equation}
where $\epsilon^{c}$ is the combination of the denoising U-Net and the ControlNet model.


\subsection{Painting Assistor}
\label{draw_guess}
\begin{figure*}[t]
    \vspace{-0.3cm}
    \begin{minipage}[b]{0.195\linewidth}
        \begin{subfigure}[b]{\linewidth}
            \centering
            \includegraphics[width=\linewidth]{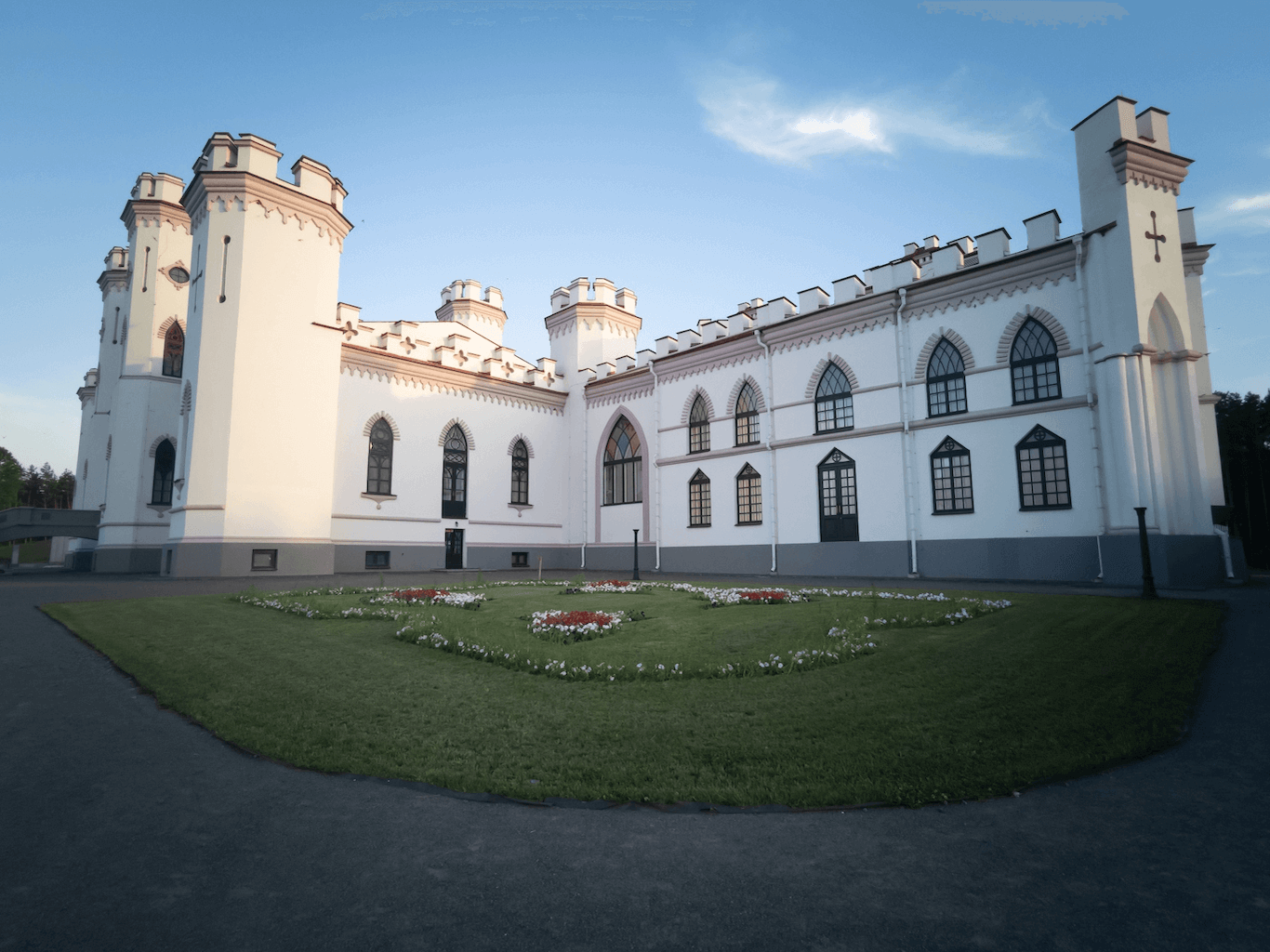}
        \end{subfigure}
    \end{minipage} 
    \begin{minipage}[b]{0.195\linewidth}
        \begin{subfigure}[b]{\linewidth}
            \centering
            \includegraphics[width=\linewidth]{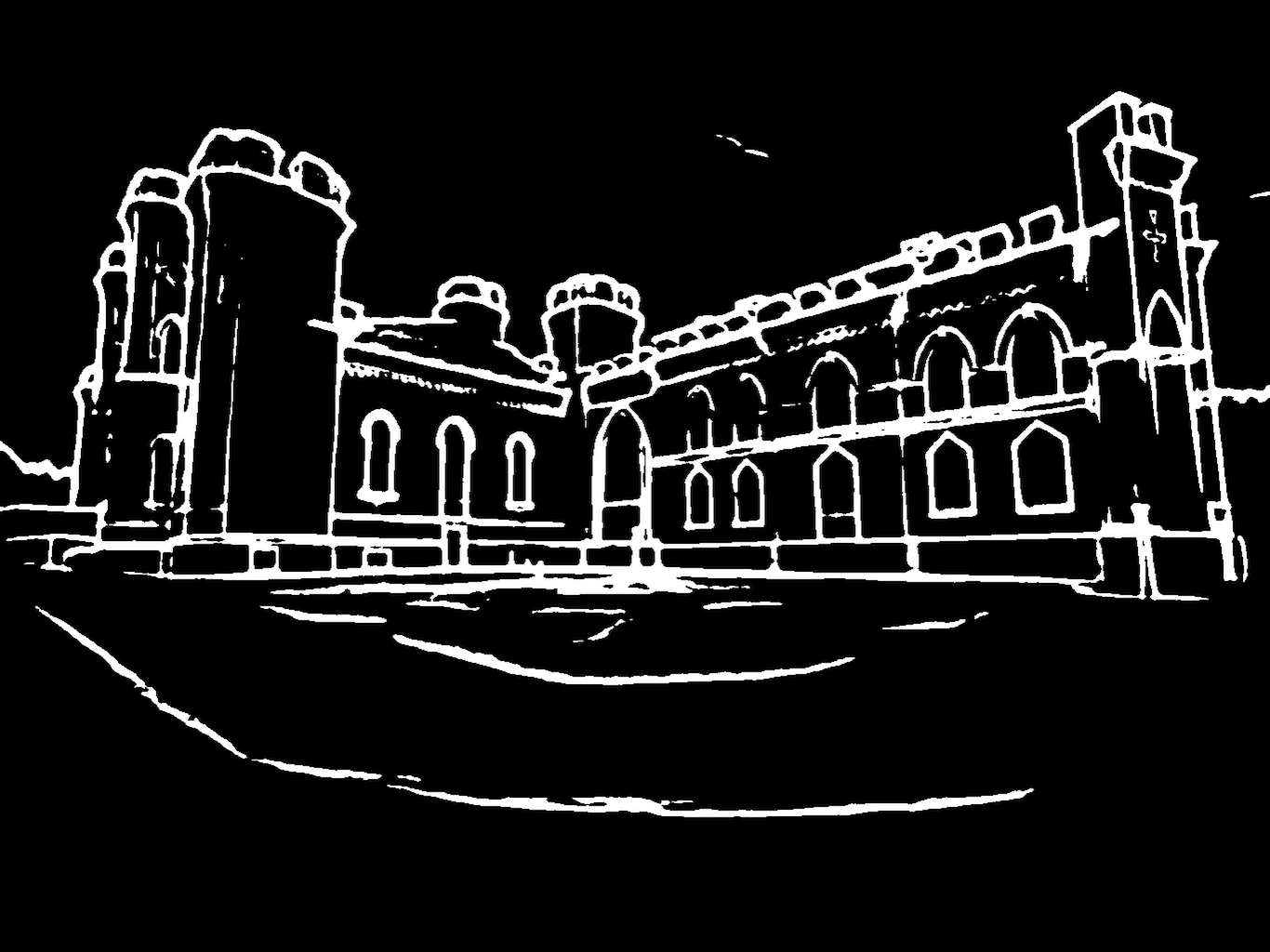}
        \end{subfigure}
    \end{minipage} 
    \begin{minipage}[b]{0.195\linewidth}
        \begin{subfigure}[b]{\linewidth}
            \centering
            \includegraphics[width=\linewidth]{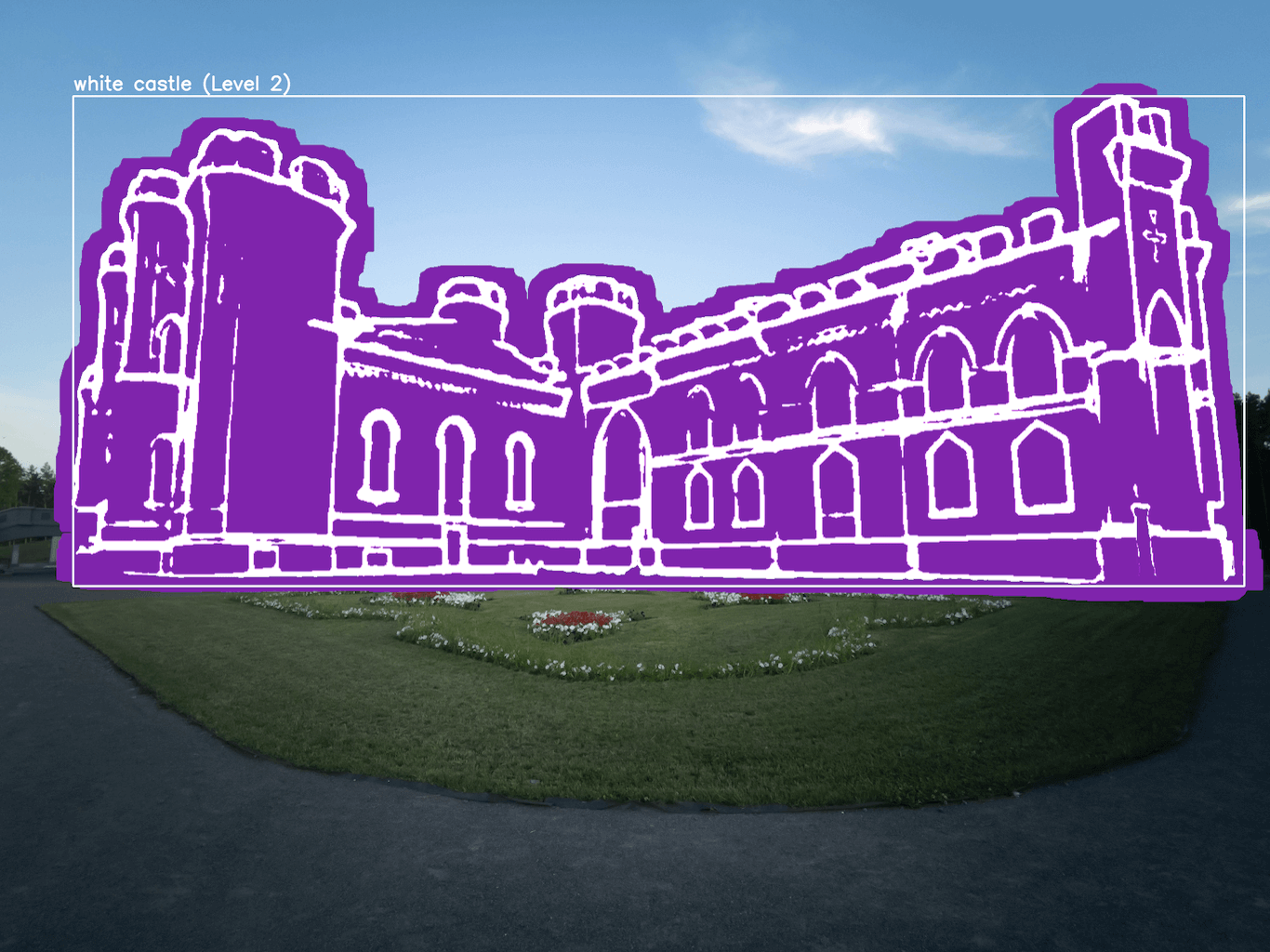}
        \end{subfigure}
    \end{minipage} 
    \begin{minipage}[b]{0.195\linewidth}
        \begin{subfigure}[b]{\linewidth}
            \centering
            \includegraphics[width=\linewidth]{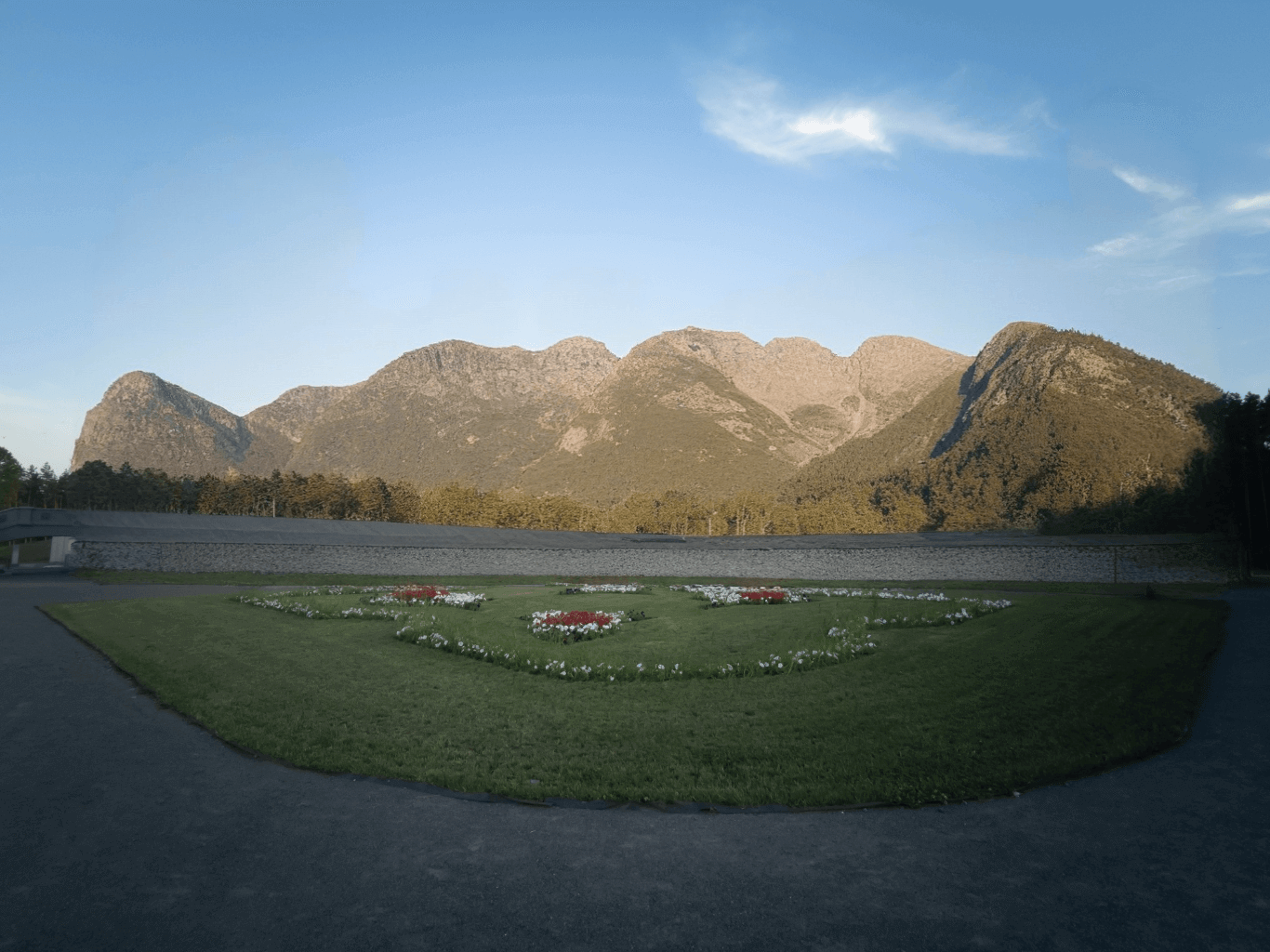}
        \end{subfigure}
    \end{minipage} 
    \begin{minipage}[b]{0.195\linewidth}
        \begin{subfigure}[b]{\linewidth}
            \centering
            \includegraphics[width=\linewidth]{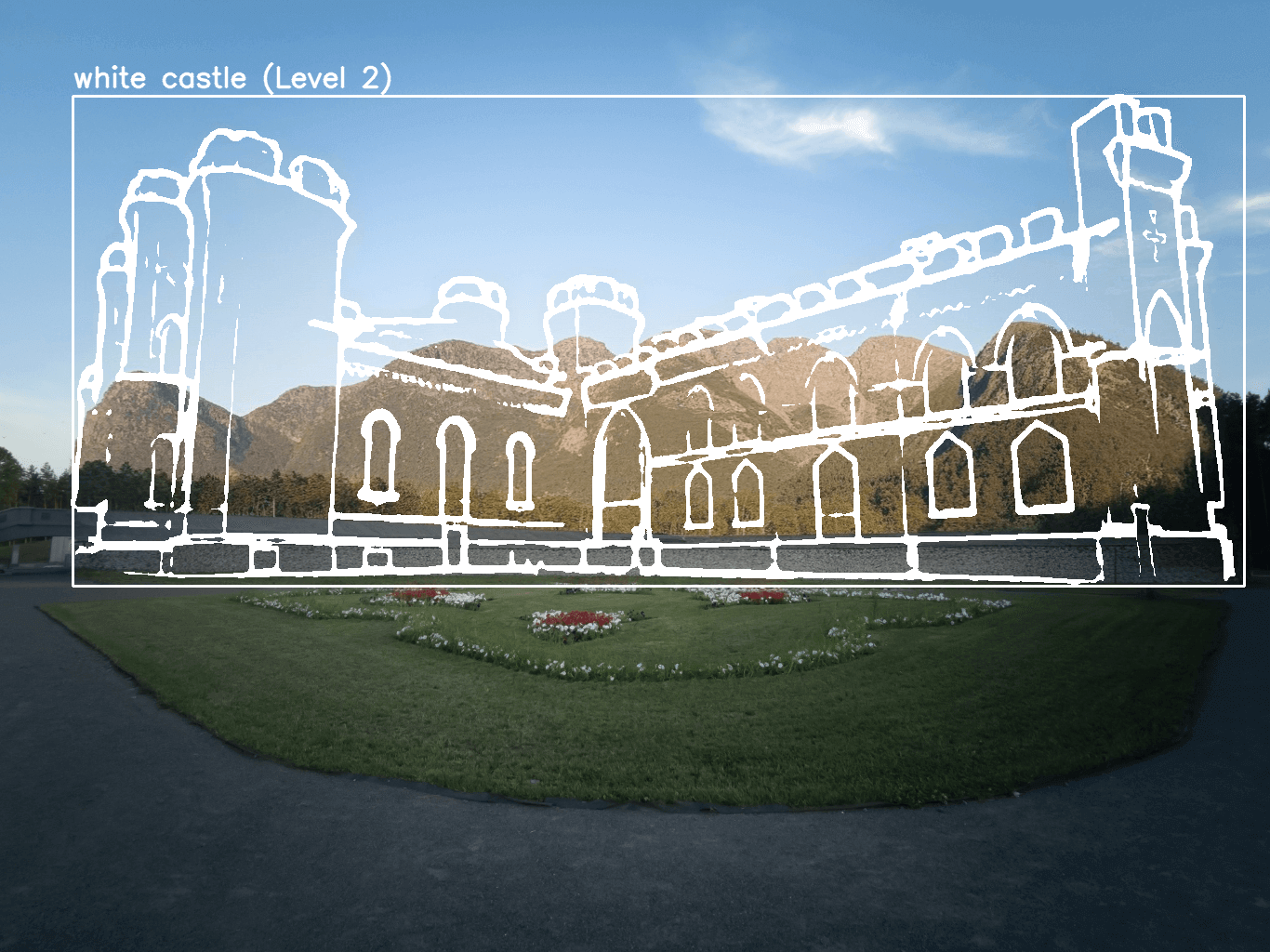}
        \end{subfigure}
    \end{minipage}

    
    \begin{minipage}[b]{0.195\linewidth}
        \begin{subfigure}[b]{\linewidth}
            \centering
            \includegraphics[width=\linewidth]{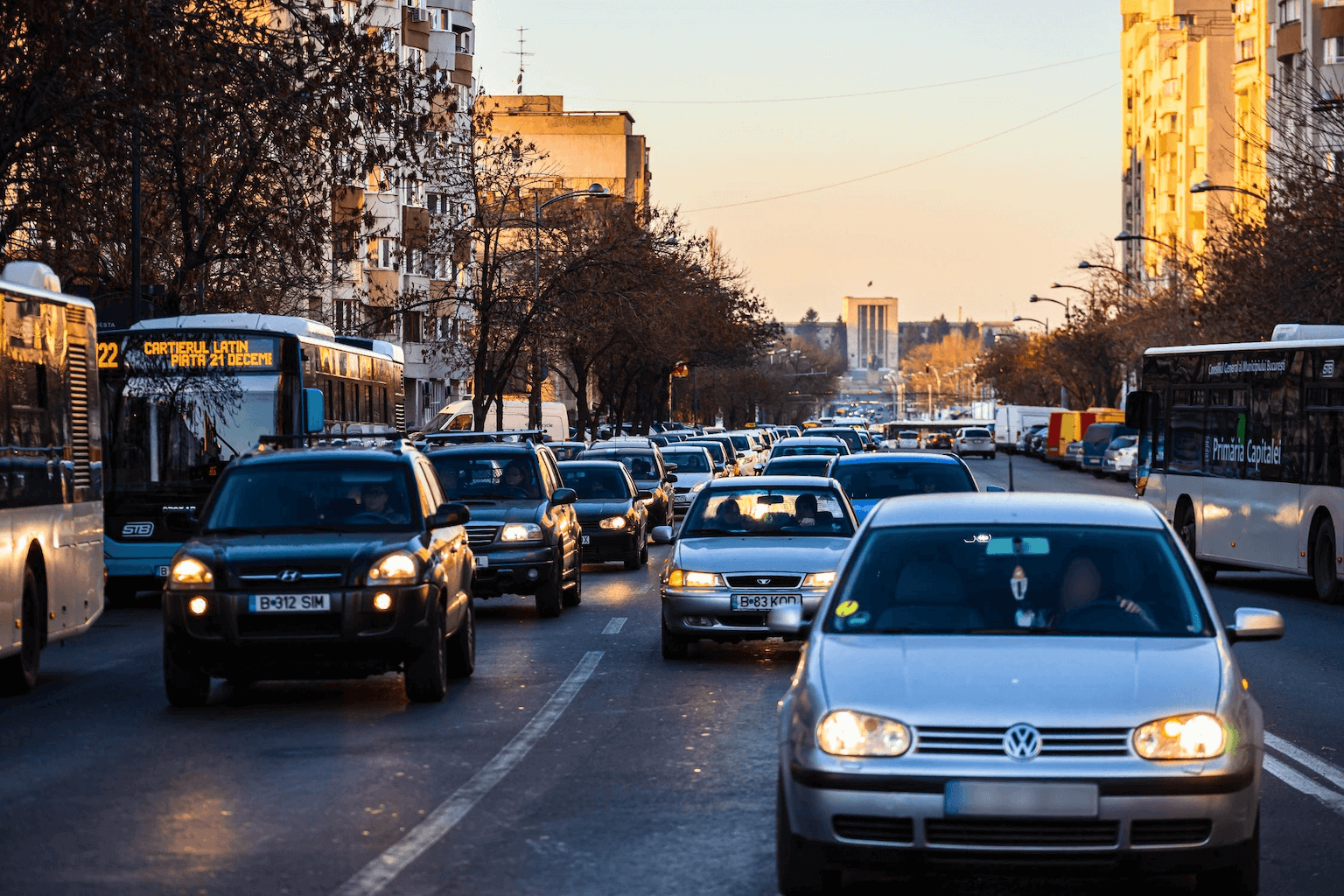}
            \caption{Original Image}
            \label{fig:column_a}
        \end{subfigure}
    \end{minipage} 
    \begin{minipage}[b]{0.195\linewidth}
        \begin{subfigure}[b]{\linewidth}
            \centering
            \includegraphics[width=\linewidth]{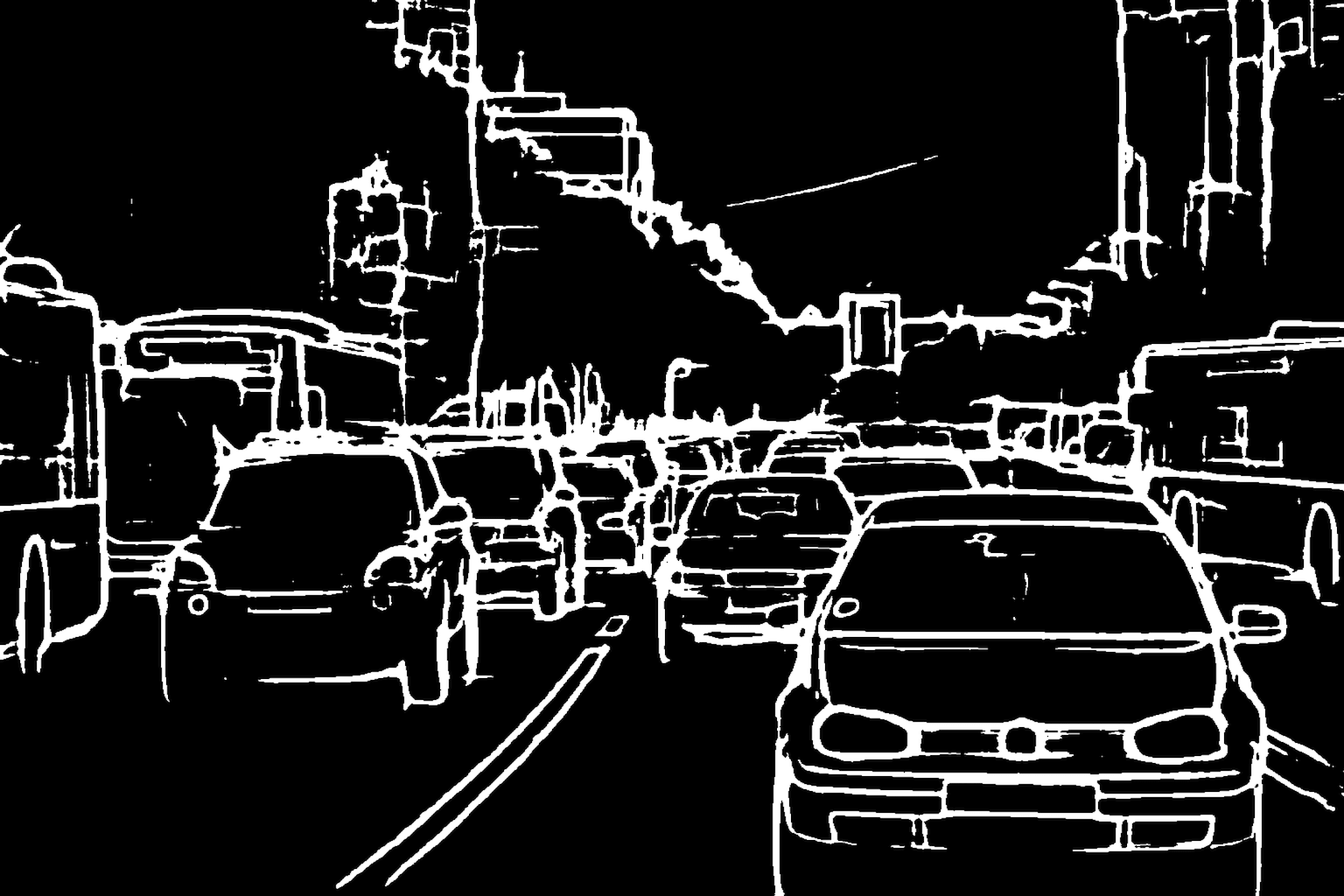}
            \caption{Edge Map}
            \label{fig:column_b}
        \end{subfigure}
    \end{minipage} 
    \begin{minipage}[b]{0.195\linewidth}
        \begin{subfigure}[b]{\linewidth}
            \centering
            \includegraphics[width=\linewidth]{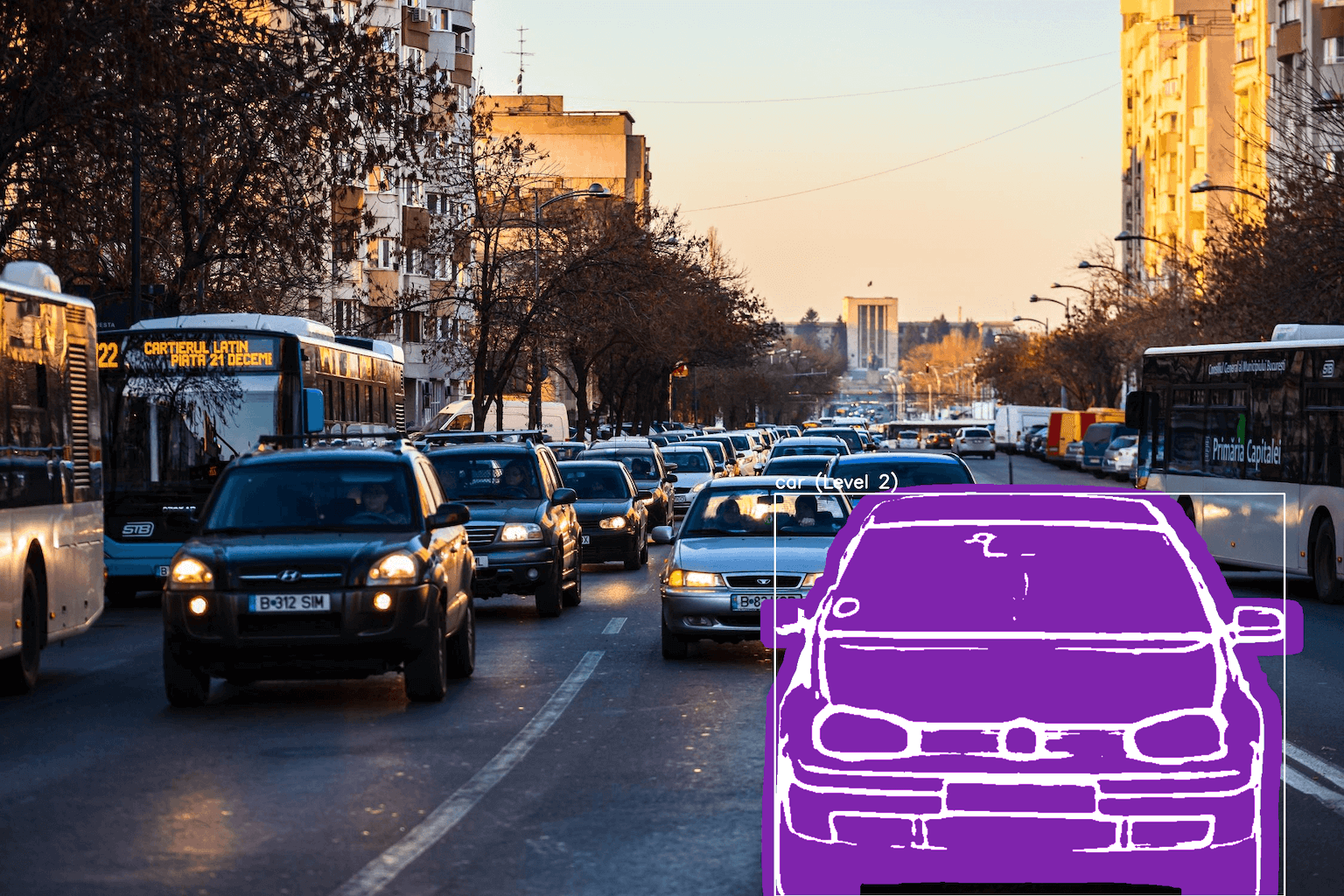}
            \caption{Chosen Mask}
            \label{fig:column_c}
        \end{subfigure}
    \end{minipage} 
    \begin{minipage}[b]{0.195\linewidth}
        \begin{subfigure}[b]{\linewidth}
            \centering
            \includegraphics[width=\linewidth]{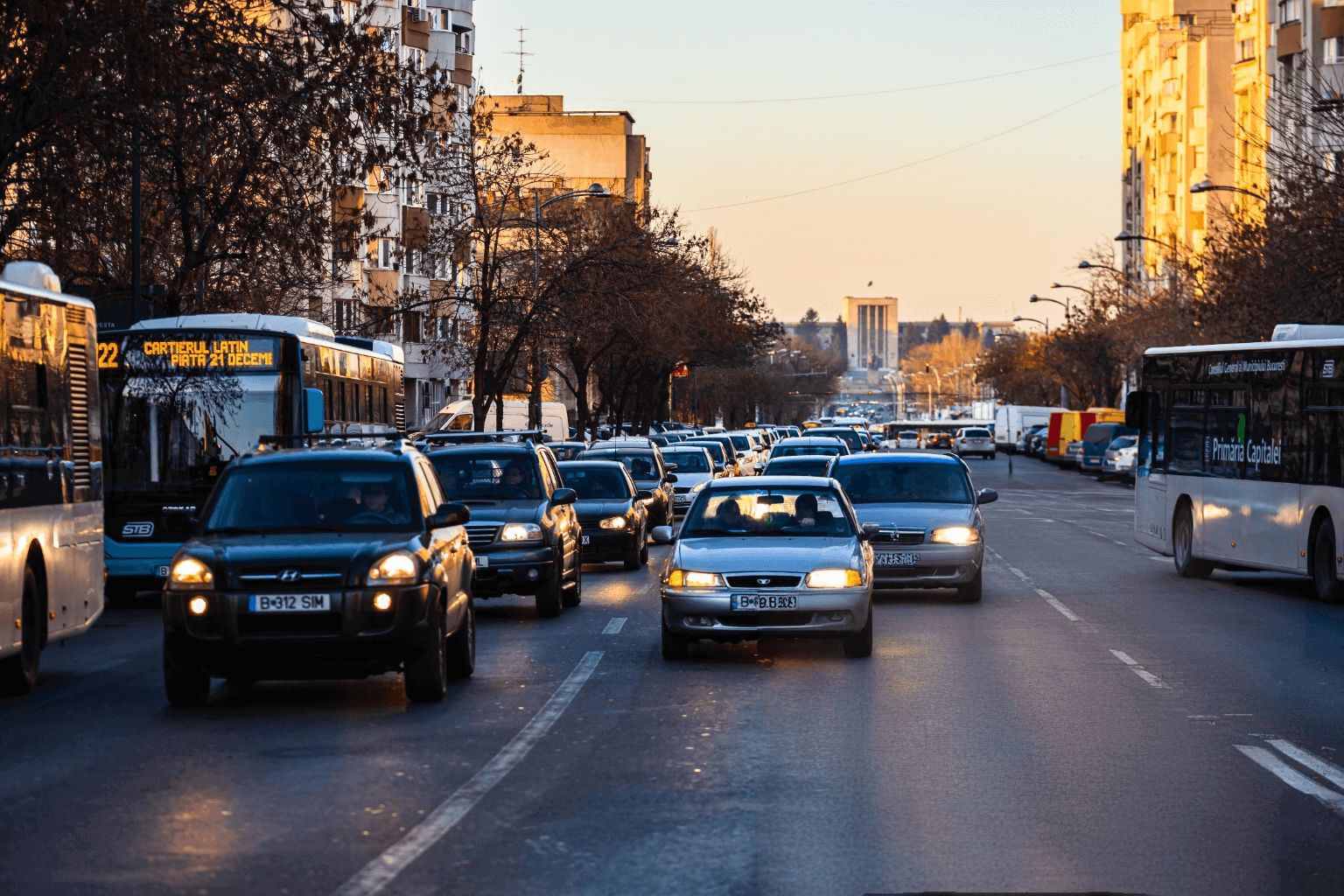}
            \caption{Inpainting Result}
            \label{fig:column_d}
        \end{subfigure}
    \end{minipage} 
    \begin{minipage}[b]{0.195\linewidth}
        \begin{subfigure}[b]{\linewidth}
            \centering
            \includegraphics[width=\linewidth]{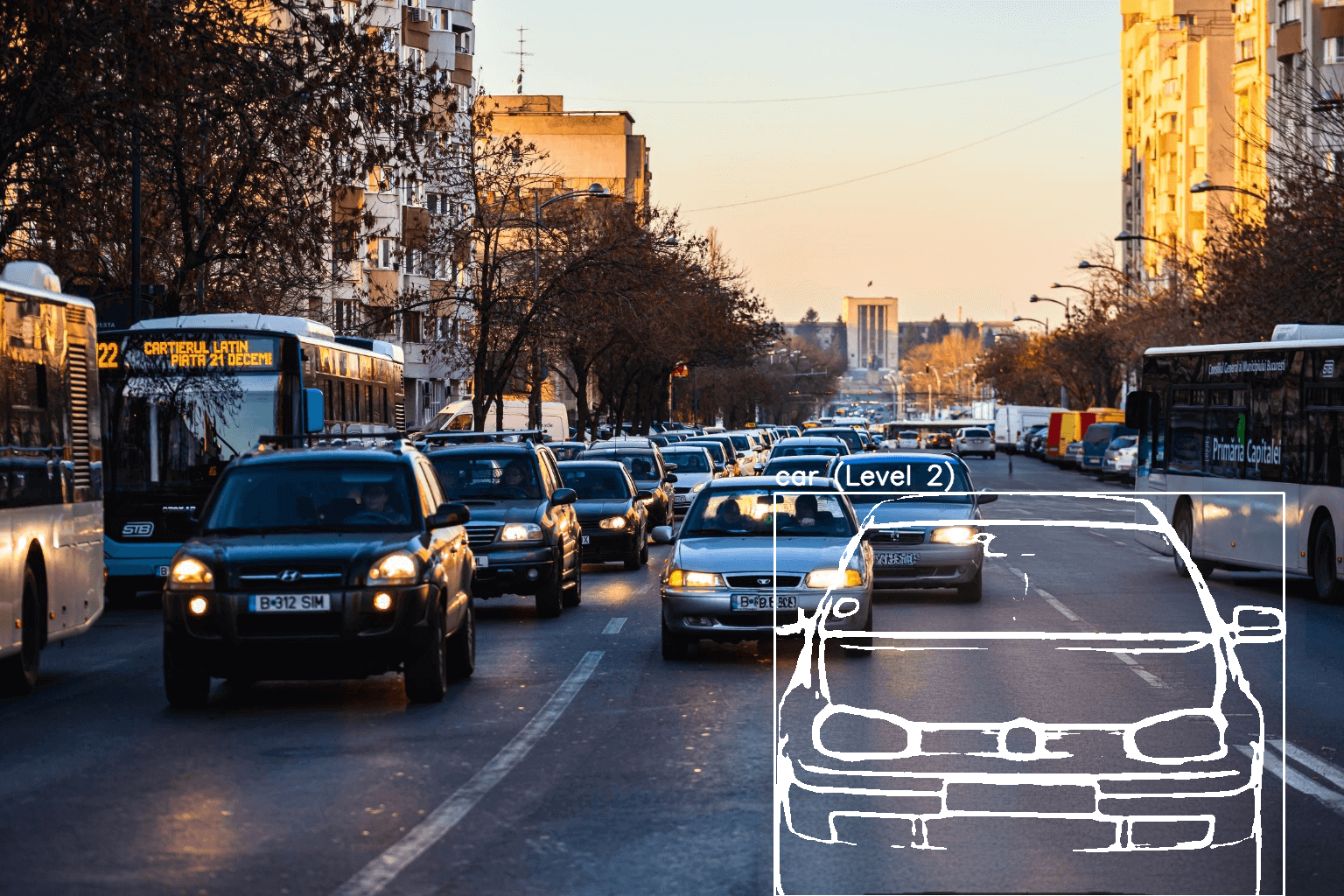}
            \caption{Edge Overlay}
            \label{fig:column_e}
        \end{subfigure}
    \end{minipage}
    \caption{Illustration of dataset construction process. (a) Original images from the DCI dataset; (b) Edge maps extracted from original images; (c) Selected masks (highlighted in purple) with highest edge density; (d) Results after BrushNet inpainting on augmented masked regions; (e) Final results with edge map overlay on selected areas. By overlaying edge maps on inpainted results, we simulate scenarios where users edit images with brush strokes, as the edge maps resemble hand-drawn sketches. The bounding box coordinates of the mask and labels are inherited from the DCI dataset.}
    \vspace{-0.4cm}
\end{figure*}

\noindent\textbf{Prompt formatting:} In our system, we implement two types of question answering (Q\&A)~\cite{vqa} tasks to facilitate the Draw\&Guess. 
For the add brush, we utilize a prompt structured as follows: 
\textit{``This is a `draw and guess' game. I will upload an image containing some strokes. To help you locate the strokes, I will give you the normalized bounding box coordinates of the stokes where their original coordinates are divided by the padded image width and height. The top-left corner of the bounding box is at $({x_1}, {y_1})$, and the bottom-right corner is at $({x_2}, {y_2})$. Now tell me in a single word a phrase, what am I trying to draw with these strokes in the image?''} 
The Q\&A output directly serves as the predicted prompt. For the subtract brush, we bypass the Q\&A process, as the results demonstrate that prompt-free generation achieves satisfactory results.

For the color brush, the Q\&A setup is similar: 
\textit{``The user will upload an image containing some contours in red color. To help you locate the contour, ... You need to identify what is inside the contours using a single word or phrase.''}, (the repetitive part is omitted). The system extracts contour information from the color brush stroke boundaries. The final predicted prompt is generated by combining the stroke's color information with Q\&A outputs. To optimize response time, we constrain Q\&A responses to concise, single-word or short-phrase formats.

For the color brush Q\&A task, accurate object recognition within contours is essential. LLaVA~\cite{llava} inherently excels in object recognition tasks, making it adept at identifying the content within color brush stroke boundaries. 
However, the interpretation of add brush strokes poses a significant challenge due to the inherent abstraction of human hand-drawn strokes or sketches. 
To address this, we find it necessary to construct a specialized dataset to fine-tune LLaVA to better understand and interpret human hand-drawn brush strokes.

\noindent \textbf{Dataset Construction:} We selected the Densely Captioned Images (DCI) dataset~\cite{DCI} as our primary source. Each image within the DCI dataset has detailed, multi-granular masks, accompanied by open-vocabulary labels and rich descriptions. This rich annotation structure enables the capture of diverse visual features and semantic contexts.

\textit{Step 1: Answer Generation for Q\&A. }
The initial stage involves generating edge maps using PiDiNet~\cite{pidinet} from images in the DCI dataset, as shown in Fig.~\ref{fig:column_b}. We calculate the edge density within the masked regions and select the top $5$ masks with the highest edge densities, as illustrated in Fig.~\ref{fig:column_c}. The labels corresponding to these selected masks serve as the ground truths for the Q\&A. To ensure the model focuses on guessing user intent rather than parsing irrelevant details, we clean the label to keep only noun components, streamlining to emphasize essential elements. 

\textit{Step 2: Simulating Brushstroke with Edge Overlay. }
In the second part of the dataset construction, we focus on the five masks identified in the first step. Each mask undergoes random shape expansion to introduce variability. We use the BrushNet~\cite{brushnet} model based on the SDXL~\cite{sdxl} to perform inpainting on these augmented masks with empty prompt, as shown in Fig.~\ref{fig:column_d}. Subsequently, the edge maps generated earlier are overlaid onto the inpainted areas as in Fig.~\ref{fig:column_e}. These overlay images simulate practical examples of how user hand-drawn strokes might alter an image.


\noindent\textbf{MLLM Fine-Tuning:} Our dataset construction method effectively prepares the model to understand and predict user edits, which contains a total of $24, 315$ images, categorized under $4, 412$ different labels, ensuring a broad spectrum of data for training. To optimize the performance of the MLLM over Draw\&Guess, we fine-tuned the LLaVA model, leveraging the Low-Rank Adaptation (LoRA)~\cite{lora} technique, allowing the efficient fine-tuning without extensively large dataset. Consistent with the original LLaVA training objectives, our approach aims to maximize the likelihood of the correct labels given the input corpora $u$, which is defined as
{
\small
\begin{equation}
\max _{\Theta^{lora}} \sum_{i=1}^{|u|} \log P\left(u_i \mid u_1, \ldots, u_{i-1} ; \{ \Theta^{pt} , \Theta^{lora} \} \right),
\label{eq:mllm}
\end{equation}
}

\noindent where $\Theta^{pt}$ and $\Theta^{lora}$ are parameters in the pre-trained MLLM and the LoRA respectively. 

\subsection{Idea Collector}
\label{idea_collector}
The user interface of \method is designed for an intuitive and streamlined image editing experience, as depicted in Figure~\ref{fig:design}. 
The interface is divided into several interactive sections, emphasizing ease of use while providing flexible control over the editing process. The interface comprises several key areas: a \textit{Prompt Area (A)} displaying MLLM-suggested prompts, a \textit{Toolbar (B)} with essential editing tools, \textit{Layer Management (C)} for organizing brush strokes, the main \textit{Canvas (D)} for editing, a \textit{Generated Images area (E)} for previewing results, \textit{Execute Button (F)}, and \textit{Parameter Adjustment (G)}.

\begin{figure*}[t]
    \centering
    \vspace{-0.2cm}
    \includegraphics[width=\linewidth]{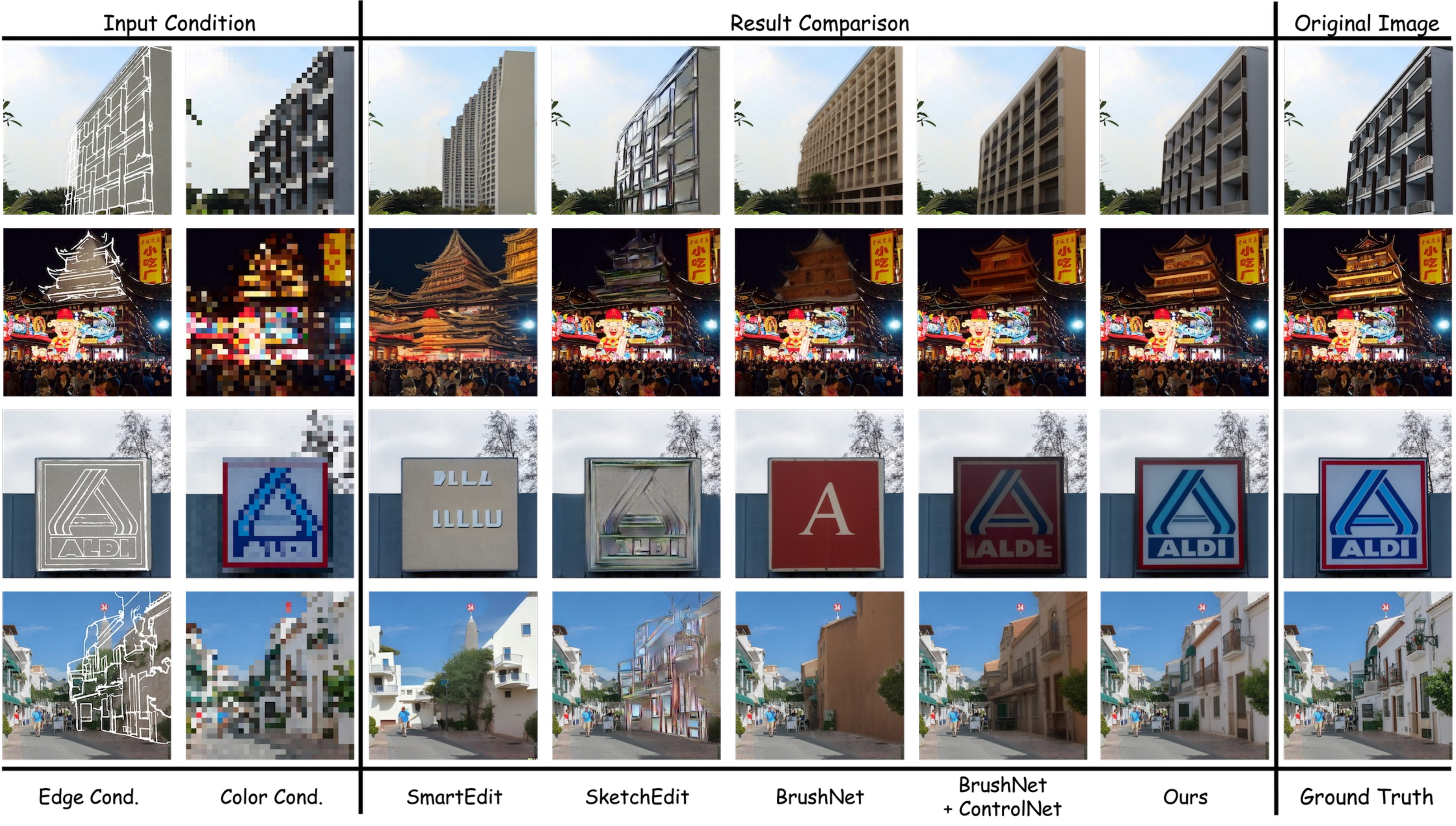}
    \vspace{-0.5cm}
    \caption{Visual result comparison. The first two columns present the edge and color conditions for editing, while the last column shows the ground truth image that the models aim to recreate. SmartEdit~\cite{smartedit} utilizes natural language for guidance, but lacks precision in controlling shape and color, often affecting non-target regions. SketchEdit~\cite{sketchedit}, a GAN-based approach~\cite{GAN}, struggles with open-domain image generation, falling short compared to models with diffusion-based generative priors. Although BrushNet~\cite{brushnet} delivers seamless image inpainting, it struggles to align edges and colors simultaneously, even with ControlNet~\cite{controlnet} enhancement. In contrast, our Editing Processor strictly adheres to both edge and color conditions, achieving high-fidelity conditional image editing.}
    \label{fig:exp}
    \vspace{-0.3cm}
\end{figure*}

\section{Experiment}
In evaluating our system, we focused on three primary modules: the \textbf{Editing Processor}, the \textbf{Painting Assistor}, and the \textbf{Idea Collector}. 
First, we assessed the quality of controllable generation provided by the Editing Processor, with particular attention to edge alignment and color fidelity. This evaluation involved analyzing how effectively users could manipulate and achieve desired visual outputs, which ensures the system responds accurately to user's control signal, detailed in Sec.~\ref{controllable_generation}.
Second, We evaluated the Painting Assistor's semantic prediction accuracy using simulated hand-drawn inputs. This assessment was critical for validating the capability of the MLLM in interpreting user intentions, ensuring contextually appropriate suggestions that align with the image semantics. Additionally, we conducted user studies to gather feedback on the system's efficiency improvements and prediction accuracy in real-world scenario, presented in Sec.~\ref{prediction_accuracy}.
Third, we assessed the usability of the user interfaces across all modules. We decomposes the assessment into four distinct dimensions spanning from operational efficiency to user satisfaction. This multi-dimensional assessment framework enabled systematic comparison with baseline systems while ensuring thorough evaluation of the interface, as shown in Sec.~\ref{idea_collection_effect}.

\subsection{Controllable Generation} 
\label{controllable_generation}
To thoroughly evaluate the controllable generation capabilities of our editing processor, we compared it with four representative baselines from different categories: (1) SmartEdit~\cite{smartedit}, an instruction-based editing method. We utilize LLaVA-Next~\cite{llava-next} to generate the editing instruction; (2) SketchEdit~\cite{sketchedit}, a GAN-based sketch-conditioned method; (3) BrushNet~\cite{brushnet}, the mask and prompt-guided inpainting method; and (4) a composite baseline combining BrushNet~\cite{brushnet} and ControlNet~\cite{controlnet}.
As illustrated in Fig.~\ref{fig:exp}, the instruction-based method, SmartEdit, tends to produce outputs that are too random, lacking the precision required for accurate editing purposes. Similarly, while BrushNet enables region-specific modifications, it struggles with maintaining predictable detail generation even with ControlNet enhancement, making precise manipulation challenging. In contrast, our model achieves more accurate edge alignment and color fidelity, which we attribute to our specialized design of the inpainting and control branch that emphasizes these aspects. 

\begin{table}[!h]
    \vspace{-0.2cm}
    \centering 
    \small
    \caption{Quantitative results and input condition comparisons between the baselines and ours. Our Editing Processor performs better than the baselines across all metrics, indicating its superiority in controllable generation over edge and color.}
    \vspace{-0.2cm}
    \SetTblrInner{rowsep=0.0pt}      
    \SetTblrInner{colsep=2.5pt}      
    \begin{tblr}{
        cells={halign=c,valign=m},   
        column{1}={halign=l},        
        hline{1,3,8}={1-9}{1.0pt},        
        hline{2}={2-4}{1.0pt},
        vline{2,5,6,7}={1-8}{1.0pt},          
        vline{3,4}={2-7}{1.0pt},
        cell{1}{1,5,6,7}={r=2}{},      
        cell{1}{2}={c=3}{},      
    }
    \ Method & Input Condition & & & LPIPS\cite{lpips} & PSNR & SSIM \\
     & Text & Edge & Color & \\
    SmartEdit & \ding{51} & \ding{55} & \ding{55} & $0.339$ & $16.695$ & $0.561$ \\
    SketchEdit & \ding{55} & \ding{51} & \ding{55} & $0.138$ & $23.288$ & $0.835$  \\
    BrushNet & \ding{51} & \ding{55} & \ding{55} & $0.0817$ & $25.455$ & $0.893$  \\
    Brush.+Cont. & \ding{51} & \ding{51} & \ding{51}  & $0.0748$ & $25.770$ & $0.894$  \\
    Ours & \ding{51} & \ding{51} & \ding{51} & $\textbf{0.0667}$ & $\textbf{27.282}$ & $\textbf{0.902}$  \\
    \end{tblr}
    \label{tab:control}
    \vspace{-0.5cm}
\end{table}

We conducted a quantitative analysis of our constructed test dataset in Sec.~\ref{draw_guess}, which contains $490$ images. Our model outperformed the baselines across all key metrics as in Tab.~\ref{tab:control}. These results demonstrate significant improvements in controllable generation.

We additionally compared two stroke-based editing methods SDEdit~\cite{sdedit} and UniPaint~\cite{unipaint}, and the qualitative results are shown below in Fig.~\ref{fig:comparison_stroke}.

\begin{figure}[h]
    \centering
    \vspace{-0.4cm}
    \includegraphics[width=\linewidth]{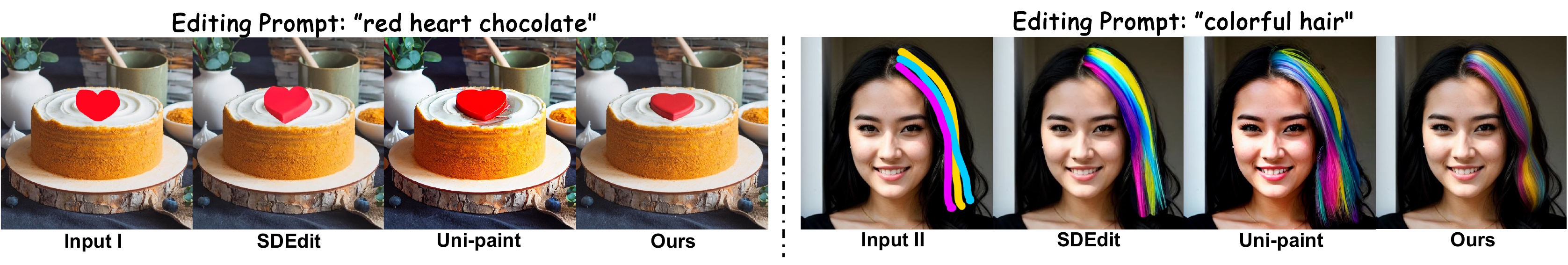}
    \vspace{-0.8cm}
    \caption{Visual comparison with stroke-based editing baselines.}
    \vspace{-0.5cm}
    \label{fig:comparison_stroke}
\end{figure}

\subsection{Prediction Accuracy \& Efficiency Facilitation}
\label{prediction_accuracy}
To evaluate the prediction accuracy of the Painting Assistor, we compared it with three state-of-the-art MLLMs: LLaVA-1.5~\cite{llava}, LLaVA-Next~\cite{llava-next}, and GPT-4o~\cite{gpt-4o} on our test dataset of $490$ images from Sec.~\ref{draw_guess}. Each model was prompted with images containing sketches and bounding box coordinates to generate semantic interpretations. The semantic outputs were assessed using three metrics: BERT~\cite{bert}, CLIP~\cite{clip}, and GPT-4~\cite{gpt4v} similarity scores, which measure the closeness of the generated descriptions to the ground truth. For GPT-4 similarity, we ask GPT-4 to rate the semantic and visual similarity between the predicted response and the ground truth on a 5-point scale, where $1$ means ``completely different'', $3$ means ``somewhat related'', and $5$ means ``exactly same''. 

\begin{table}[t]
    \vspace{-0.3cm}
    \centering 
    \small
    \caption{Performance comparison between our Painting Assistor and other MLLMs, demonstrating superior visual and semantic consistency in predictions.}
    \vspace{-0.3cm}
    \SetTblrInner{rowsep=0.0pt}      
    \SetTblrInner{colsep=3.0pt}      
    \begin{tblr}{
        cells={halign=c,valign=m},   
        column{1}={halign=l},        
        hline{1,3,7}={1-7}{1.0pt},        
        vline{2,3,4}={1-7}{1.0pt},          
        cell{1}{1}={r=2}{},      
    }
    \ Method & GPT-4~\cite{gpt4v} & BERT~\cite{bert}  & CLIP~\cite{clip}  \\
            & Similarity & Similarity  & Similarity \\
    LLaVA-1.5 & $1.894$ & $0.721$ & $0.795$  \\
    LLaVA-Next & $1.941$ & $0.716$ & $0.794$ \\
    GPT-4o & $1.976$ & $0.684$ & $0.790$ \\
    Ours & $\textbf{2.712}$ & $\textbf{0.749}$ & $\textbf{0.824}$ \\
    \end{tblr}
    \label{tab:predict}
    \vspace{-0.3cm}
\end{table}

The evaluation results are presented in Tab.~\ref{tab:predict}, illustrating that our model achieves the highest prediction accuracy among all tested MLLMs. This superior performance indicates that our Painting Assistor more accurately captures and predicts the semantic meanings of user drawings.

\begin{figure}[t]
    \centering
    \includegraphics[width=\linewidth]{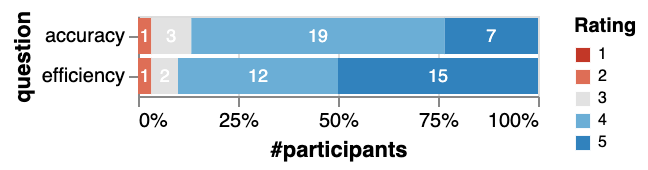}
    \vspace{-0.7cm}
    \caption{User ratings for the Painting Assistor, focusing on its prediction accuracy and efficiency enhancement capabilities.}
    \vspace{-0.6cm}
    \label{fig:user_study_mllm}
\end{figure}

To qualitatively evaluate the Painting Assistor, we conducted a user study with $30$ participants who freely edited images using our system. Participants rated the Painting Assistor on a $5$-point scale for prediction accuracy ($1$: very poor, $5$: excellent) and efficiency facilitation ($1$: significantly reduced, $5$: significantly enhanced). As shown in Fig.~\ref{fig:user_study_mllm}, $86.67\%$ of users rated prediction accuracy at least $4$, validating the ability of our fine-tuned MLLM to interpret user intentions. Similarly, $90\%$ rated efficiency facilitation $4$ or above, confirming that Draw\&Guess effectively streamlines the editing process by reducing manual prompt inputs. The average scores for accuracy and efficiency were $4.07$ and $4.37$.  We further provide a quantitative analysis with 10 users performing 10 edits, showing an average time savings of 24.92\% on iPad per edit and 19.58\% on PC per edit, as in the Tab.~\ref{tab:editing_time}.

\begin{table}[h]
    \vspace{-8pt}
    \centering 
    \scriptsize
    \SetTblrInner{rowsep=0.0pt}
    \SetTblrInner{colsep=3.0pt}
    \caption{Editing Time Comparison w./w.o. Painting Assistor.}
    \vspace{-10pt}
    \begin{tblr}{
        cells={halign=c,valign=m},
        column{1}={halign=l},
        hline{1,Z}={1-6}{1.0pt},
        hline{2}={1-6}{0.5pt},
        vline{4}={1-3}{1.0pt},
        cell{1}{2}={c=2}{},
        cell{1}{4}={c=2}{},
    }
    & \SetCell[c=2]{c} iPad & & \SetCell[c=2]{c} PC & \\
    & w. Paint. Assit. & w.o. Paint. Assit. & w. Paint. Assit. & w.o. Paint. Assit.\\
    \hline
    & 13.29s & 17.70s (+4.41s) & 12.49s & 15.53s (+3.04s) \\
    \end{tblr}
    \label{tab:editing_time}
    \vspace{-12pt}
\end{table}

\subsection{Idea Collection Effectiveness and Efficiency}
\label{idea_collection_effect}
Collecting user ideas effectively and efficiently is critical for the usability and adoption of interactive systems, especially in creative applications where user engagement is crucial. 
To evaluate the Idea Collector, we conducted a user study with $30$ participants, comparing our system against a baseline system on the following dimensions:

\begin{itemize}
    \item \textit{Complexity and Efficiency} measures how streamlined and intuitive the user finds the system for creative editing.
    \item \textit{Consistency and Integration} assesses whether the system maintains a cohesive interface and interaction design.
    \item \textit{Ease of Use} captures the learnability of the system, especially for users with varying backgrounds.
    \item \textit{Overall Satisfaction} reflects users' general satisfaction with the design, features, and usability of the system.
\end{itemize}


\noindent\textbf{Baseline:} The baseline system was implemented as a customized ComfyUI workflow, replacing our Idea Collector interface with an open-source canvas, Painter Node~\cite{PainterNode}. This setup enables the focus on the value provided with our Idea Collector by controlling other variables.

\noindent\textbf{Procedure:} The study lasted approximately $30$ minutes for each participant with two systems (our system and the baseline).
Each session began with a brief introduction to the system using the case illustrated in Fig.~\ref{fig:teaser}. 
Participants then had $5$ minutes to freely explore and edit images. 
After using both systems, participants completed a questionnaire with $22$ questions ($10$ questions per system covering all four dimensions and $2$ questions regarding the Painting Assistor detailed in Sec.~\ref{prediction_accuracy}). We employed the System Usability Scale (SUS)~\cite{SUS} for scoring, using a Likert scale from $1$ (strongly disagree) to $5$ (strongly agree), to capture a global view of subjective usability for each system.

\begin{figure}[t]
    \centering
    \vspace{-0.3cm}
    \includegraphics[width=\linewidth]{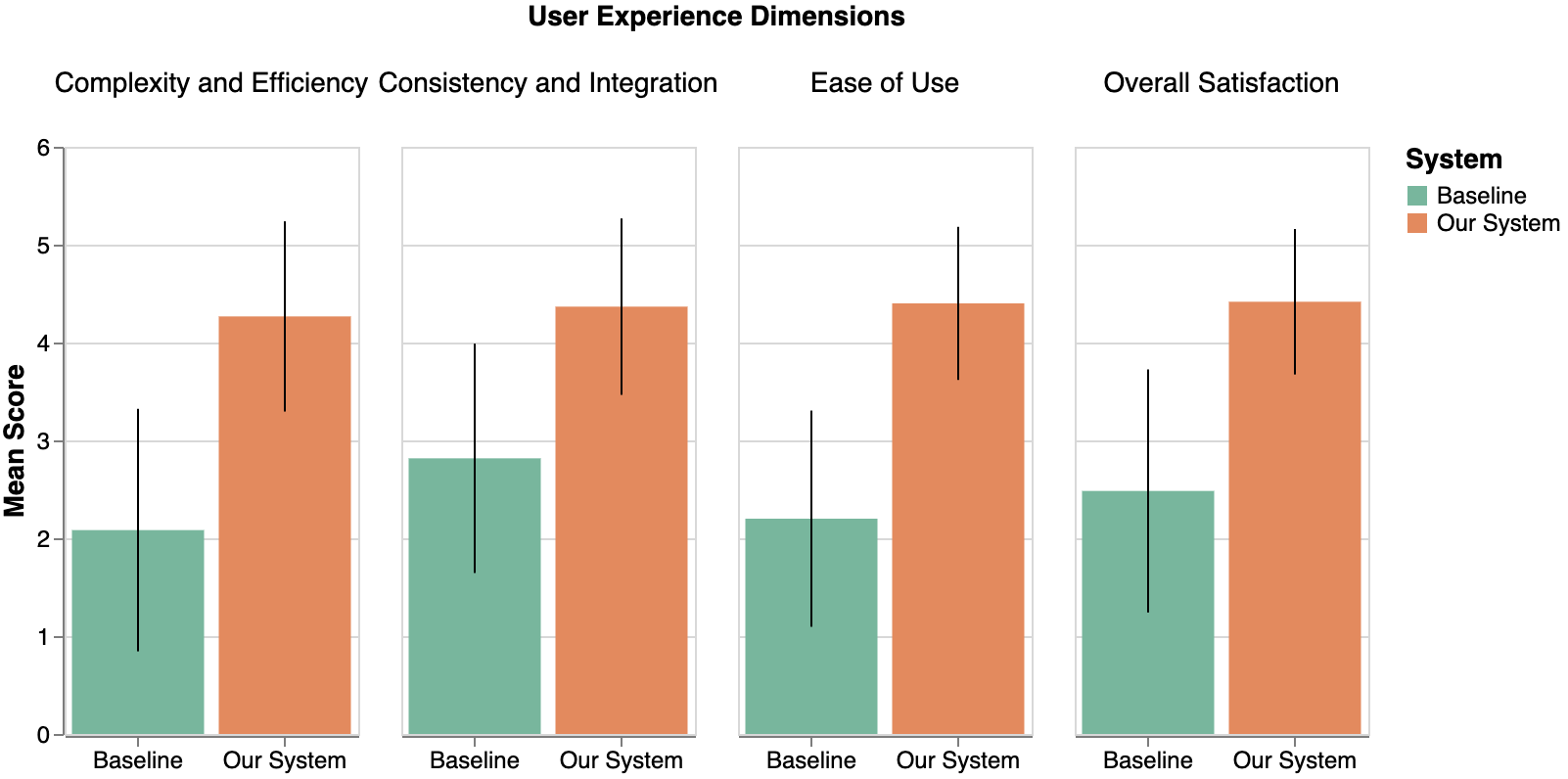}
    \vspace{-0.5cm}
    \caption{Comparative user ratings between our system and the baseline, with standard deviation shown as error bars.}
    \label{fig:user_study_4_dimensions}
    \vspace{-0.6cm}
\end{figure}

As shown in Fig.~\ref{fig:user_study_4_dimensions}, our system demonstrated significantly higher scores across all dimensions compared to the baseline. Indicating the effectiveness of our Idea Collector. Further details can be found in the supplementary.

\section{Conclusion}
In conclusion, our interactive image editing system \method effectively addresses the challenges of performing precise and efficient edits by combining the strengths of the Editing Processor, Painting Assistor, and Idea Collector. Our comprehensive evaluations demonstrate significant improvements over existing methods in terms of controllable generation quality, editing intent prediction accuracy, and user interface efficiency. For future work, we aim to expand the capabilities of our system by incorporating additional editing types, such as reference-based editing, which would allow users to guide modifications using external images. We also plan to implement layered image generation to provide better editing flexibility and support for complex compositions. Moreover, enhancing typography support will enable more robust manipulation of textual elements within images. These developments will further enrich our framework, offering users a more versatile and powerful tool for creative expression. The system is available at \href{https://magic-quill.github.io}{https://magic-quill.github.io}.

\noindent\textbf{Acknowledgments.} This work was supported by the Research Grant Council of the Hong Kong Special Administrative Region under grant number 16212623 and the Ant Group Research Intern Program.


{
    \small
    \bibliographystyle{ieeenat_fullname}
    \bibliography{main}

\begin{thebibliography}{84}
\providecommand{\natexlab}[1]{#1}
\providecommand{\url}[1]{\texttt{#1}}
\expandafter\ifx\csname urlstyle\endcsname\relax
  \providecommand{\doi}[1]{doi: #1}\else
  \providecommand{\doi}{doi: \begingroup \urlstyle{rm}\Url}\fi

\bibitem[Abid et~al.(2019)Abid, Abdalla, Abid, Khan, Alfozan, and Zou]{abid2019gradio}
Abubakar Abid, Ali Abdalla, Ali Abid, Dawood Khan, Abdulrahman Alfozan, and James Zou.
\newblock Gradio: Hassle-free sharing and testing of ml models in the wild.
\newblock \emph{arXiv preprint arXiv:1906.02569}, 2019.

\bibitem[Achiam et~al.(2023)Achiam, Adler, Agarwal, Ahmad, Akkaya, Aleman, Almeida, Altenschmidt, Altman, Anadkat, et~al.]{gpt4v}
Josh Achiam, Steven Adler, Sandhini Agarwal, Lama Ahmad, Ilge Akkaya, Florencia~Leoni Aleman, Diogo Almeida, Janko Altenschmidt, Sam Altman, Shyamal Anadkat, et~al.
\newblock Gpt-4 technical report.
\newblock \emph{arXiv preprint arXiv:2303.08774}, 2023.

\bibitem[Antol et~al.(2015)Antol, Agrawal, Lu, Mitchell, Batra, Zitnick, and Parikh]{vqa}
Stanislaw Antol, Aishwarya Agrawal, Jiasen Lu, Margaret Mitchell, Dhruv Batra, C~Lawrence Zitnick, and Devi Parikh.
\newblock Vqa: Visual question answering.
\newblock In \emph{Proceedings of the IEEE international conference on computer vision}, 2015.

\bibitem[Bai et~al.(2024)Bai, Ouyang, Xu, Wang, Yang, Cheng, Shen, and Chen]{edicho}
Qingyan Bai, Hao Ouyang, Yinghao Xu, Qiuyu Wang, Ceyuan Yang, Ka~Leong Cheng, Yujun Shen, and Qifeng Chen.
\newblock Edicho: Consistent image editing in the wild.
\newblock \emph{arXiv preprint arXiv:2412.21079}, 2024.

\bibitem[Brack et~al.(2023)Brack, Friedrich, Kornmeier, Tsaban, Schramowski, Kersting, and Passos]{ledits++}
Manuel Brack, Felix Friedrich, Katharina Kornmeier, Linoy Tsaban, Patrick Schramowski, Kristian Kersting, and Apolinário Passos.
\newblock Ledits++: Limitless image editing using text-to-image models.
\newblock \emph{Proceedings of the IEEE/CVF Conference on Computer Vision and Pattern Recognition}, 2023.

\bibitem[Brade et~al.(2023)Brade, Wang, Sousa, Oore, and Grossman]{promptify}
Stephen Brade, Bryan Wang, Mauricio Sousa, Sageev Oore, and Tovi Grossman.
\newblock Promptify: Text-to-image generation through interactive prompt exploration with large language models.
\newblock In \emph{Proceedings of the 36th Annual ACM Symposium on User Interface Software and Technology}, 2023.

\bibitem[Brooke et~al.(1996)]{SUS}
John Brooke et~al.
\newblock Sus-a quick and dirty usability scale.
\newblock \emph{Usability evaluation in industry}, 1996.

\bibitem[Brooks et~al.(2023)Brooks, Holynski, and Efros]{instructpix2pix}
Tim Brooks, Aleksander Holynski, and Alexei~A Efros.
\newblock Instructpix2pix: Learning to follow image editing instructions.
\newblock In \emph{Proceedings of the IEEE/CVF Conference on Computer Vision and Pattern Recognition}, 2023.

\bibitem[Chen et~al.(2023)Chen, Huang, Liu, Shen, Zhao, and Zhao]{anydoor}
Xi Chen, Lianghua Huang, Yu Liu, Yujun Shen, Deli Zhao, and Hengshuang Zhao.
\newblock Anydoor: Zero-shot object-level image customization.
\newblock \emph{arXiv preprint arXiv:2307.09481}, 2023.

\bibitem[Cheng et~al.(2025)Cheng, Wang, Shi, Zheng, Xu, Ouyang, Chen, and Shen]{agap}
Ka~Leong Cheng, Qiuyu Wang, Zifan Shi, Kecheng Zheng, Yinghao Xu, Hao Ouyang, Qifeng Chen, and Yujun Shen.
\newblock Learning naturally aggregated appearance for efficient 3d editing.
\newblock In \emph{Proceedings of the International Conference on 3D Vision}, 2025.

\bibitem[ComfyUI(2024)]{ComfyUI}
ComfyUI.
\newblock The most powerful and modular diffusion model gui, api and backend with a graph/nodes interface.
\newblock \url{https://github.com/comfyanonymous/ComfyUI}, 2024.

\bibitem[Devlin(2018)]{bert}
Jacob Devlin.
\newblock Bert: Pre-training of deep bidirectional transformers for language understanding.
\newblock \emph{arXiv preprint arXiv:1810.04805}, 2018.

\bibitem[Dong et~al.(2023)Dong, Han, Peng, Qi, Ge, Yang, Zhao, Sun, Zhou, Wei, et~al.]{dreamllm}
Runpei Dong, Chunrui Han, Yuang Peng, Zekun Qi, Zheng Ge, Jinrong Yang, Liang Zhao, Jianjian Sun, Hongyu Zhou, Haoran Wei, et~al.
\newblock Dreamllm: Synergistic multimodal comprehension and creation.
\newblock \emph{arXiv preprint arXiv:2309.11499}, 2023.

\bibitem[Epstein et~al.(2023)Epstein, Jabri, Poole, Efros, and Holynski]{self-guidance}
Dave Epstein, Allan Jabri, Ben Poole, Alexei Efros, and Aleksander Holynski.
\newblock Diffusion self-guidance for controllable image generation.
\newblock \emph{Advances in Neural Information Processing Systems}, 2023.

\bibitem[Feng et~al.(2024{\natexlab{a}})Feng, Ma, Wang, Qi, Chen, Chen, and Wang]{dit4edit}
Kunyu Feng, Yue Ma, Bingyuan Wang, Chenyang Qi, Haozhe Chen, Qifeng Chen, and Zeyu Wang.
\newblock Dit4edit: Diffusion transformer for image editing.
\newblock \emph{arXiv preprint arXiv:2411.03286}, 2024{\natexlab{a}}.

\bibitem[Feng et~al.(2024{\natexlab{b}})Feng, Wang, Wong, Wang, Lu, Zhu, Wang, and Chen]{PromptMagician}
Yingchaojie Feng, Xingbo Wang, Kam~Kwai Wong, Sijia Wang, Yuhong Lu, Minfeng Zhu, Baicheng Wang, and Wei Chen.
\newblock Promptmagician: Interactive prompt engineering for text-to-image creation.
\newblock \emph{IEEE Transactions on Visualization and Computer Graphics}, 2024{\natexlab{b}}.

\bibitem[Fu et~al.(2023)Fu, Hu, Du, Wang, Yang, and Gan]{MGIE}
Tsu-Jui Fu, Wenze Hu, Xianzhi Du, William~Yang Wang, Yinfei Yang, and Zhe Gan.
\newblock Guiding instruction-based image editing via multimodal large language models.
\newblock \emph{arXiv preprint arXiv:2309.17102}, 2023.

\bibitem[Ge et~al.(2024)Ge, Zhao, Zhu, Ge, Yi, Song, Li, Ding, and Shan]{seedx}
Yuying Ge, Sijie Zhao, Jinguo Zhu, Yixiao Ge, Kun Yi, Lin Song, Chen Li, Xiaohan Ding, and Ying Shan.
\newblock Seed-x: Multimodal models with unified multi-granularity comprehension and generation.
\newblock \emph{arXiv preprint arXiv:2404.14396}, 2024.

\bibitem[Geng et~al.(2024)Geng, Yang, Hang, Li, Gu, Zhang, Bao, Zhang, Li, Hu, et~al.]{instructdiffusion}
Zigang Geng, Binxin Yang, Tiankai Hang, Chen Li, Shuyang Gu, Ting Zhang, Jianmin Bao, Zheng Zhang, Houqiang Li, Han Hu, et~al.
\newblock Instructdiffusion: A generalist modeling interface for vision tasks.
\newblock In \emph{Proceedings of the IEEE/CVF Conference on Computer Vision and Pattern Recognition}, 2024.

\bibitem[Goodfellow et~al.(2020)Goodfellow, Pouget-Abadie, Mirza, Xu, Warde-Farley, Ozair, Courville, and Bengio]{GAN}
Ian Goodfellow, Jean Pouget-Abadie, Mehdi Mirza, Bing Xu, David Warde-Farley, Sherjil Ozair, Aaron Courville, and Yoshua Bengio.
\newblock Generative adversarial networks.
\newblock \emph{Communications of the ACM}, 2020.

\bibitem[He et~al.(2024)He, Liu, Chen, Tian, Liu, Chi, Liu, Yuan, Xing, Wang, et~al.]{survey_mllm_generation}
Yingqing He, Zhaoyang Liu, Jingye Chen, Zeyue Tian, Hongyu Liu, Xiaowei Chi, Runtao Liu, Ruibin Yuan, Yazhou Xing, Wenhai Wang, et~al.
\newblock Llms meet multimodal generation and editing: A survey.
\newblock \emph{arXiv preprint arXiv:2405.19334}, 2024.

\bibitem[Ho et~al.(2020)Ho, Jain, and Abbeel]{DDPM}
Jonathan Ho, Ajay Jain, and Pieter Abbeel.
\newblock Denoising diffusion probabilistic models.
\newblock \emph{Advances in neural information processing systems}, 2020.

\bibitem[Hu et~al.(2021)Hu, Shen, Wallis, Allen-Zhu, Li, Wang, Wang, and Chen]{lora}
Edward~J Hu, Yelong Shen, Phillip Wallis, Zeyuan Allen-Zhu, Yuanzhi Li, Shean Wang, Lu Wang, and Weizhu Chen.
\newblock Lora: Low-rank adaptation of large language models.
\newblock \emph{arXiv preprint arXiv:2106.09685}, 2021.

\bibitem[Huang et~al.(2024{\natexlab{a}})Huang, Huang, Liu, Yan, Lv, Liu, Xiong, Zhang, Chen, and Cao]{Survey_DiffusionImageEditing}
Yi Huang, Jiancheng Huang, Yifan Liu, Mingfu Yan, Jiaxi Lv, Jianzhuang Liu, Wei Xiong, He Zhang, Shifeng Chen, and Liangliang Cao.
\newblock Diffusion model-based image editing: A survey.
\newblock \emph{arXiv preprint arXiv:2402.17525}, 2024{\natexlab{a}}.

\bibitem[Huang et~al.(2024{\natexlab{b}})Huang, Xie, Wang, Yuan, Cun, Ge, Zhou, Dong, Huang, Zhang, et~al.]{smartedit}
Yuzhou Huang, Liangbin Xie, Xintao Wang, Ziyang Yuan, Xiaodong Cun, Yixiao Ge, Jiantao Zhou, Chao Dong, Rui Huang, Ruimao Zhang, et~al.
\newblock Smartedit: Exploring complex instruction-based image editing with multimodal large language models.
\newblock In \emph{Proceedings of the IEEE/CVF Conference on Computer Vision and Pattern Recognition}, 2024{\natexlab{b}}.

\bibitem[Hurst et~al.(2024)Hurst, Lerer, Goucher, Perelman, Ramesh, Clark, Ostrow, Welihinda, Hayes, Radford, et~al.]{gpt-4o}
Aaron Hurst, Adam Lerer, Adam~P Goucher, Adam Perelman, Aditya Ramesh, Aidan Clark, AJ Ostrow, Akila Welihinda, Alan Hayes, Alec Radford, et~al.
\newblock Gpt-4o system card.
\newblock \emph{arXiv preprint arXiv:2410.21276}, 2024.

\bibitem[Jo and Park(2019)]{sc-fegan}
Youngjoo Jo and Jongyoul Park.
\newblock Sc-fegan: Face editing generative adversarial network with user's sketch and color.
\newblock In \emph{Proceedings of the IEEE/CVF international conference on computer vision}, 2019.

\bibitem[Ju et~al.(2024{\natexlab{a}})Ju, Liu, Wang, Bian, Shan, and Xu]{brushnet}
Xuan Ju, Xian Liu, Xintao Wang, Yuxuan Bian, Ying Shan, and Qiang Xu.
\newblock Brushnet: A plug-and-play image inpainting model with decomposed dual-branch diffusion.
\newblock \emph{arXiv preprint arXiv:2403.06976}, 2024{\natexlab{a}}.

\bibitem[Ju et~al.(2024{\natexlab{b}})Ju, Zeng, Bian, Liu, and Xu]{pnp}
Xuan Ju, Ailing Zeng, Yuxuan Bian, Shaoteng Liu, and Qiang Xu.
\newblock Pnp inversion: Boosting diffusion-based editing with 3 lines of code.
\newblock \emph{International Conference on Learning Representations}, 2024{\natexlab{b}}.

\bibitem[Karras et~al.(2022)Karras, Aittala, Aila, and Laine]{karras}
Tero Karras, Miika Aittala, Timo Aila, and Samuli Laine.
\newblock Elucidating the design space of diffusion-based generative models.
\newblock \emph{Advances in neural information processing systems}, 2022.

\bibitem[Kawar et~al.(2023)Kawar, Zada, Lang, Tov, Chang, Dekel, Mosseri, and Irani]{imagic}
Bahjat Kawar, Shiran Zada, Oran Lang, Omer Tov, Huiwen Chang, Tali Dekel, Inbar Mosseri, and Michal Irani.
\newblock Imagic: Text-based real image editing with diffusion models.
\newblock In \emph{Conference on Computer Vision and Pattern Recognition 2023}, 2023.

\bibitem[Kim et~al.(2023)Kim, Park, Lee, and Choo]{reference_sketch}
Kangyeol Kim, Sunghyun Park, Junsoo Lee, and Jaegul Choo.
\newblock Reference-based image composition with sketch via structure-aware diffusion model.
\newblock \emph{arXiv preprint arXiv:2304.09748}, 2023.

\bibitem[Kingma(2013)]{vae}
Diederik~P Kingma.
\newblock Auto-encoding variational bayes.
\newblock \emph{arXiv preprint arXiv:1312.6114}, 2013.

\bibitem[Ko et~al.(2023)Ko, Park, Jeon, Jo, Kim, and Seo]{Ko2023LargeScale}
Hyung-Kwon Ko, Gwanmo Park, Hyeon Jeon, Jaemin Jo, Juho Kim, and Jinwook Seo.
\newblock Large-scale text-to-image generation models for visual artists’ creative works.
\newblock In \emph{Proceedings of the 28th International Conference on Intelligent User Interfaces}, 2023.

\bibitem[Koh et~al.(2024)Koh, Fried, and Salakhutdinov]{gill}
Jing~Yu Koh, Daniel Fried, and Russ~R Salakhutdinov.
\newblock Generating images with multimodal language models.
\newblock \emph{Advances in Neural Information Processing Systems}, 2024.

\bibitem[Liu et~al.(2024{\natexlab{a}})Liu, Li, Li, and Lee]{improved_instruction}
Haotian Liu, Chunyuan Li, Yuheng Li, and Yong~Jae Lee.
\newblock Improved baselines with visual instruction tuning.
\newblock In \emph{Proceedings of the IEEE/CVF Conference on Computer Vision and Pattern Recognition}, 2024{\natexlab{a}}.

\bibitem[Liu et~al.(2024{\natexlab{b}})Liu, Li, Li, Li, Zhang, Shen, and Lee]{llava-next}
Haotian Liu, Chunyuan Li, Yuheng Li, Bo Li, Yuanhan Zhang, Sheng Shen, and Yong~Jae Lee.
\newblock Llava-next: Improved reasoning, ocr, and world knowledge, 2024{\natexlab{b}}.

\bibitem[Liu et~al.(2024{\natexlab{c}})Liu, Li, Wu, and Lee]{llava}
Haotian Liu, Chunyuan Li, Qingyang Wu, and Yong~Jae Lee.
\newblock Visual instruction tuning.
\newblock \emph{Advances in neural information processing systems}, 2024{\natexlab{c}}.

\bibitem[Liu et~al.(2023{\natexlab{a}})Liu, Feng, Zhu, Zhang, Zheng, Liu, Zhao, Zhou, and Cao]{cones}
Zhiheng Liu, Ruili Feng, Kai Zhu, Yifei Zhang, Kecheng Zheng, Yu Liu, Deli Zhao, Jingren Zhou, and Yang Cao.
\newblock Cones: Concept neurons in diffusion models for customized generation.
\newblock \emph{arXiv preprint arXiv:2303.05125}, 2023{\natexlab{a}}.

\bibitem[Liu et~al.(2023{\natexlab{b}})Liu, Zhang, Shen, Zheng, Zhu, Feng, Liu, Zhao, Zhou, and Cao]{liu2023cones2}
Zhiheng Liu, Yifei Zhang, Yujun Shen, Kecheng Zheng, Kai Zhu, Ruili Feng, Yu Liu, Deli Zhao, Jingren Zhou, and Yang Cao.
\newblock Cones 2: Customizable image synthesis with multiple subjects.
\newblock In \emph{Proceedings of the 37th International Conference on Neural Information Processing Systems}, 2023{\natexlab{b}}.

\bibitem[Liu et~al.(2025)Liu, Cheng, Chen, Xiao, Ouyang, Zhu, Liu, Shen, Chen, and Luo]{manganinja}
Zhiheng Liu, Ka~Leong Cheng, Xi Chen, Jie Xiao, Hao Ouyang, Kai Zhu, Yu Liu, Yujun Shen, Qifeng Chen, and Ping Luo.
\newblock Manganinja: Line art colorization with precise reference following.
\newblock \emph{arXiv preprint arXiv:2501.08332}, 2025.

\bibitem[Mao et~al.(2023)Mao, Han, and Wang]{sketchffusion}
Weihang Mao, Bo Han, and Zihao Wang.
\newblock Sketchffusion: Sketch-guided image editing with diffusion model.
\newblock In \emph{2023 IEEE International Conference on Image Processing (ICIP)}, 2023.

\bibitem[Matsunaga et~al.(2022)Matsunaga, Ishii, Hayakawa, Suzuki, and Narihira]{pixel-diffusion}
Naoki Matsunaga, Masato Ishii, Akio Hayakawa, Kenji Suzuki, and Takuya Narihira.
\newblock Fine-grained image editing by pixel-wise guidance using diffusion models.
\newblock \emph{arXiv preprint arXiv:2212.02024}, 2022.

\bibitem[Meng et~al.(2022)Meng, He, Song, Song, Wu, Zhu, and Ermon]{sdedit}
Chenlin Meng, Yutong He, Yang Song, Jiaming Song, Jiajun Wu, Jun-Yan Zhu, and Stefano Ermon.
\newblock {SDE}dit: Guided image synthesis and editing with stochastic differential equations.
\newblock In \emph{International Conference on Learning Representations}, 2022.

\bibitem[Mou et~al.(2023)Mou, Wang, Song, Shan, and Zhang]{dragondiffusion}
Chong Mou, Xintao Wang, Jiechong Song, Ying Shan, and Jian Zhang.
\newblock Dragondiffusion: Enabling drag-style manipulation on diffusion models.
\newblock \emph{arXiv preprint arXiv:2307.02421}, 2023.

\bibitem[Nie et~al.(2023)Nie, Guo, Lu, Zhou, Zheng, and Li]{sde-drag}
Shen Nie, Hanzhong~Allan Guo, Cheng Lu, Yuhao Zhou, Chenyu Zheng, and Chongxuan Li.
\newblock The blessing of randomness: Sde beats ode in general diffusion-based image editing.
\newblock \emph{arXiv preprint arXiv:2311.01410}, 2023.

\bibitem[Pan et~al.(2023)Pan, Tewari, Leimk{\"u}hler, Liu, Meka, and Theobalt]{draggan}
Xingang Pan, Ayush Tewari, Thomas Leimk{\"u}hler, Lingjie Liu, Abhimitra Meka, and Christian Theobalt.
\newblock Drag your gan: Interactive point-based manipulation on the generative image manifold.
\newblock In \emph{ACM SIGGRAPH 2023 Conference Proceedings}, 2023.

\bibitem[Peng et~al.(2024)Peng, Koch, and Mackay]{DesignPrompt}
Xiaohan Peng, Janin Koch, and Wendy~E. Mackay.
\newblock Designprompt: Using multimodal interaction for design exploration with generative ai.
\newblock In \emph{Proceedings of the 2024 ACM Designing Interactive Systems Conference}, 2024.

\bibitem[Petrov(2024)]{PainterNode}
Aleksey Petrov.
\newblock Comfyui custom nodes alekpet.
\newblock \url{https://github.com/AlekPet/ComfyUI_Custom_Nodes_AlekPet}, 2024.

\bibitem[Podell et~al.(2023)Podell, English, Lacey, Blattmann, Dockhorn, M{\"u}ller, Penna, and Rombach]{sdxl}
Dustin Podell, Zion English, Kyle Lacey, Andreas Blattmann, Tim Dockhorn, Jonas M{\"u}ller, Joe Penna, and Robin Rombach.
\newblock Sdxl: Improving latent diffusion models for high-resolution image synthesis.
\newblock \emph{arXiv preprint arXiv:2307.01952}, 2023.

\bibitem[Portenier et~al.(2018)Portenier, Hu, Szabo, Bigdeli, Favaro, and Zwicker]{faceshop}
Tiziano Portenier, Qiyang Hu, Attila Szabo, Siavash~Arjomand Bigdeli, Paolo Favaro, and Matthias Zwicker.
\newblock Faceshop: Deep sketch-based face image editing.
\newblock \emph{arXiv preprint arXiv:1804.08972}, 2018.

\bibitem[Radford et~al.(2021)Radford, Kim, Hallacy, Ramesh, Goh, Agarwal, Sastry, Askell, Mishkin, Clark, et~al.]{clip}
Alec Radford, Jong~Wook Kim, Chris Hallacy, Aditya Ramesh, Gabriel Goh, Sandhini Agarwal, Girish Sastry, Amanda Askell, Pamela Mishkin, Jack Clark, et~al.
\newblock Learning transferable visual models from natural language supervision.
\newblock In \emph{International conference on machine learning}, 2021.

\bibitem[Rombach et~al.(2022)Rombach, Blattmann, Lorenz, Esser, and Ommer]{LDM}
Robin Rombach, Andreas Blattmann, Dominik Lorenz, Patrick Esser, and Bj{\"o}rn Ommer.
\newblock High-resolution image synthesis with latent diffusion models.
\newblock In \emph{Proceedings of the IEEE/CVF conference on computer vision and pattern recognition}, 2022.

\bibitem[Ronneberger et~al.(2015)Ronneberger, Fischer, and Brox]{UNet}
Olaf Ronneberger, Philipp Fischer, and Thomas Brox.
\newblock U-net: Convolutional networks for biomedical image segmentation.
\newblock In \emph{Medical image computing and computer-assisted intervention--MICCAI 2015: 18th international conference, Munich, Germany, October 5-9, 2015, proceedings, part III 18}, 2015.

\bibitem[Schuhmann et~al.(2022)Schuhmann, Beaumont, Vencu, Gordon, Wightman, Cherti, Coombes, Katta, Mullis, Wortsman, et~al.]{laion5b}
Christoph Schuhmann, Romain Beaumont, Richard Vencu, Cade Gordon, Ross Wightman, Mehdi Cherti, Theo Coombes, Aarush Katta, Clayton Mullis, Mitchell Wortsman, et~al.
\newblock Laion-5b: An open large-scale dataset for training next generation image-text models.
\newblock \emph{Advances in Neural Information Processing Systems}, 2022.

\bibitem[Sheynin et~al.(2024)Sheynin, Polyak, Singer, Kirstain, Zohar, Ashual, Parikh, and Taigman]{emuedit}
Shelly Sheynin, Adam Polyak, Uriel Singer, Yuval Kirstain, Amit Zohar, Oron Ashual, Devi Parikh, and Yaniv Taigman.
\newblock Emu edit: Precise image editing via recognition and generation tasks.
\newblock In \emph{Proceedings of the IEEE/CVF Conference on Computer Vision and Pattern Recognition}, 2024.

\bibitem[Singh et~al.(2024)Singh, Zhang, Liu, Smith, Lin, and Zheng]{smartmask}
Jaskirat Singh, Jianming Zhang, Qing Liu, Cameron Smith, Zhe Lin, and Liang Zheng.
\newblock Smartmask: Context aware high-fidelity mask generation for fine-grained object insertion and layout control.
\newblock In \emph{Proceedings of the IEEE/CVF Conference on Computer Vision and Pattern Recognition}, 2024.

\bibitem[Song et~al.(2020)Song, Meng, and Ermon]{DDIM}
Jiaming Song, Chenlin Meng, and Stefano Ermon.
\newblock Denoising diffusion implicit models.
\newblock \emph{arXiv preprint arXiv:2010.02502}, 2020.

\bibitem[Song et~al.(2024)Song, Zhang, Lin, Cohen, Price, Zhang, Kim, Zhang, Xiong, and Aliaga]{imprint}
Yizhi Song, Zhifei Zhang, Zhe Lin, Scott Cohen, Brian Price, Jianming Zhang, Soo~Ye Kim, He Zhang, Wei Xiong, and Daniel Aliaga.
\newblock Imprint: Generative object compositing by learning identity-preserving representation.
\newblock \emph{arXiv preprint arXiv:2403.10701}, 2024.

\bibitem[Soria et~al.(2023)Soria, Li, Rouhani, and Sappa]{teed}
Xavier Soria, Yachuan Li, Mohammad Rouhani, and Angel~D. Sappa.
\newblock Tiny and efficient model for the edge detection generalization.
\newblock In \emph{Proceedings of the IEEE/CVF International Conference on Computer Vision Workshops}, 2023.

\bibitem[Su et~al.(2021)Su, Liu, Yu, Hu, Liao, Tian, Pietik{\"a}inen, and Liu]{pidinet}
Zhuo Su, Wenzhe Liu, Zitong Yu, Dewen Hu, Qing Liao, Qi Tian, Matti Pietik{\"a}inen, and Li Liu.
\newblock Pixel difference networks for efficient edge detection.
\newblock In \emph{Proceedings of the IEEE/CVF international conference on computer vision}, 2021.

\bibitem[Sun et~al.(2023{\natexlab{a}})Sun, Fu, Hu, Wang, Rassin, Juan, Alon, Herrmann, van Steenkiste, Krishna, et~al.]{dreamsync}
Jiao Sun, Deqing Fu, Yushi Hu, Su Wang, Royi Rassin, Da-Cheng Juan, Dana Alon, Charles Herrmann, Sjoerd van Steenkiste, Ranjay Krishna, et~al.
\newblock Dreamsync: Aligning text-to-image generation with image understanding feedback.
\newblock In \emph{Synthetic Data for Computer Vision Workshop@ CVPR 2024}, 2023{\natexlab{a}}.

\bibitem[Sun et~al.(2023{\natexlab{b}})Sun, Yu, Cui, Zhang, Zhang, Wang, Gao, Liu, Huang, and Wang]{emu}
Quan Sun, Qiying Yu, Yufeng Cui, Fan Zhang, Xiaosong Zhang, Yueze Wang, Hongcheng Gao, Jingjing Liu, Tiejun Huang, and Xinlong Wang.
\newblock Generative pretraining in multimodality.
\newblock \emph{arXiv preprint arXiv:2307.05222}, 2023{\natexlab{b}}.

\bibitem[Sun et~al.(2024)Sun, Cui, Zhang, Zhang, Yu, Wang, Rao, Liu, Huang, and Wang]{emu2}
Quan Sun, Yufeng Cui, Xiaosong Zhang, Fan Zhang, Qiying Yu, Yueze Wang, Yongming Rao, Jingjing Liu, Tiejun Huang, and Xinlong Wang.
\newblock Generative multimodal models are in-context learners.
\newblock In \emph{Proceedings of the IEEE/CVF Conference on Computer Vision and Pattern Recognition}, 2024.

\bibitem[Urbanek et~al.(2024)Urbanek, Bordes, Astolfi, Williamson, Sharma, and Romero-Soriano]{DCI}
Jack Urbanek, Florian Bordes, Pietro Astolfi, Mary Williamson, Vasu Sharma, and Adriana Romero-Soriano.
\newblock A picture is worth more than 77 text tokens: Evaluating clip-style models on dense captions.
\newblock In \emph{Proceedings of the IEEE/CVF Conference on Computer Vision and Pattern Recognition}, 2024.

\bibitem[Wang et~al.(2024{\natexlab{a}})Wang, Lin, Yuan, Cheng, Wang, GH, Chen, and Peng]{flexedit}
Jue Wang, Yuxiang Lin, Tianshuo Yuan, Zhi-Qi Cheng, Xiaolong Wang, Jiao GH, Wei Chen, and Xiaojiang Peng.
\newblock Flexedit: Marrying free-shape masks to vllm for flexible image editing.
\newblock \emph{arXiv preprint arXiv:2408.12429}, 2024{\natexlab{a}}.

\bibitem[Wang et~al.(2021)Wang, Ouyang, and Chen]{inpainting_OU}
Tengfei Wang, Hao Ouyang, and Qifeng Chen.
\newblock Image inpainting with external-internal learning and monochromic bottleneck.
\newblock In \emph{Proceedings of the IEEE/CVF Conference on Computer Vision and Pattern Recognition}, 2021.

\bibitem[Wang et~al.(2024{\natexlab{b}})Wang, Huang, Song, Ma, and Zhang]{promptcharm}
Zhijie Wang, Yuheng Huang, Da Song, Lei Ma, and Tianyi Zhang.
\newblock Promptcharm: Text-to-image generation through multi-modal prompting and refinement.
\newblock In \emph{Proceedings of the CHI Conference on Human Factors in Computing Systems}, 2024{\natexlab{b}}.

\bibitem[Wang et~al.(2024{\natexlab{c}})Wang, Li, Li, and Liu]{genartist}
Zhenyu Wang, Aoxue Li, Zhenguo Li, and Xihui Liu.
\newblock Genartist: Multimodal llm as an agent for unified image generation and editing.
\newblock \emph{arXiv preprint arXiv:2407.05600}, 2024{\natexlab{c}}.

\bibitem[Winnemöller et~al.(2012)Winnemöller, Kyprianidis, and Olsen]{xdog}
Holger Winnemöller, Jan~Eric Kyprianidis, and Sven~C. Olsen.
\newblock Xdog: An extended difference-of-gaussians compendium including advanced image stylization.
\newblock \emph{Computers and Graphics}, 2012.

\bibitem[Xiao and Fu(2024)]{customsketching}
Chufeng Xiao and Hongbo Fu.
\newblock Customsketching: Sketch concept extraction for sketch-based image synthesis and editing.
\newblock \emph{arXiv preprint arXiv:2402.17624}, 2024.

\bibitem[Xiao et~al.(2024)Xiao, Wang, Zhou, Yuan, Xing, Yan, Wang, Huang, and Liu]{omnigen}
Shitao Xiao, Yueze Wang, Junjie Zhou, Huaying Yuan, Xingrun Xing, Ruiran Yan, Shuting Wang, Tiejun Huang, and Zheng Liu.
\newblock Omnigen: Unified image generation.
\newblock \emph{arXiv preprint arXiv:2409.11340}, 2024.

\bibitem[Xu et~al.(2024)Xu, Huang, Pan, Ma, and Chai]{infedit}
Sihan Xu, Yidong Huang, Jiayi Pan, Ziqiao Ma, and Joyce Chai.
\newblock Inversion-free image editing with natural language.
\newblock \emph{Proceedings of the IEEE/CVF Conference on Computer Vision and Pattern Recognition}, 2024.

\bibitem[Yang et~al.(2024{\natexlab{a}})Yang, Yu, Meng, Xu, Ermon, and Bin]{RPG}
Ling Yang, Zhaochen Yu, Chenlin Meng, Minkai Xu, Stefano Ermon, and CUI Bin.
\newblock Mastering text-to-image diffusion: Recaptioning, planning, and generating with multimodal llms.
\newblock In \emph{Forty-first International Conference on Machine Learning}, 2024{\natexlab{a}}.

\bibitem[Yang et~al.(2020)Yang, Wang, Liu, and Guo]{plasticsurgery}
Shuai Yang, Zhangyang Wang, Jiaying Liu, and Zongming Guo.
\newblock Deep plastic surgery: Robust and controllable image editing with human-drawn sketches.
\newblock In \emph{Computer Vision--ECCV 2020: 16th European Conference, Glasgow, UK, August 23--28, 2020, Proceedings, Part XV 16}, 2020.

\bibitem[Yang et~al.(2023{\natexlab{a}})Yang, Chen, and Liao]{unipaint}
Shiyuan Yang, Xiaodong Chen, and Jing Liao.
\newblock Uni-paint: A unified framework for multimodal image inpainting with pretrained diffusion model.
\newblock In \emph{Proceedings of the 31st ACM International Conference on Multimedia}, 2023{\natexlab{a}}.

\bibitem[Yang et~al.(2024{\natexlab{b}})Yang, Peng, Shen, Yang, Hu, Qiu, Koike, et~al.]{imagebrush}
Yifan Yang, Houwen Peng, Yifei Shen, Yuqing Yang, Han Hu, Lili Qiu, Hideki Koike, et~al.
\newblock Imagebrush: Learning visual in-context instructions for exemplar-based image manipulation.
\newblock \emph{Advances in Neural Information Processing Systems}, 2024{\natexlab{b}}.

\bibitem[Yang et~al.(2023{\natexlab{b}})Yang, Wang, Li, Lin, Lin, Liu, and Wang]{idea2img}
Zhengyuan Yang, Jianfeng Wang, Linjie Li, Kevin Lin, Chung-Ching Lin, Zicheng Liu, and Lijuan Wang.
\newblock Idea2img: Iterative self-refinement with gpt-4v (ision) for automatic image design and generation.
\newblock \emph{arXiv preprint arXiv:2310.08541}, 2023{\natexlab{b}}.

\bibitem[Zeng et~al.(2022)Zeng, Lin, and Patel]{sketchedit}
Yu Zeng, Zhe Lin, and Vishal~M Patel.
\newblock Sketchedit: Mask-free local image manipulation with partial sketches.
\newblock In \emph{Proceedings of the IEEE/CVF conference on computer vision and pattern recognition}, 2022.

\bibitem[Zhang et~al.(2024)Zhang, Mo, Chen, Sun, and Su]{magicbrush}
Kai Zhang, Lingbo Mo, Wenhu Chen, Huan Sun, and Yu Su.
\newblock Magicbrush: A manually annotated dataset for instruction-guided image editing.
\newblock \emph{Advances in Neural Information Processing Systems}, 2024.

\bibitem[Zhang et~al.(2023{\natexlab{a}})Zhang, Rao, and Agrawala]{controlnet}
Lvmin Zhang, Anyi Rao, and Maneesh Agrawala.
\newblock Adding conditional control to text-to-image diffusion models.
\newblock In \emph{Proceedings of the IEEE/CVF International Conference on Computer Vision}, 2023{\natexlab{a}}.

\bibitem[Zhang et~al.(2018)Zhang, Isola, Efros, Shechtman, and Wang]{lpips}
Richard Zhang, Phillip Isola, Alexei~A Efros, Eli Shechtman, and Oliver Wang.
\newblock The unreasonable effectiveness of deep features as a perceptual metric.
\newblock In \emph{Proceedings of the IEEE conference on computer vision and pattern recognition}, 2018.

\bibitem[Zhang et~al.(2023{\natexlab{b}})Zhang, Guo, Yoo, Matsuo, and Iwasawa]{phd}
Xin Zhang, Jiaxian Guo, Paul Yoo, Yutaka Matsuo, and Yusuke Iwasawa.
\newblock Paste, inpaint and harmonize via denoising: Subject-driven image editing with pre-trained diffusion model.
\newblock \emph{arXiv preprint arXiv:2306.07596}, 2023{\natexlab{b}}.

\bibitem[Zhuang et~al.(2023)Zhuang, Zeng, Liu, Yuan, and Chen]{powerpaint}
Junhao Zhuang, Yanhong Zeng, Wenran Liu, Chun Yuan, and Kai Chen.
\newblock A task is worth one word: Learning with task prompts for high-quality versatile image inpainting.
\newblock \emph{arXiv preprint arXiv:2312.03594}, 2023.

\end{thebibliography}
}
\clearpage
\setcounter{page}{1}
\maketitlesupplementary

\section{Implementation Details}
\subsection{Editing Processor}
Our Editing Processor is built upon Stable Diffusion v1.5~\cite{LDM} and is compatible with all customized fine-tuned weights. We set the control parameters with inpainting strength $w_I=1.0$ and control strength $w_C=0.5$ for both edge and color control signals, while expanding the mask region by $15$ pixels during controllable inpainting.
We use \textit{\textbf{separate}} ControlNets to independently control edge and color. Although the two signals may conflict, our model blends them using adjustable control weights ($0.5$ by default), allowing users to achieve more precise control.
The generation process employs the Euler ancestral sampler with Karras scheduler~\cite{karras}, requiring $20$ steps per generation. On standard hardware, generating a $512 \times 512$ resolution image takes approximately $3$ seconds with $10$ GB VRAM consumption.
For the control branch, we conduct fine-tuning on the LAION-Aesthetics dataset~\cite{laion5b}, specifically selecting images with aesthetic scores above $6.5$. The training process spans $3$ epochs with a learning rate of $5e-6$ and batch size of $8$.

\begin{figure}[h]
    \centering
    \vspace{-5pt}
    \includegraphics[width=\linewidth]{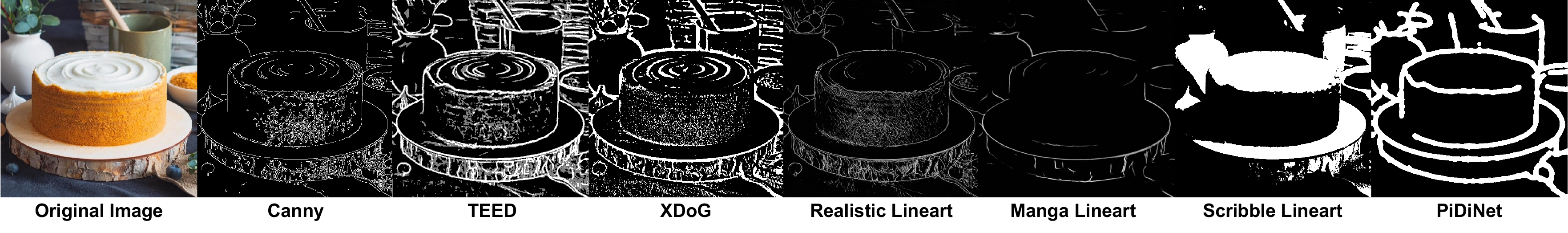}
    \vspace{-20pt}
    \caption{Extracted edge between different methods~\cite{pidinet, xdog, teed}.}
    \label{fig:edge}
    \vspace{-15pt}
\end{figure}

We choose PiDiNet~\cite{pidinet} as the edge extractor. Fig.~\ref{fig:edge} shows that it strikes a better balance between geometric structure preservation and simulation of human-like strokes.

\subsection{Painting Assistor} We fine-tune a LLaVA-1.5 model with $7B$ parameters for Draw\&Guess task on our own constructed dataset in Sec.~\ref{draw_guess}, leveraging LoRA~\cite{lora}. The LoRA rank and alpha are $64$ and $16$ respectively. The model is trained for $3$ epochs with a learning rate of $2e-5$ and batch size of $8$, taking $5$ hours on 3$\times$A6000 GPUs. Under $4$-bit quantization, the model achieves real-time prompt inference within $0.3$ seconds using only $5$ GB VRAM, enabling efficient on-the-fly prompt generation with satisfactory accuracy.  

\subsection{Idea Collector}
\noindent\textbf{Cross-platform Support:}
We implement the Idea Collector as a modular ReactJS component library, designed for cross-platform compatibility with various generative AI frameworks, such as Gradio~\cite{abid2019gradio} and ComfyUI~\cite{ComfyUI}. The architecture separates client-side user interactions from server-side model computations through HTTP protocols, enabling platform-independent deployment via standard HTML rendering.

\begin{figure}[ht]
    \centering
    \vspace{-0.4cm}
    \includegraphics[width=0.8\linewidth]{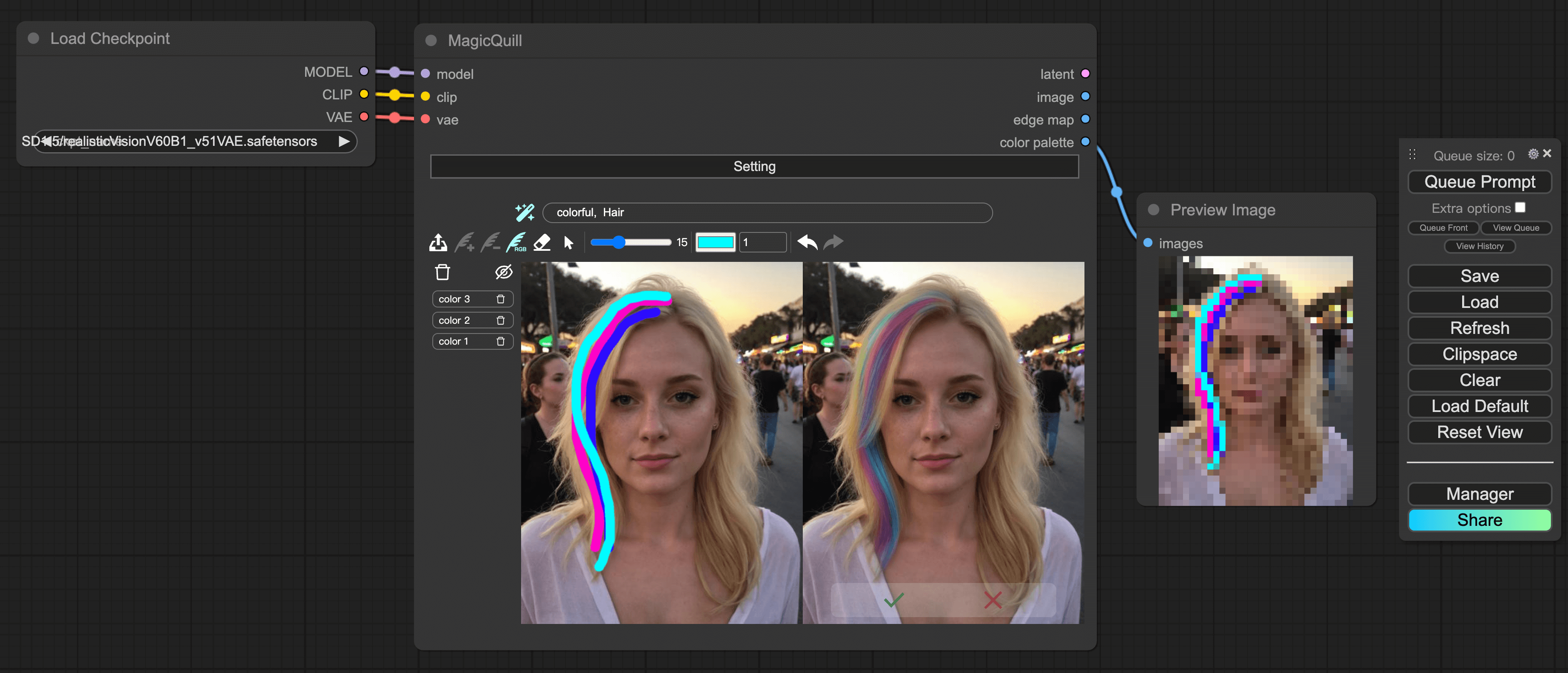}
    \vspace{-0.2cm}
    \caption{\method as a custom node in ComfyUI.}
    \vspace{-0.5cm}
    \label{fig:comfyui}
\end{figure}

Besides Gradio, \method can also be integrated into ComfyUI as a custom node, as shown in Fig.~\ref{fig:comfyui}, with customizable widgets for parameter settings and extensible architecture for future platform integrations.

\noindent\textbf{Usage Scenario:}
To demonstrate the user-friendly workflow of \method, we present an illustrative scenario: A user wants to modify an image of a complete cake, cutting a slice out of it, as shown in Fig~\ref{fig:design}.
The user begins by uploading the image through the toolbar, which provides access to a range of tools (Fig.~\ref{fig:design}-B).
Using the add brush, the user outlines the slice to be cut directly on the canvas (Fig.~\ref{fig:design}-D).
Meanwhile, the \drawguess feature introduced in Sec.~\ref{draw_guess} predicts that the user intends to manipulate a ``cake'' and suggests the relevant prompt automatically in the prompt area (Fig.~\ref{fig:design}-A).
Afterward, the user switches to the subtract brush to fill in the outlined slice, visually marking the area to be removed from the cake.
For additional precision, the eraser tool is available to refine the cut.
Once the adjustments are made, the user generates the image by clicking the Run button (Fig.~\ref{fig:design}-F), which runs the model detailed in Sec.~\ref{Editing_Processor}.

The resulting image appears in the generated image area (Fig.~\ref{fig:design}-E). Users can confirm changes via the tick icon to update the canvas, or click the cross icon to revert modifications. This workflow enables iterative refinement of edits, providing flexible control throughout the process.




\section{Comparison on MagicBrush Benchmark}
\begin{figure}[h]
    \centering
    \vspace{-10pt}
    \includegraphics[width=\linewidth]{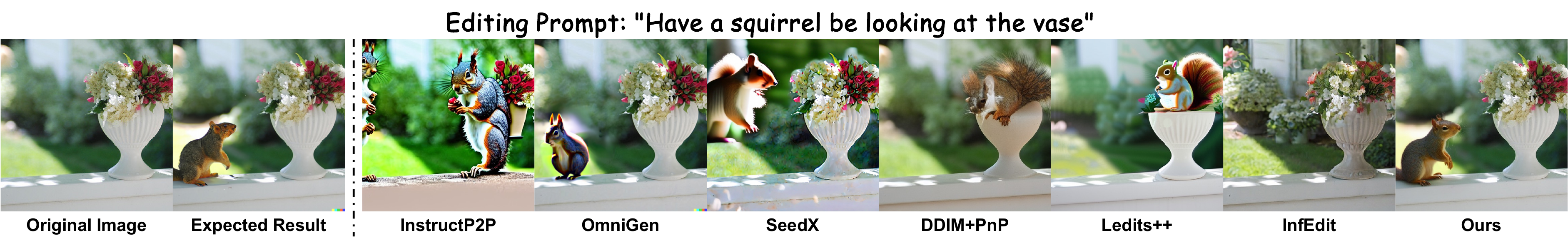}
    \vspace{-20pt}
    \caption{Visual comparison of editing result on MagicBrush}
    \label{fig:magicbrush}
    \vspace{-15pt}
\end{figure}

We evaluate our approach against existing instruction-based image editing methods using the MagicBrush~\cite{magicbrush} benchmark. This benchmark provides high-quality image pairs for both single-round and multi-round editing scenarios, aligning well with our settings. Our comprehensive evaluation against six state-of-the-art instruction-based editing baselines~\cite{instructpix2pix, omnigen, seedx, pnp, ledits++, infedit} demonstrates superior performance in both quantitative metrics and qualitative results, as show in Tab.~\ref{tab:magicbrush} and Fig.~\ref{fig:magicbrush}.

\begin{table}[h]
    \centering 
    \scriptsize
    \vspace{-6pt}
    \caption{Quantitative comparison on MagicBrush benchmark}
    \vspace{-10pt}
    \SetTblrInner{rowsep=0.0pt}
    \SetTblrInner{colsep=1.0pt}
    \begin{tblr}{
        cells={halign=c,valign=m},
        column{1}={halign=l},
        hline{1,3,Z}={1-11}{1.0pt},
        hline{2}={2-11}{},
        vline{2,7}={1-9}{1.0pt},
        cell{1}{1}={r=2}{}, 
        cell{1}{2}={c=5}{},
        cell{1}{7}={c=5}{},
    }
    Method & \SetCell[c=5]{c} Single-Turn & & & & & \SetCell[c=5]{c} Multi-Turn & & & & \\
    & L1 & L2 & CLIP-I & DINO & CLIP-T & L1 & L2 & CLIP-I & DINO & CLIP-T \\
    InstructP2P & 0.115 & 0.039 & 0.849 & 0.741 & 0.265 & 0.141 & 0.050 & 0.817 & 0.678 & 0.270 \\
    OmniGEN & 0.092 & 0.037 & 0.903 & 0.837 & 0.268 & 0.152 & 0.062 & 0.839 & 0.685 & 0.272 \\
    SeedX & 0.187 & 0.090 & 0.857 & 0.747 & 0.268 & 0.258 & 0.130 & 0.785 & 0.564 & 0.269 \\
    DDIM+PNP & 0.100 & 0.026 & 0.858 & 0.785 & 0.278 & 0.131 & 0.039 & 0.824 & 0.709 & 0.281 \\
    Ledits++ & 0.094 & 0.027 & 0.853 & 0.774 & 0.274 & 0.121 & 0.039 & 0.811 & 0.684 & 0.276 \\
    InfEdit & 0.122 & 0.034 & 0.849 & 0.770 & \textbf{0.283} & 0.155 & 0.050 & 0.815 & 0.698 & \textbf{0.288} \\
    Ours & \textbf{0.033} & \textbf{0.011} & \textbf{0.949} & \textbf{0.927} & 0.279 & \textbf{0.035} & \textbf{0.010} & \textbf{0.939} & \textbf{0.913} & 0.284 \\
    \end{tblr}
    \label{tab:magicbrush}
\end{table}

\section{Editing Results under Complex Prompt}
Our system allows users to refine the suggested prompts. Fig.~\ref{fig:complex_prompt} below shows our Editing Processor accurately reflects complex prompts.

\begin{figure}[h]
    \centering
    \vspace{-10pt}
    \includegraphics[width=\linewidth]{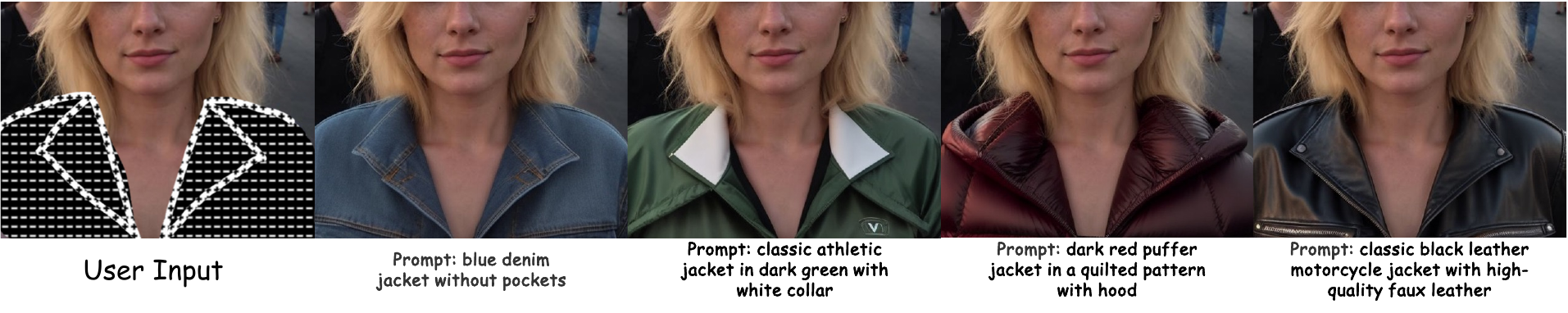}
    \vspace{-20pt}
    \caption{Same image being edited under complex prompts.}
    \vspace{-20pt}
    \label{fig:complex_prompt}
\end{figure}

\section{Control Brush Area Size}
\begin{figure}[h]
    \centering
    \vspace{-10pt}
    \includegraphics[width=0.9\linewidth]{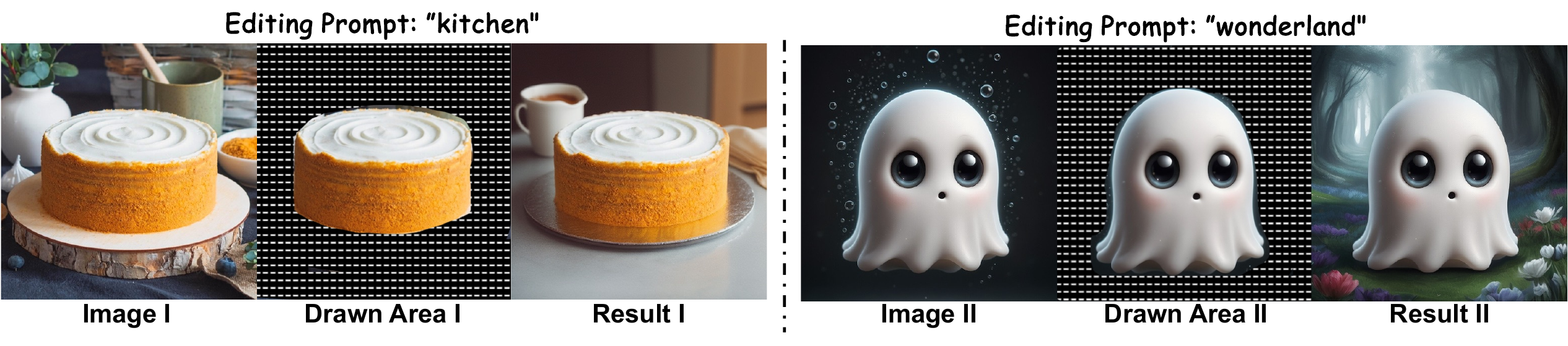}
    \vspace{-10pt}
    \caption{Editing results when user draws large area.}
    \label{fig:large_area}
    \vspace{-20pt}
\end{figure}

Our UI supports brush size adjustment and an eraser to easily correct the control area.
Fig.~\ref{fig:large_area} shows our method generates realistic results for large-scale drawn content.

\section{Failure Case}
\subsection{Failure Case of Editing Processor}
\noindent\textbf{Scribble-Prompt Trade-Off:} We observe quality degradation when user-provided add brush strokes deviate from the semantic content specified in the prompt, a common occurrence among users with limited artistic skills. This creates a fundamental trade-off: strictly following the scribble structure may compromise the generation quality with respect to the text prompt. To address this issue, we propose adjusting the edge control strength.

\begin{figure}[ht]
    \centering  
    \vspace{-0.2cm}
    \begin{minipage}[b]{0.3\linewidth}
        \begin{subfigure}[b]{\linewidth}
            \centering
            \includegraphics[width=\linewidth]{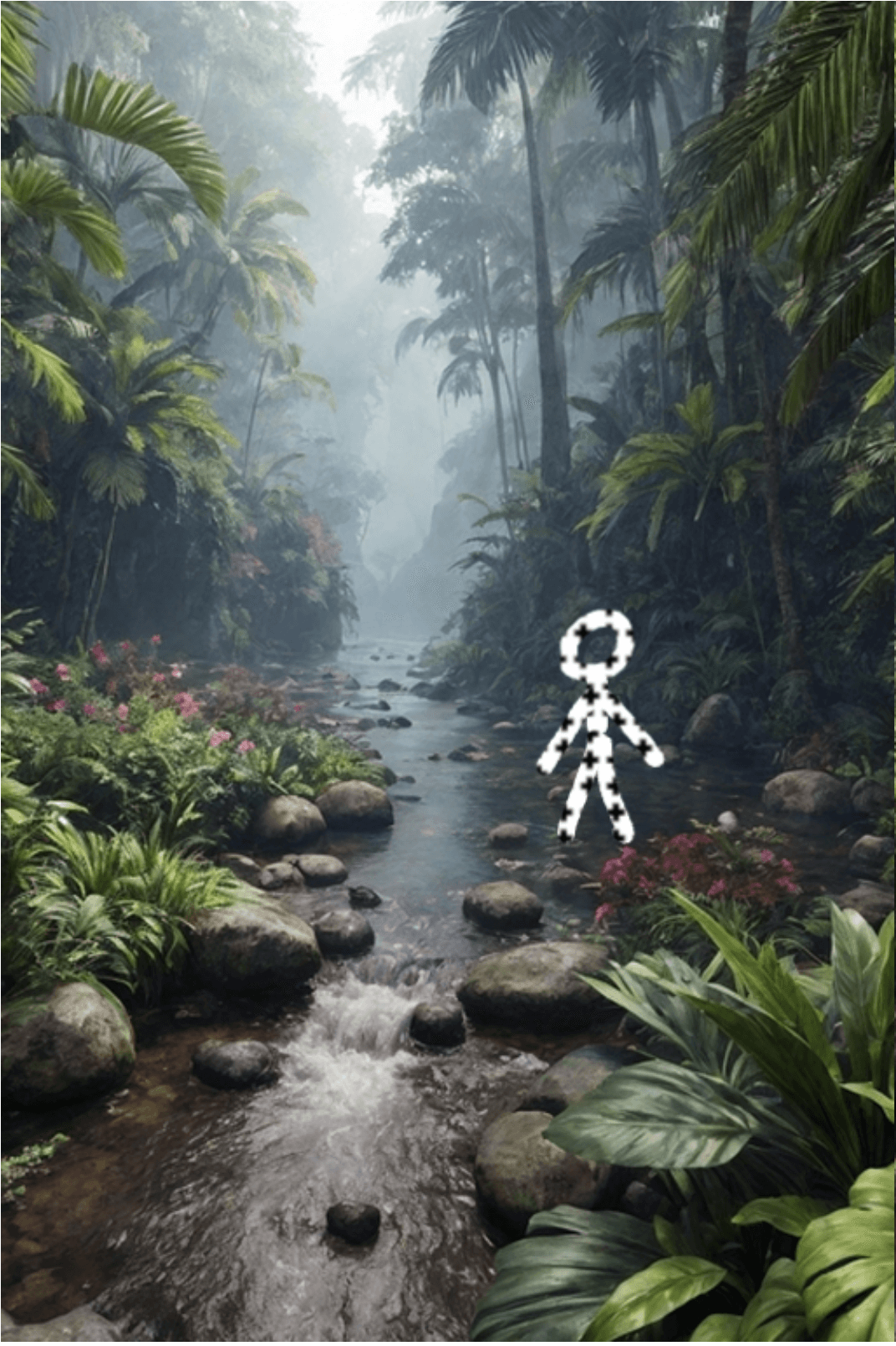}
            \caption{User's Input}
        \end{subfigure}
    \end{minipage}
    \begin{minipage}[b]{0.3\linewidth}
        \begin{subfigure}[b]{\linewidth}
            \centering
            \includegraphics[width=\linewidth]{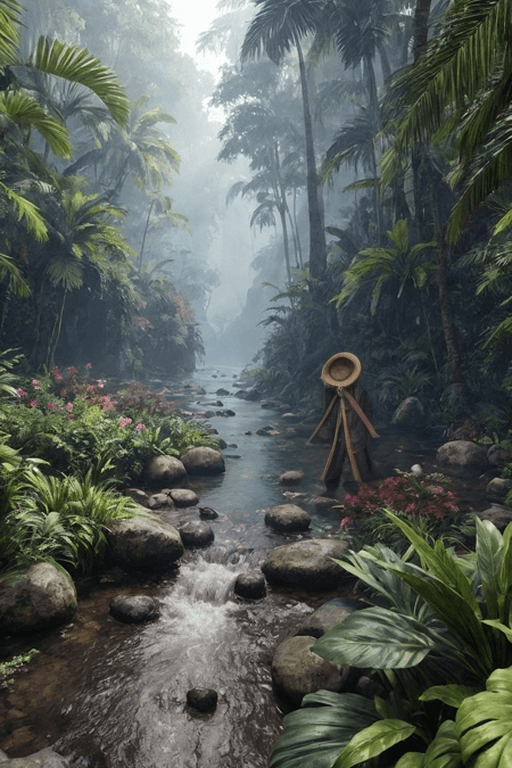}
            \caption{Edge Strength: 0.6}
        \end{subfigure}
    \end{minipage}
    \begin{minipage}[b]{0.3\linewidth}
        \begin{subfigure}[b]{\linewidth}
            \centering
            \includegraphics[width=\linewidth]{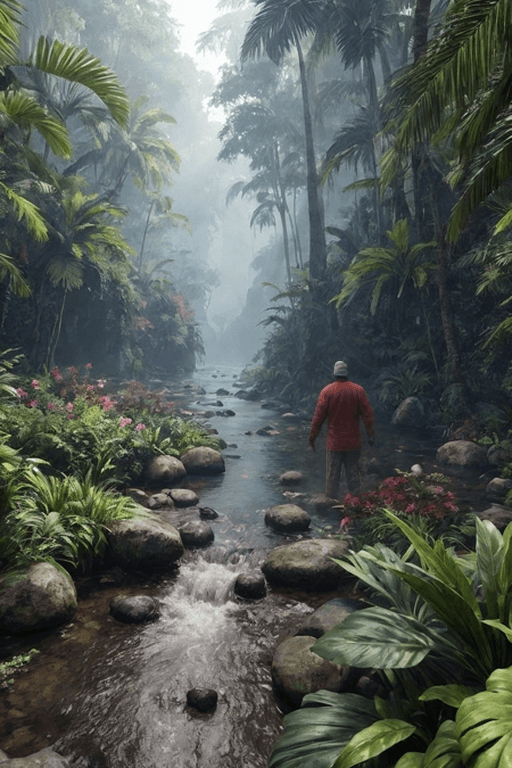}
            \caption{Edge Strength: 0.2}
        \end{subfigure}
    \end{minipage}
    \vspace{-0.2cm}
    \caption{Illustration of the Scribble-Prompt Trade-Off. Given user-provided brush strokes (a) with the text prompt ``man'', we show generation results with different edge control strengths: (b) with edge control of $0.6$ and (c) with edge control of $0.2$.}
    \vspace{-0.3cm}
    \label{fig:failure_case1}
\end{figure}

As demonstrated in Fig.~\ref{fig:failure_case1}, when presented with an oversimplified sketch that substantially deviates from the prompt ``man'', a high edge strength of $0.6$ produces results that, while faithful to the sketch, appear inharmonious. By reducing the edge strength to $0.2$, we achieve notably improved generation quality.

\noindent\textbf{Colorization-Details Trade-Off:} We observe a trade-off between colorization accuracy and detail preservation. Since our conditional image inpainting pipeline relies on downsampled color blocks and CNN-extracted edge maps as input, structrual details in the edited regions may be compromised during the generation process. 

\begin{figure}[ht]
    \centering  
    \vspace{-0.2cm}
    \begin{minipage}[b]{0.3\linewidth}
        \begin{subfigure}[b]{\linewidth}
            \centering
            \includegraphics[width=\linewidth]{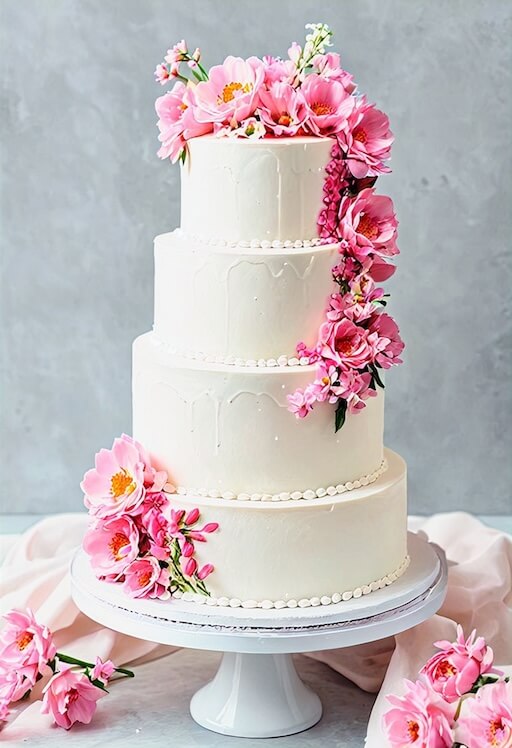}
            \caption{Original Image}
        \end{subfigure}
    \end{minipage}
    \begin{minipage}[b]{0.3\linewidth}
        \begin{subfigure}[b]{\linewidth}
            \centering
            \includegraphics[width=\linewidth]{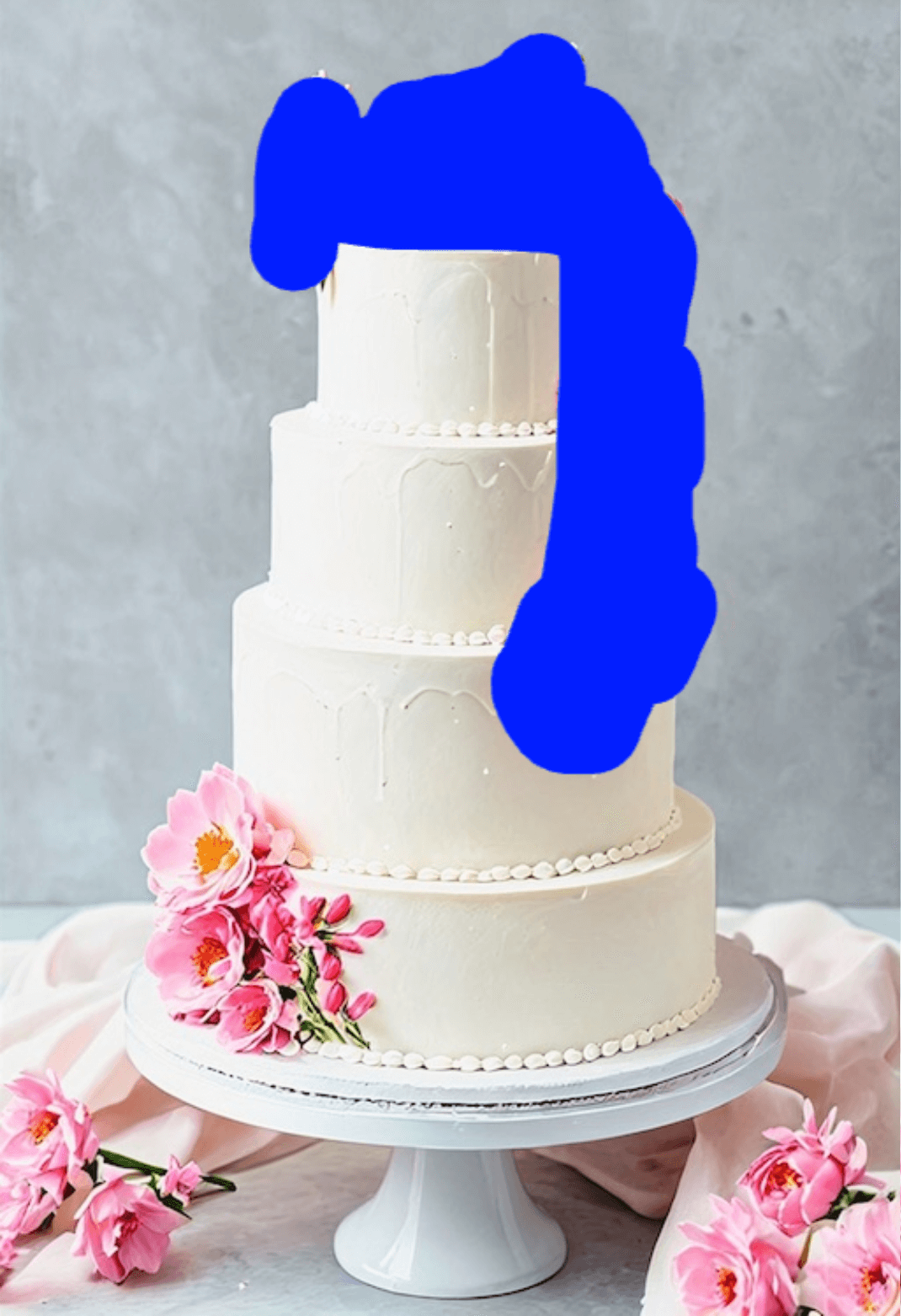}
            \caption{Color brush, $\alpha$ 1.0}
        \end{subfigure}
    \end{minipage}
    \begin{minipage}[b]{0.3\linewidth}
        \begin{subfigure}[b]{\linewidth}
            \centering
            \includegraphics[width=\linewidth]{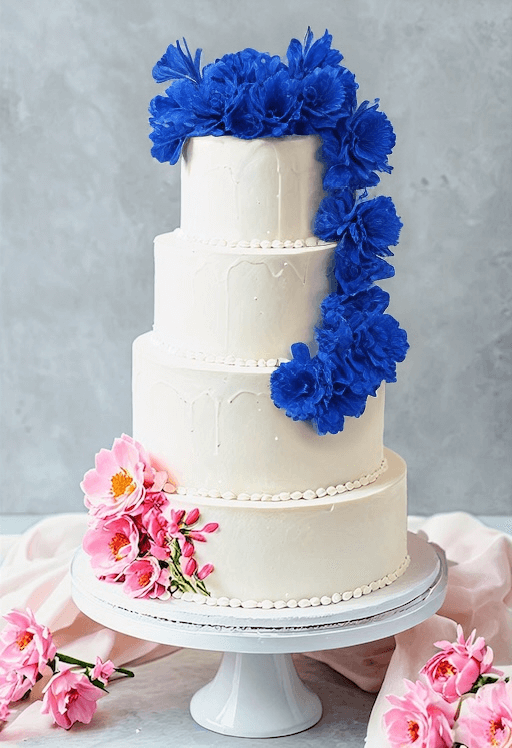}
            \caption{Result for $\alpha$ 1.0}
        \end{subfigure}
    \end{minipage}
    
    \begin{minipage}[b]{0.3\linewidth}
        \begin{subfigure}[b]{\linewidth}
            \centering
            \includegraphics[width=\linewidth]{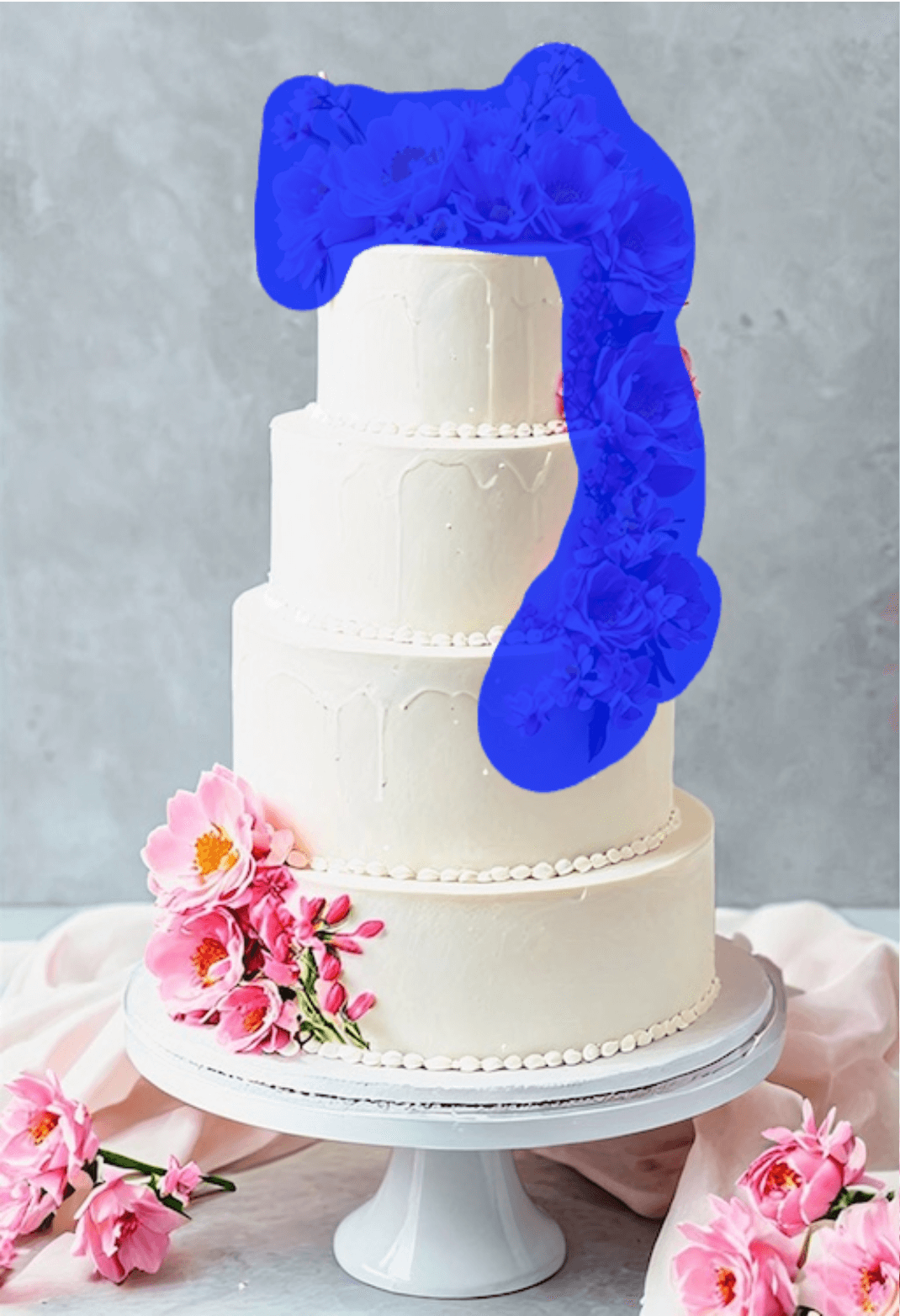}
            \caption{Color brush, $\alpha$ 0.8}
        \end{subfigure}
    \end{minipage}
    \hspace{0.3cm} 
    \begin{minipage}[b]{0.3\linewidth}
        \begin{subfigure}[b]{\linewidth}
            \centering
            \includegraphics[width=\linewidth]{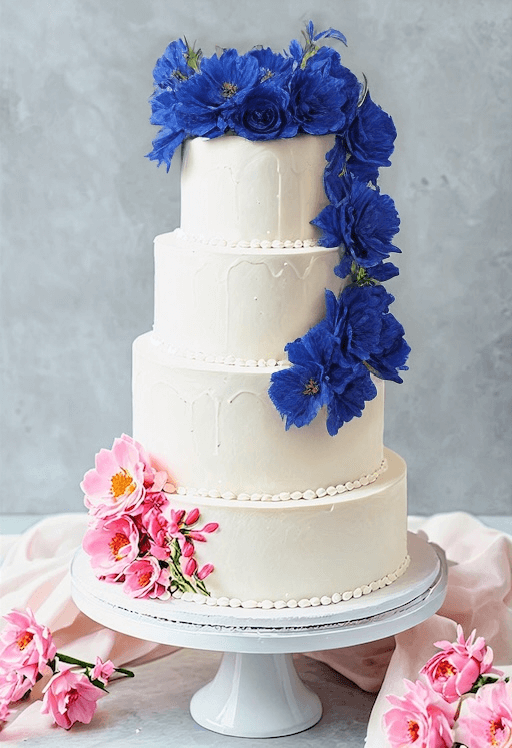}
            \caption{Result for $\alpha$ 0.8}
        \end{subfigure}
    \end{minipage}
    \vspace{-0.2cm}
    \caption{Illustration of the Colorization-Detail Trade-Off. Results of color brush strokes with different alpha values: (b, c) using alpha value $1.0$, and (d, e) using alpha value $0.8$, where the latter better preserves more structural details of the original image.}
    \label{fig:failure_case2}
\end{figure}

As illustrated in Fig.~\ref{fig:failure_case2}, this limitation can be partially mitigated by reducing the alpha value of the color brush trokes, which preserves more information from the original image when downsampled to color blocks. Future work could explore using grayscale images as the control condition to achieve colorization while maintaining fine-grained structural details.

\subsection{Failure Case of Painting Assistor}

\begin{figure}[t]
    \centering  
    \vspace{-0.2cm}
    \begin{minipage}[b]{0.32\linewidth}
        \begin{subfigure}[b]{\linewidth}
            \centering
            \includegraphics[width=\linewidth]{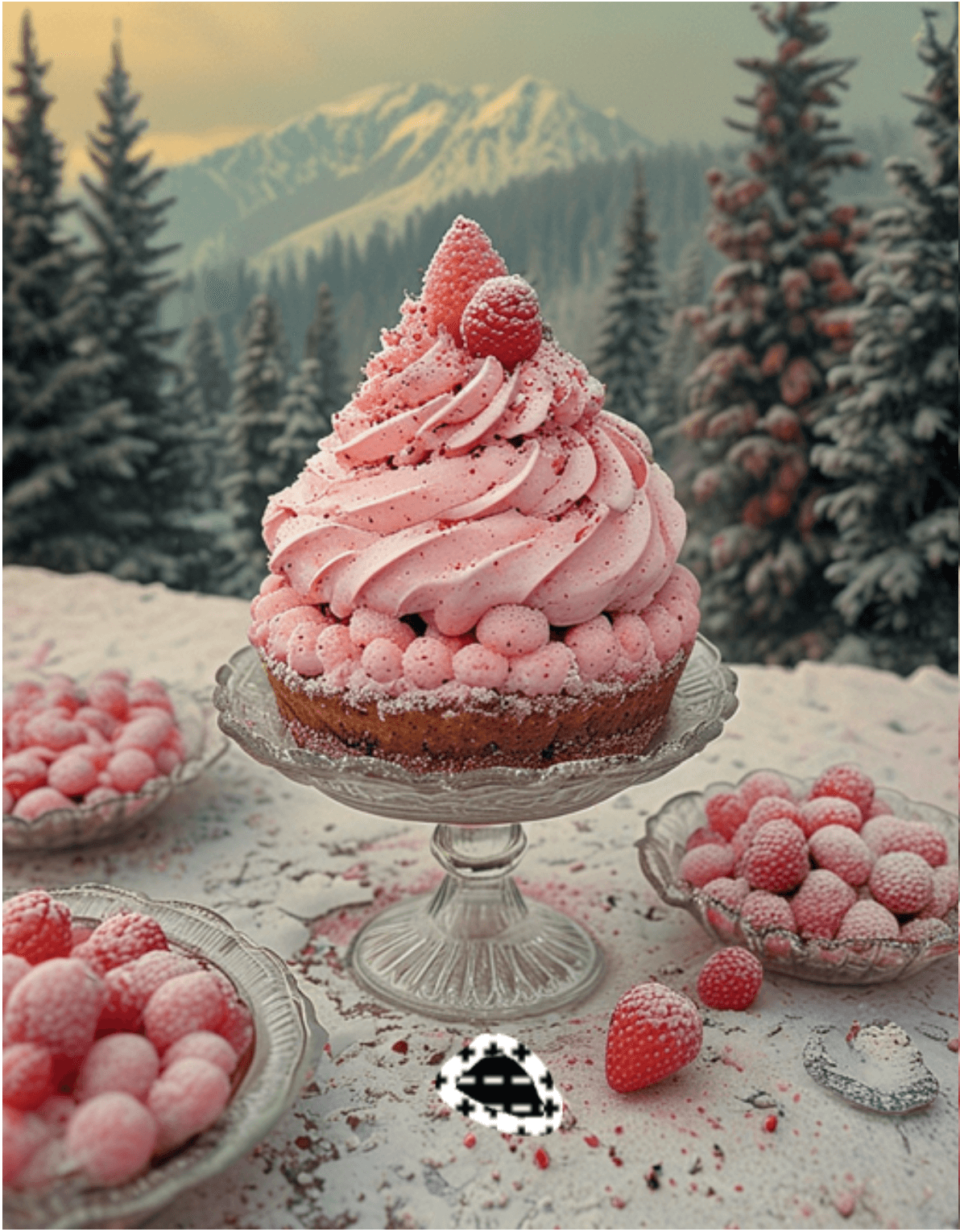}
            \caption{User's Input}
        \end{subfigure}
    \end{minipage}
    \begin{minipage}[b]{0.32\linewidth}
        \begin{subfigure}[b]{\linewidth}
            \centering
            \includegraphics[width=\linewidth]{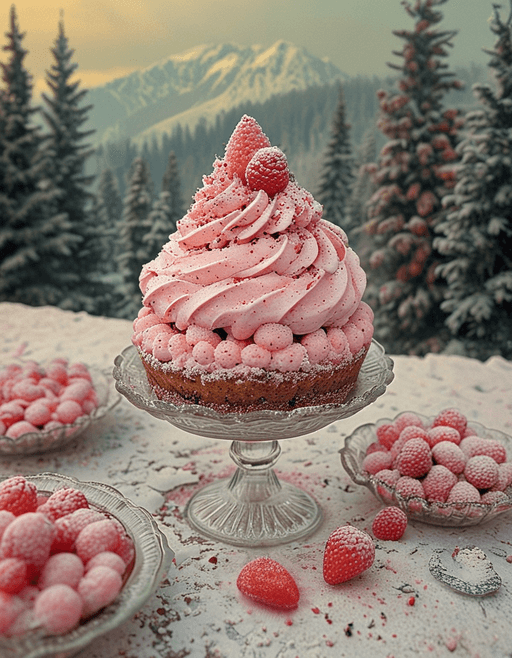}
            \caption{Prompt: Candy}
        \end{subfigure}
    \end{minipage}
    \begin{minipage}[b]{0.32\linewidth}
        \begin{subfigure}[b]{\linewidth}
            \centering
            \includegraphics[width=\linewidth]{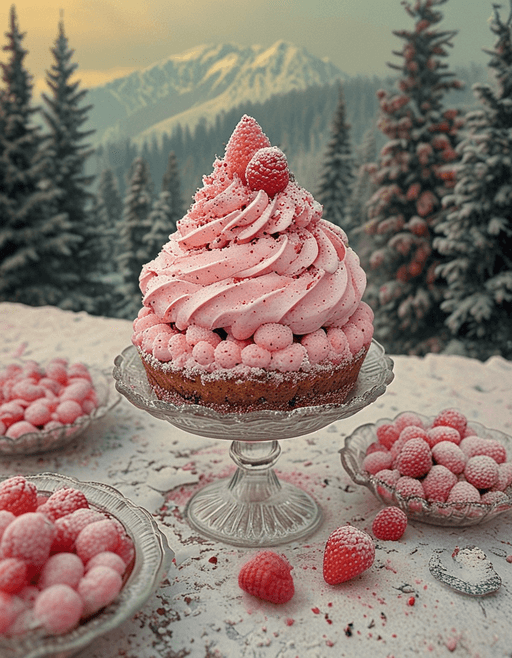}
            \caption{Prompt: Raspberry}
        \end{subfigure}
    \end{minipage}
    \vspace{-0.2cm}
    \caption{Demonstration of semantic ambiguity in sketch interpretation. (A) User's sketch intended to represent a raspberry; (B) Our Draw\&Guess model incorrectly interprets the sketch as candy, leading to a misaligned generation; (C) The expected generation result with correct raspberry interpretation.}
    \label{fig:failure_case3}
    \vspace{-5pt}
\end{figure}

\noindent\textbf{Ambiguity of the Sketch:} Our system enables users to express their editing intentions through brush strokes, which are then interpreted by the Painting Assistor via Draw\&Guess. However, this approach faces inherent limitations due to the ambiguous nature of user-provided sketches. For instance, a simple circular sketch could represent various objects like strawberry, raspberry, or candy, making it challenging for the model to accurately infer the user's intended modification, as shown in Fig.~\ref{fig:failure_case3}. 

This ambiguity in sketch interpretation can lead to misaligned generations that deviate from the user's expectations. Fortunately, our user study reveals that participants were generally understanding of such interpretation errors and considered the model's predictions to be reasonable attempts at disambiguating their sketches.

\section{Generalizability of Editing Processor}
\begin{figure}[ht]
    \centering  
    \vspace{-0.2cm}
    \begin{minipage}[b]{0.32\linewidth}
        \begin{subfigure}[b]{\linewidth}
            \centering
            \includegraphics[width=\linewidth]{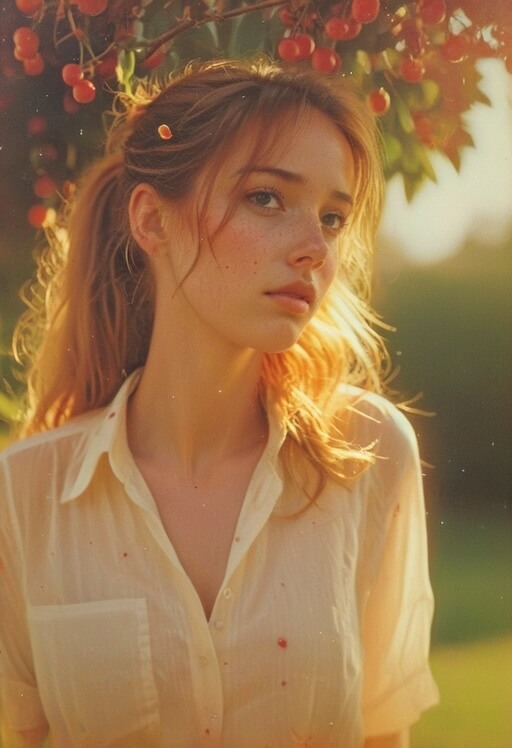}
        \end{subfigure}
    \end{minipage}
    \begin{minipage}[b]{0.32\linewidth}
        \begin{subfigure}[b]{\linewidth}
            \centering
            \includegraphics[width=\linewidth]{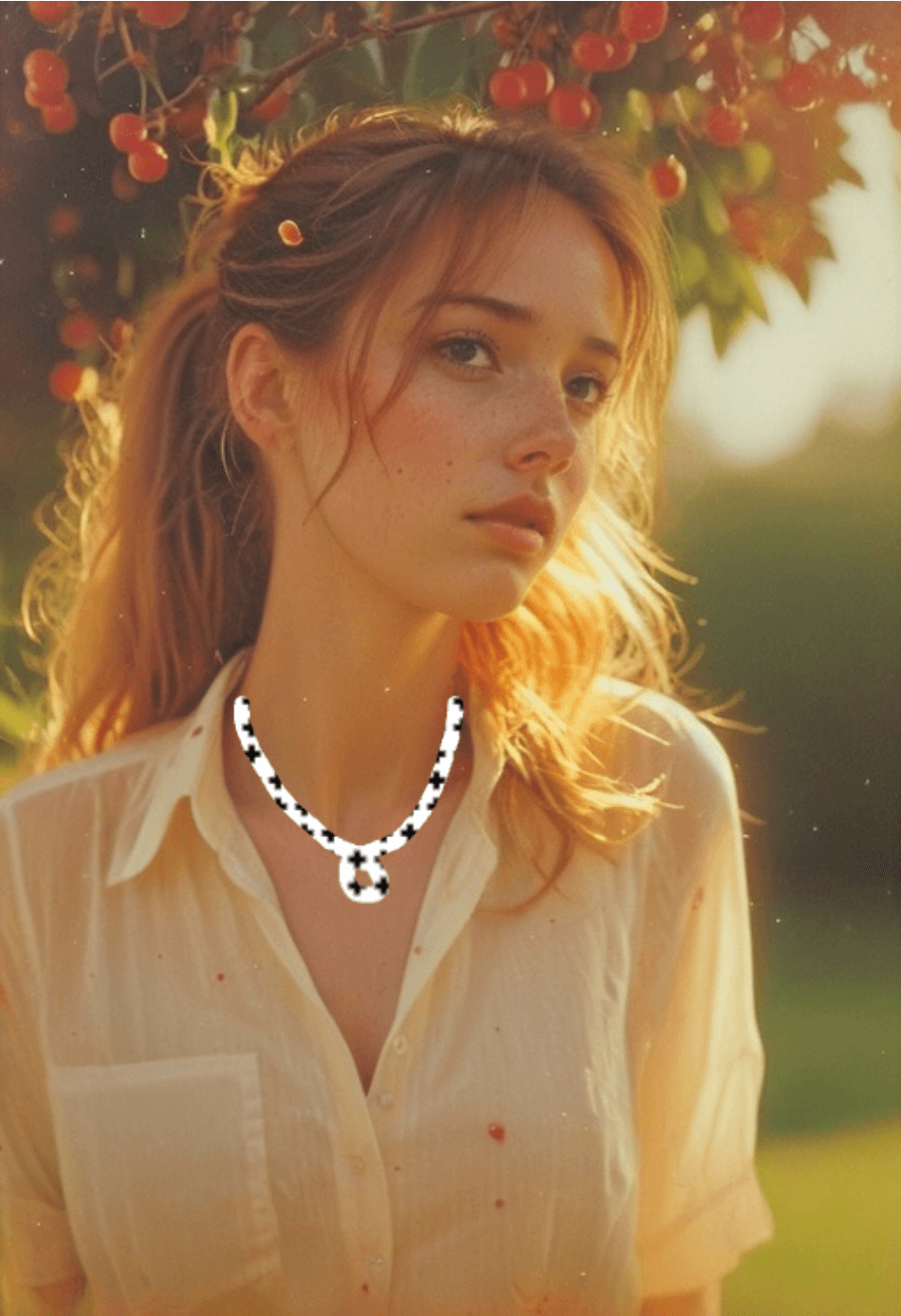}
        \end{subfigure}
    \end{minipage}
    \begin{minipage}[b]{0.32\linewidth}
        \begin{subfigure}[b]{\linewidth}
            \centering
            \includegraphics[width=\linewidth]{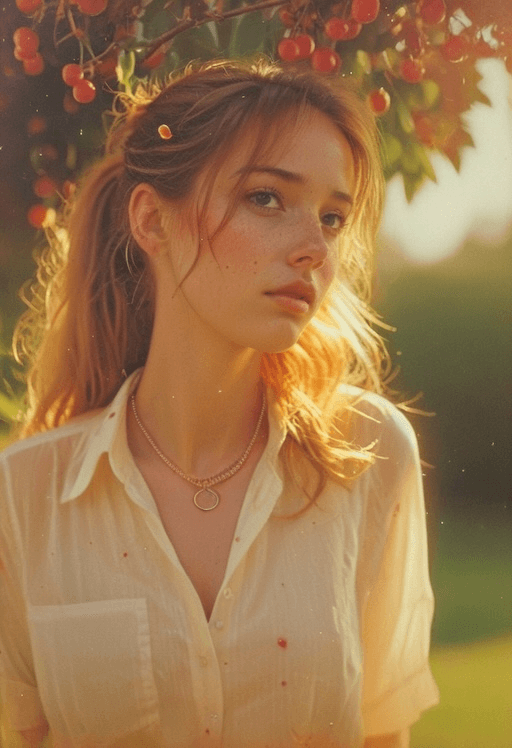}
        \end{subfigure}
    \end{minipage}
    \begin{minipage}[b]{0.32\linewidth}
        \begin{subfigure}[b]{\linewidth}
            \centering
            \includegraphics[width=\linewidth]{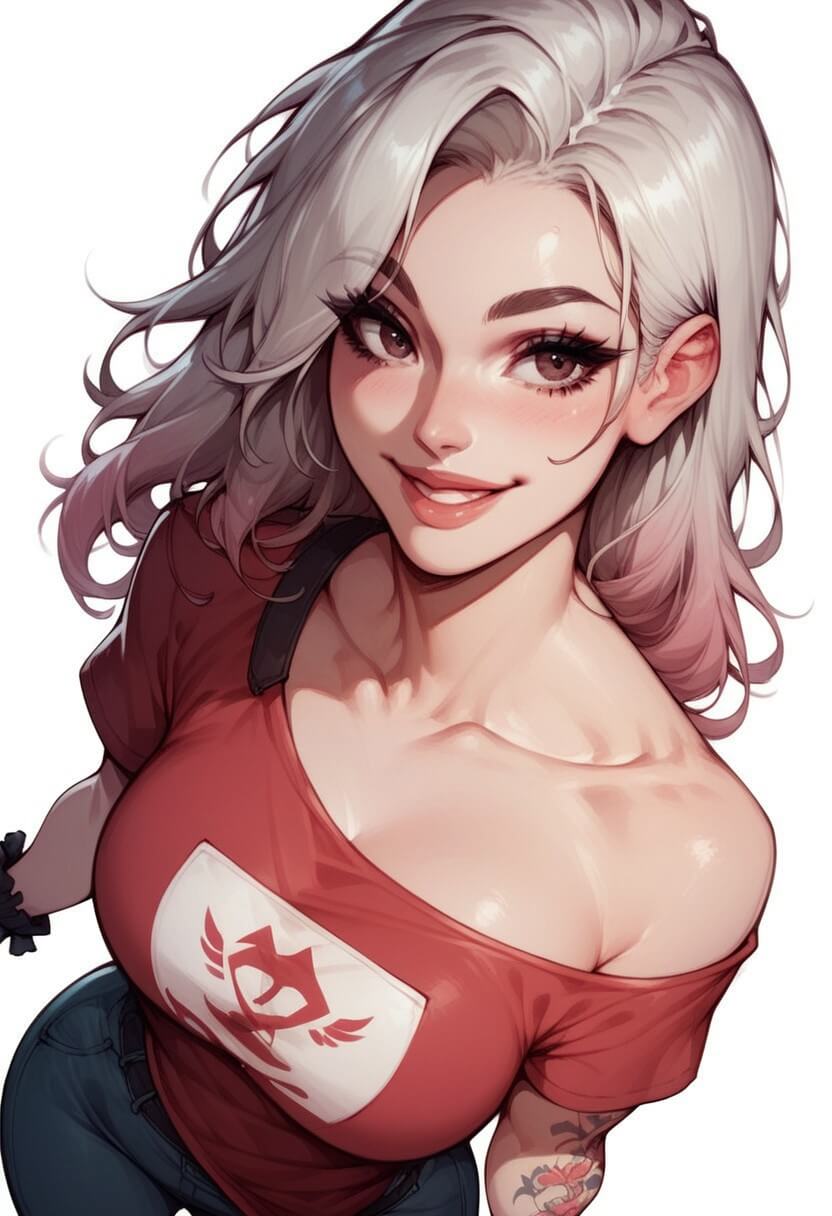}
        \end{subfigure}
    \end{minipage}
    \begin{minipage}[b]{0.32\linewidth}
        \begin{subfigure}[b]{\linewidth}
            \centering
            \includegraphics[width=\linewidth]{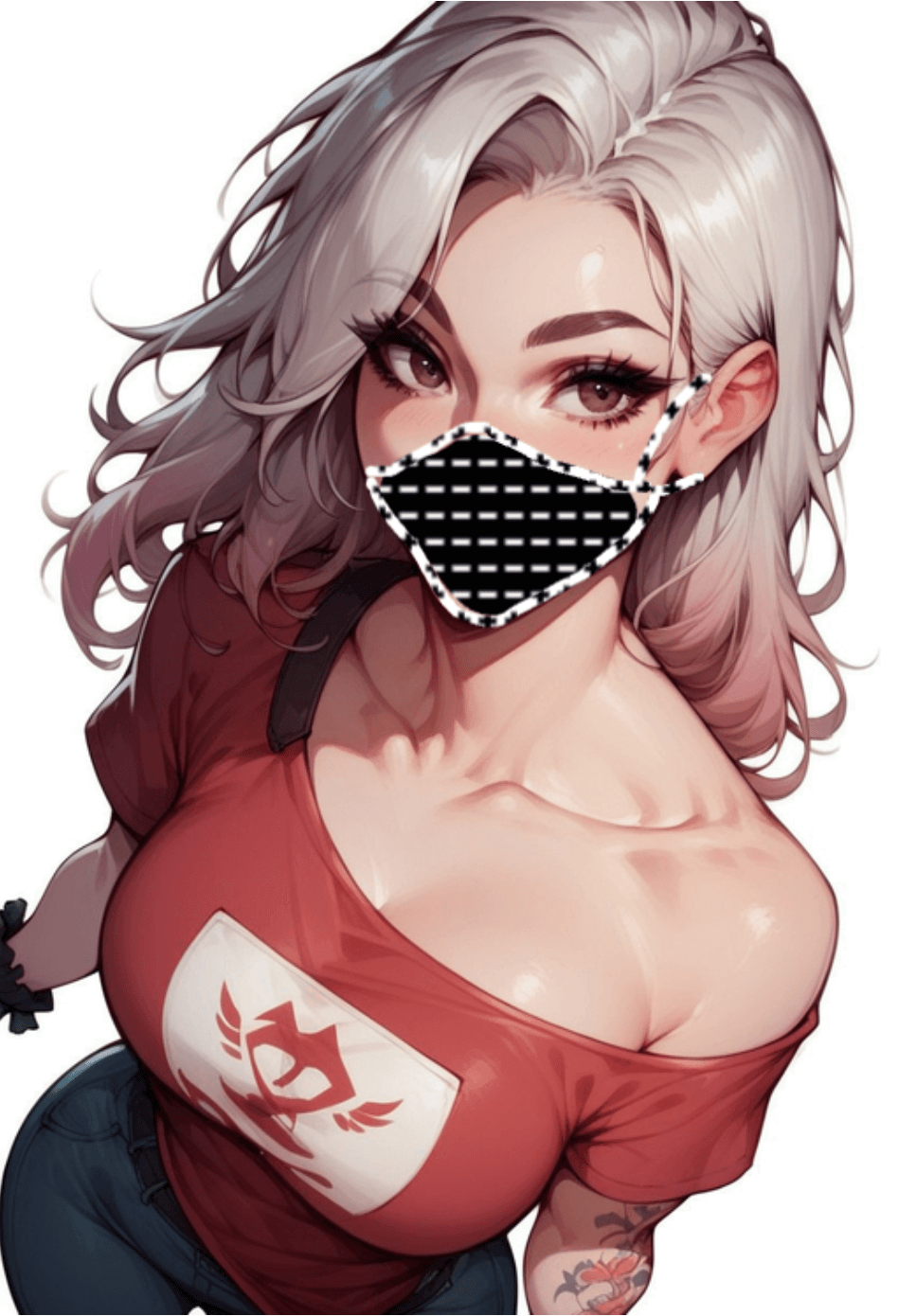}
        \end{subfigure}
    \end{minipage}
    \begin{minipage}[b]{0.32\linewidth}
        \begin{subfigure}[b]{\linewidth}
            \centering
            \includegraphics[width=\linewidth]{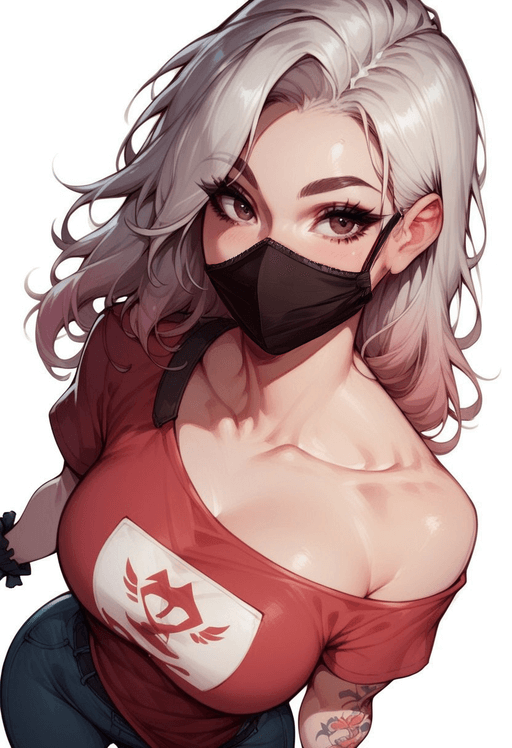}
        \end{subfigure}
    \end{minipage}  
    \begin{minipage}[b]{0.32\linewidth}
        \begin{subfigure}[b]{\linewidth}
            \centering
            \includegraphics[width=\linewidth]{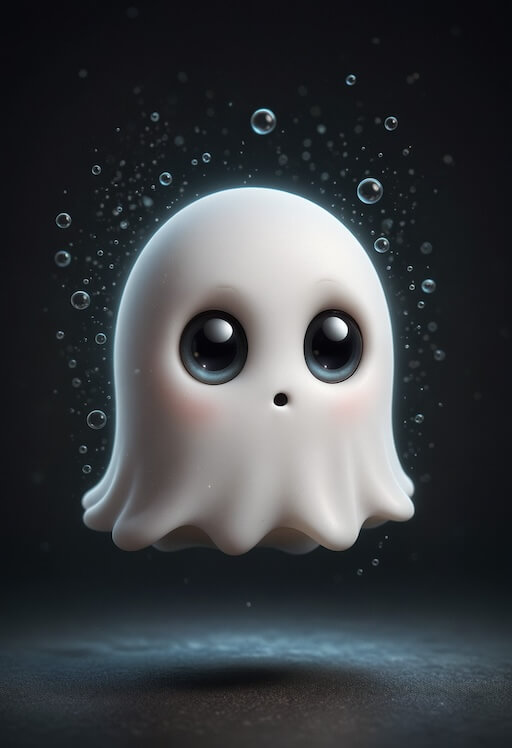}
            \caption{Original Image}
        \end{subfigure}
    \end{minipage}
    \begin{minipage}[b]{0.32\linewidth}
        \begin{subfigure}[b]{\linewidth}
            \centering
            \includegraphics[width=\linewidth]{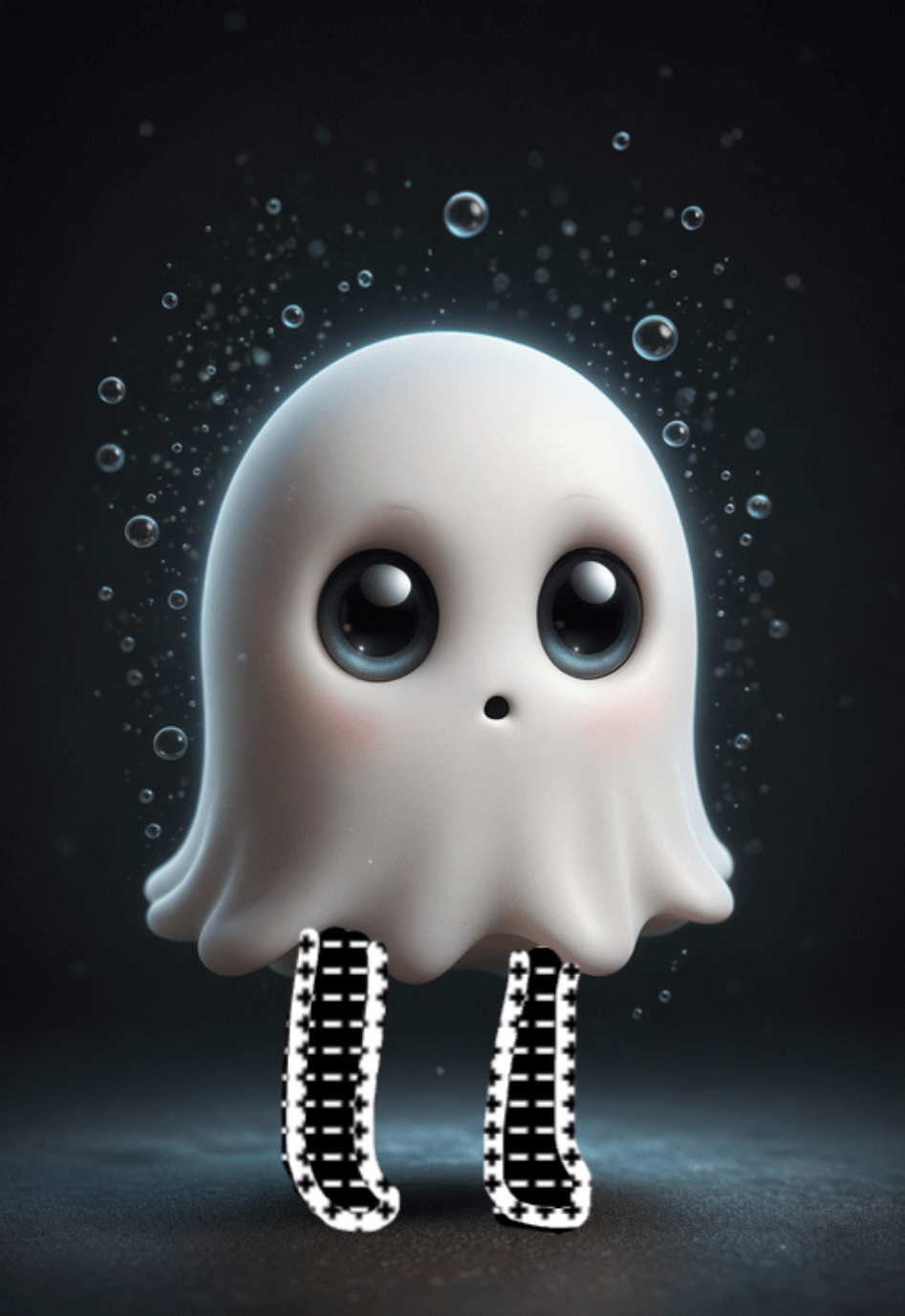}
            \caption{User's Input}
        \end{subfigure}
    \end{minipage}
    \begin{minipage}[b]{0.32\linewidth}
        \begin{subfigure}[b]{\linewidth}
            \centering
            \includegraphics[width=\linewidth]{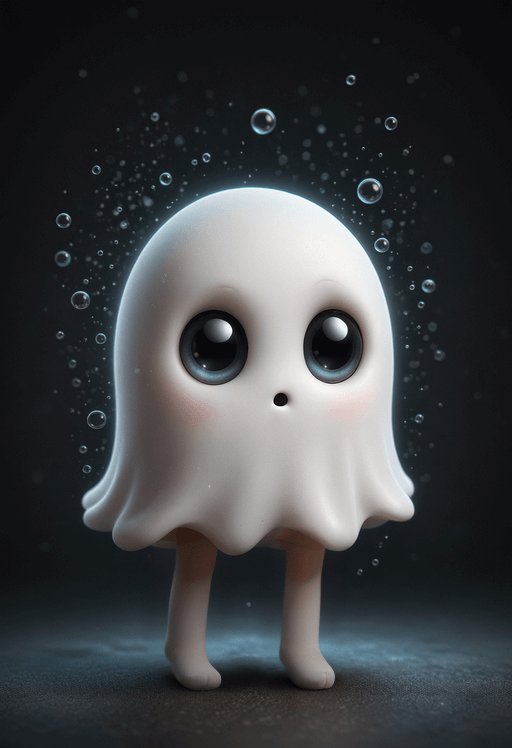}
            \caption{Editing Result}
        \end{subfigure}
    \end{minipage}
    \vspace{-0.2cm}
    \caption{Demonstration of our method's generalization capability across different fine-tuned Stable Diffusion models. Results shown using RealisticVision (top row), GhostMix (middle row), and DreamShaper (bottom row) as base models, all achieving consistent editing performance.}
    \vspace{-5pt}
    \label{fig:generalization}
\end{figure}

Our Editing Processor demonstrates generalization capabilities across various fine-tuned Stable Diffusion v1.5 models. Since both the inpainting and control branches preserve the weights of pre-trained diffusion models, our method seamlessly integrates with any community fine-tuned model as a plug-and-play component. We validate this versatility by testing on several popular fine-tuned models including RealisticVision, GhostMix, and DreamShaper, achieving consistent editing performance while inheriting the unique stylistic characteristics of each model, as shown in Fig.~\ref{fig:generalization}. This compatibility highlights the practical value of our Editing Processor, as users can leverage their preferred fine-tuned models or LoRA~\cite{lora} weight while maintaining the editing capabilities provided by our framework.

\section{In-Context Editing Intent Interpretation}
\begin{figure}[t]
    \centering  
    \vspace{-0.2cm}
    \begin{minipage}[b]{0.32\linewidth}
        \begin{subfigure}[b]{\linewidth}
            \centering
            \includegraphics[width=\linewidth]{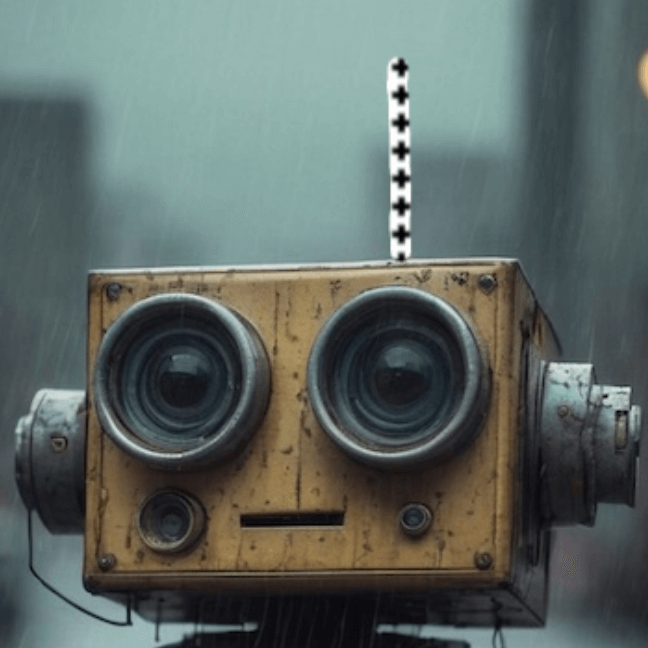}
            \caption{Guess: Antenna}
        \end{subfigure}
    \end{minipage}
    \begin{minipage}[b]{0.32\linewidth}
        \begin{subfigure}[b]{\linewidth}
            \centering
            \includegraphics[width=\linewidth]{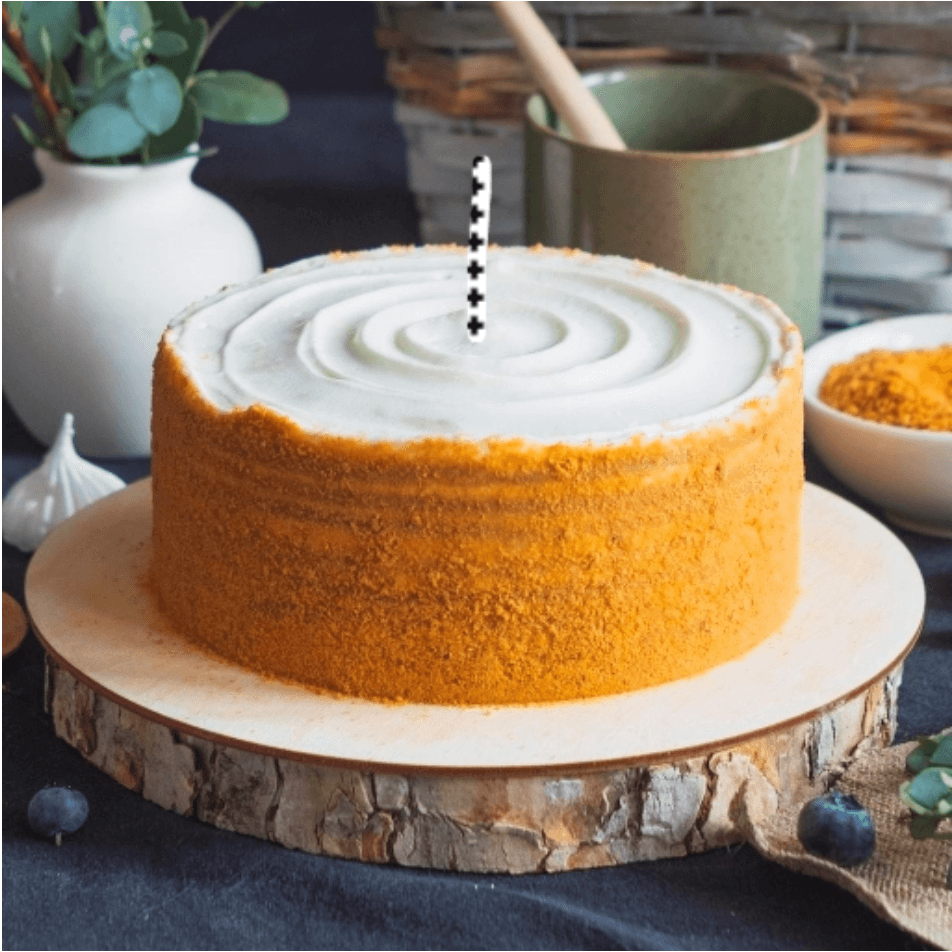}
            \caption{Guess: Candle}
        \end{subfigure}
    \end{minipage}
    \begin{minipage}[b]{0.32\linewidth}
        \begin{subfigure}[b]{\linewidth}
            \centering
            \includegraphics[width=\linewidth]{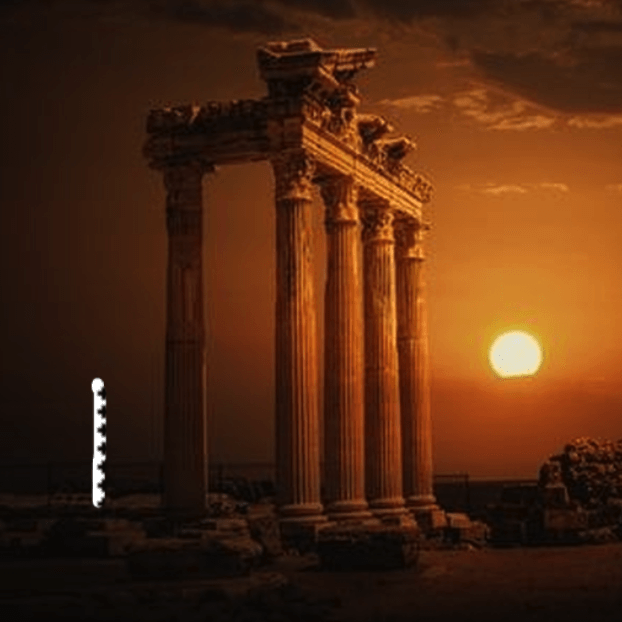}
            \caption{Guess: Column}
        \end{subfigure}
    \end{minipage}
    \vspace{-0.2cm}
    \caption{Examples of context-aware editing intention interpretation. The MLLM interprets the same vertical line sketch differently based on surrounding context: (a) as an antenna on a robot's head, (b) as a candle on a birthday cake, and (c) as a column among ancient ruins.}
    \label{fig:context}
    \vspace{-0.5cm}
\end{figure}

The MLLM in Painting Assistor, fine-tuned on our own constructed dataset in Sec.~\ref{draw_guess}, demonstrates sophisticated in-context reasoning capabilities for editing intent interpretation. The model effectively leverages contextual visual information to interpret user brush strokes based on their surrounding environment. For instance, a simple vertical line is interpreted differently based on its context: as a candle on a cake, a column on ruins, or an antenna on a robot, as illustrated in Fig.~\ref{fig:context}. These context-aware interpretations validate the effectiveness of our dataset construction approach and highlight the model's ability to incorporate environmental cues in its reasoning process.

\begin{figure*}[h]
    \centering
    \vspace{-0.2cm}
    \begin{subfigure}[t]{\linewidth}
        \centering
        \includegraphics[width=\linewidth]{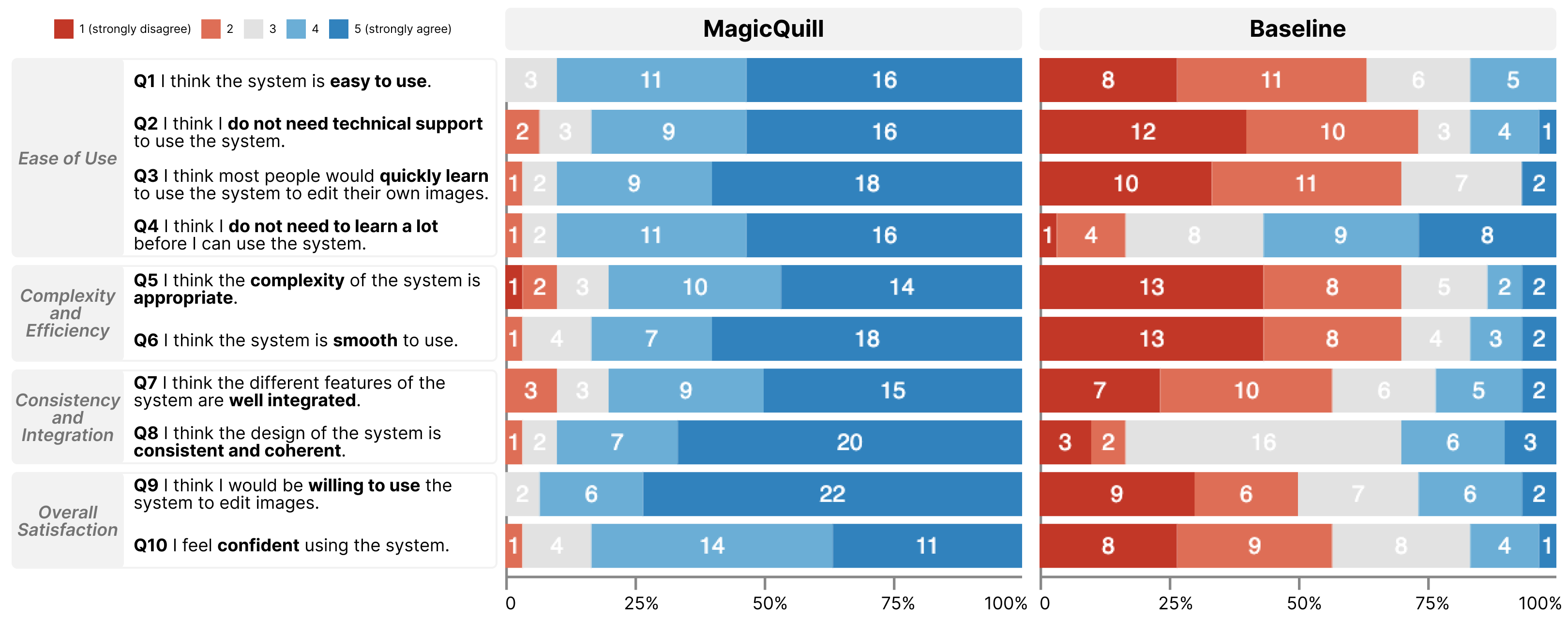}
    \end{subfigure}
    \vspace{-13pt}
    \caption{The questionnaire and user ratings comparing \method to the baseline system ($1$=strongly disagree, $5$=strongly agree).}
    \label{fig:evaluation-likert}
\end{figure*}

\begin{figure}[h]
    \centering
    \includegraphics[width=\linewidth]{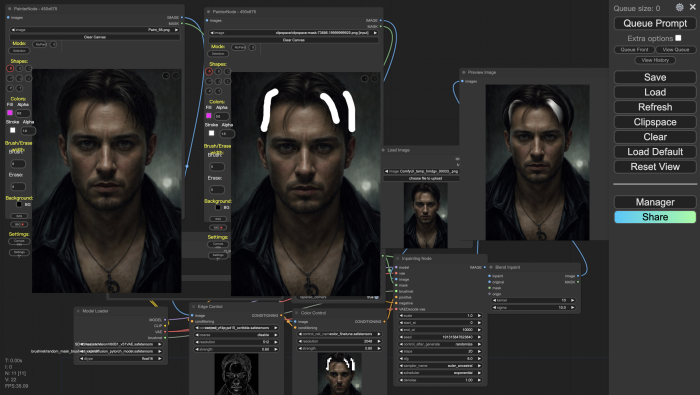}
    \vspace{-0.5cm}
    \caption{The baseline system implemented in ComfyUI.}
    \vspace{-0.3cm}
    \label{fig:evaluation-baseline}
\end{figure}

\section{User Study Details and Questionnaires}

To assess the effectiveness and usability of the Painting Assistor and Idea Collector, we recruited $30$ participants from diverse backgrounds, including postgraduate students, artists, and computer vision researchers. All had image editing experience, with varying skill levels, providing a realistic range of user proficiency.

To control for learning effects, we randomly divided participants into two groups: Group A used \method before the baseline (Fig.~\ref{fig:evaluation-baseline}), while Group B followed the reverse order. Each participant completed a comprehensive evaluation consisting of $10$ questions per system, modified from the System Usability Scale (SUS)~\cite{SUS}, spanning four key categories: \textit{Complexity and Efficiency}, \textit{Consistency and Integration}, \textit{Ease of Use}, and \textit{Overall Satisfaction}. .
The detailed evaluation results are presented in Fig.~\ref{fig:evaluation-likert}.
Additionally, participants responded to $2$ specific questions addressing the Painting Assistor's accuracy and efficiency.

In the \textit{Ease of Use} category, all participants rated the easiness (Q1) with a score of $3$ or above, and most reported learning our system more quickly (Q3, Q4) and independently (Q2) compared to the baseline. These findings indicate a lower barrier to entry for creative tasks with our system.
For \textit{Complexity and Efficiency}, $80\%$ of participants found our system’s complexity appropriate (Q5), contrasting with perceptions of excessive complexity in the baseline. Additionally, $83.3\%$ felt our system was smooth to use (Q6), suggesting that our design lowered cognitive load and supported efficient task completion.
In \textit{Consistency and Integration}, $80\%$ agreed on effective feature integration (Q7), and $90\%$ of participants agreed that our system was consistent and coherent (Q8). This feedback suggests our system provided a cohesive and intuitive user experience.
Lastly, for \textit{Overall Satisfaction}, $93\%$ expressed willingness to use our system (Q9), and $83\%$ reported confidence in using it (Q10). This high satisfaction rate reflects positive user reception and highlights the system’s overall effectiveness in meeting user expectations in editing. 

\begin{figure*}[h]
    \centering  
    \begin{minipage}[b]{0.3\linewidth}
        \begin{subfigure}[b]{\linewidth}
            \centering
            \includegraphics[height=3.8cm]{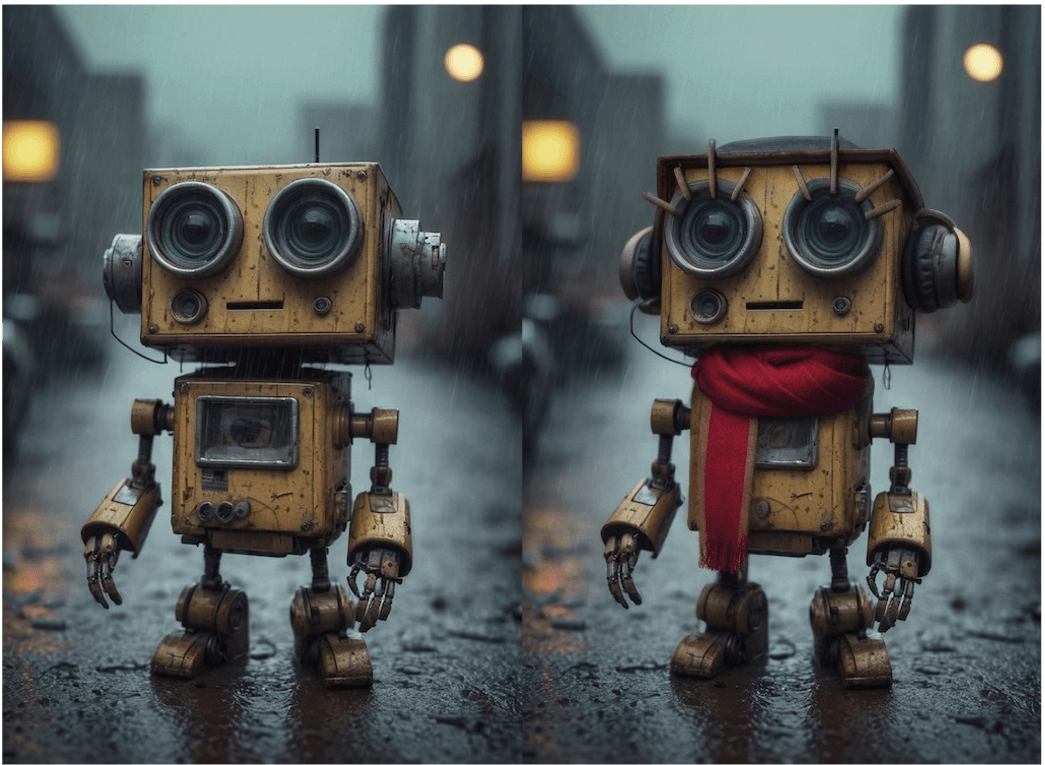}
        \end{subfigure}
    \end{minipage}
    \hspace{0.2cm}
    \begin{minipage}[b]{0.3\linewidth}
        \begin{subfigure}[b]{\linewidth}
            \centering
            \includegraphics[height=3.8cm]{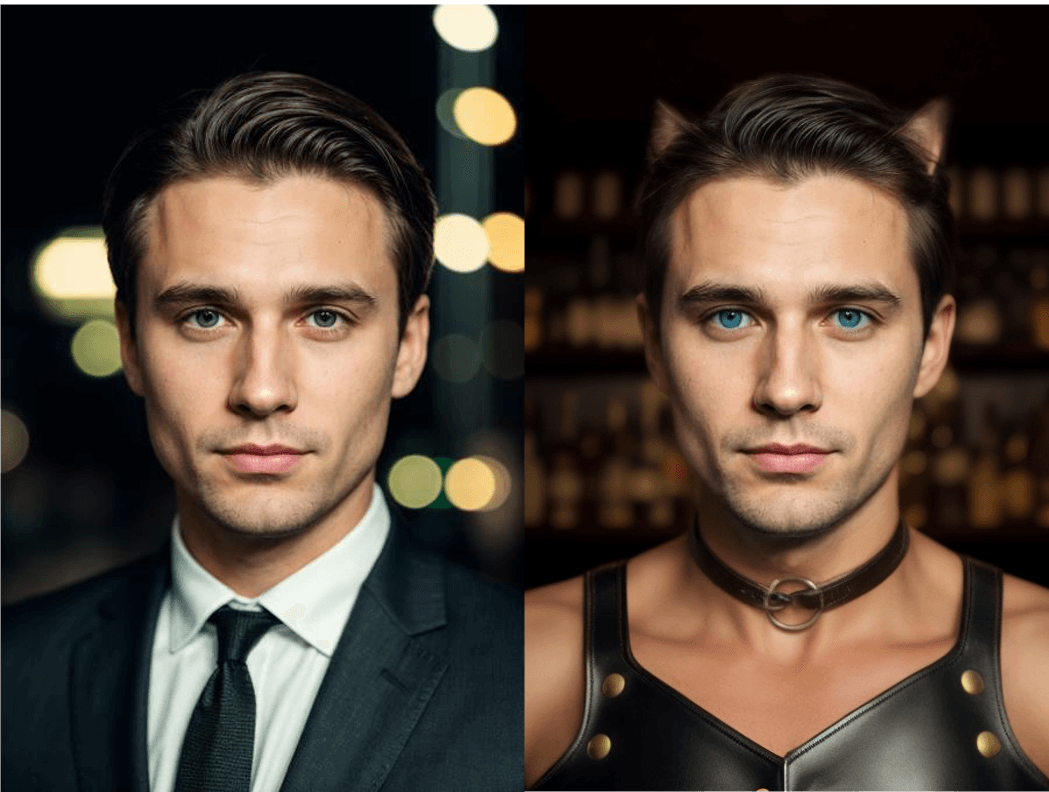}
        \end{subfigure}
    \end{minipage}
    \hspace{0.2cm}
    \begin{minipage}[b]{0.3\linewidth}
        \begin{subfigure}[b]{\linewidth}
            \centering
            \includegraphics[height=3.8cm]{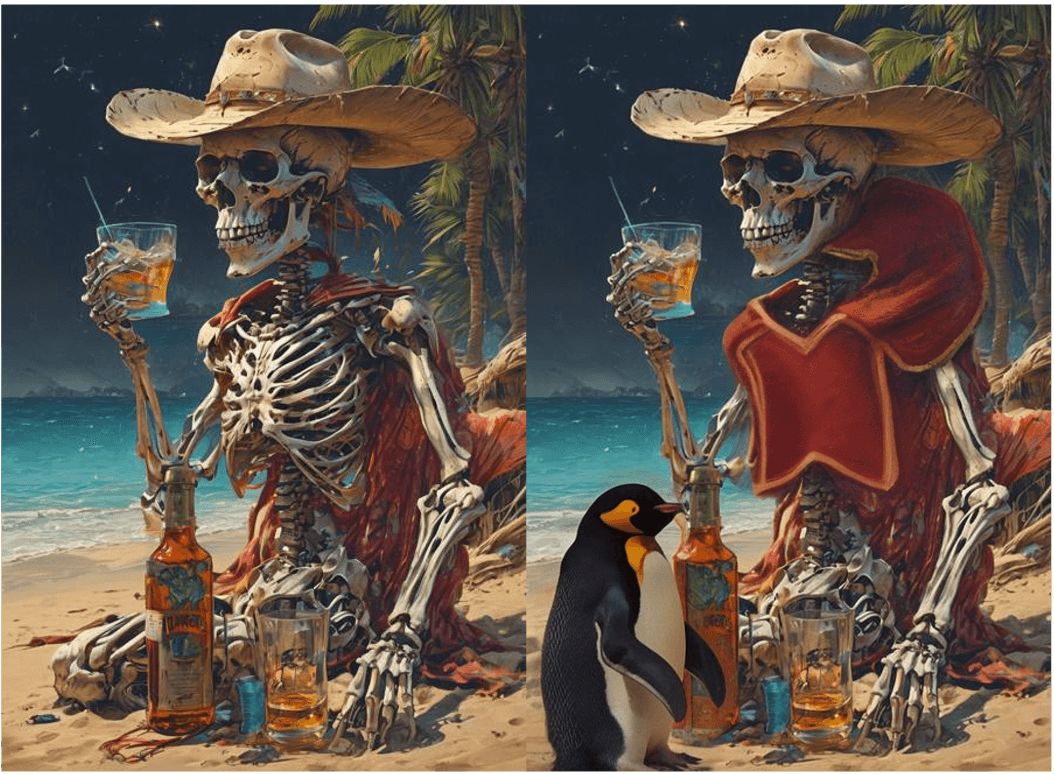}
        \end{subfigure}
    \end{minipage}

    \hspace{1cm}
    
    \begin{minipage}[b]{0.3\linewidth}
        \begin{subfigure}[b]{\linewidth}
            \centering
            \includegraphics[height=3.8cm]{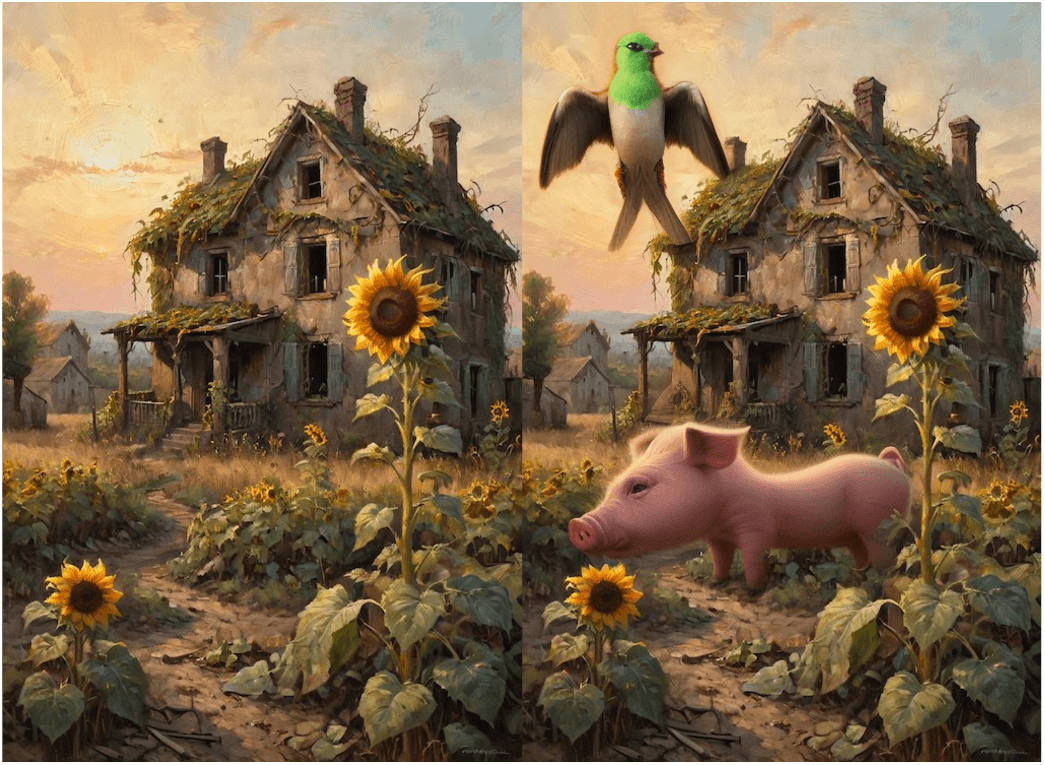}
        \end{subfigure}
    \end{minipage}
    \hspace{0.2cm}
    \begin{minipage}[b]{0.3\linewidth}
        \begin{subfigure}[b]{\linewidth}
            \centering
            \includegraphics[height=3.8cm]{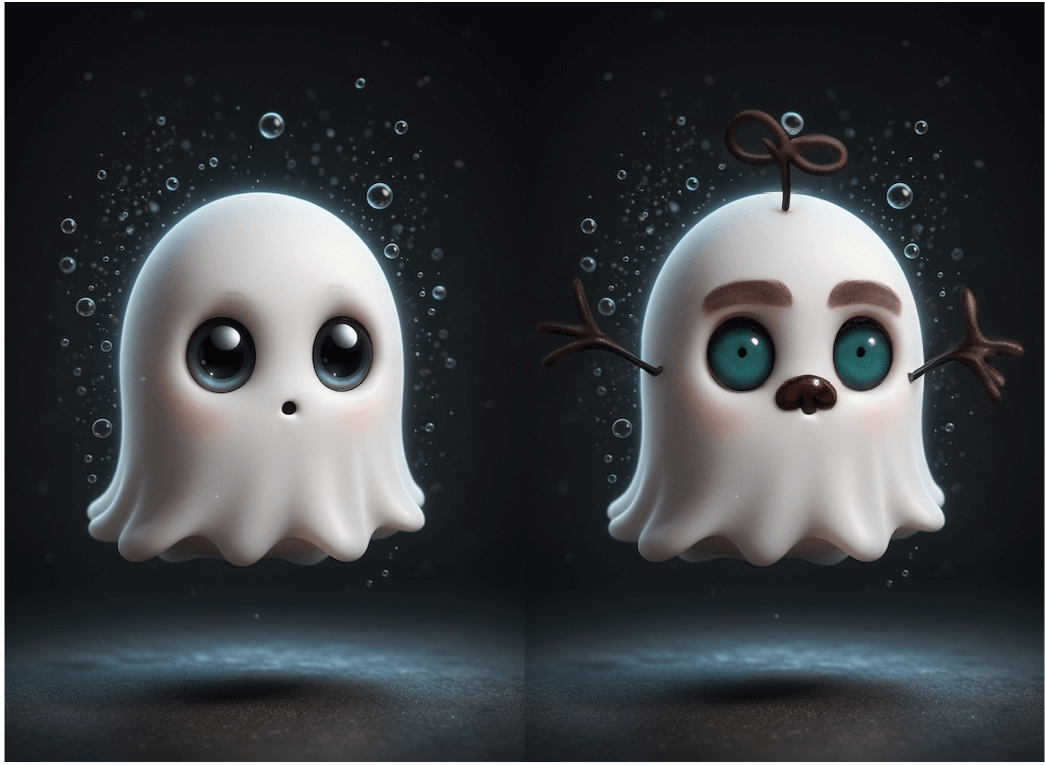}
        \end{subfigure}
    \end{minipage}
    \hspace{0.2cm}
    \begin{minipage}[b]{0.3\linewidth}
        \begin{subfigure}[b]{\linewidth}
            \centering
            \includegraphics[height=3.8cm]{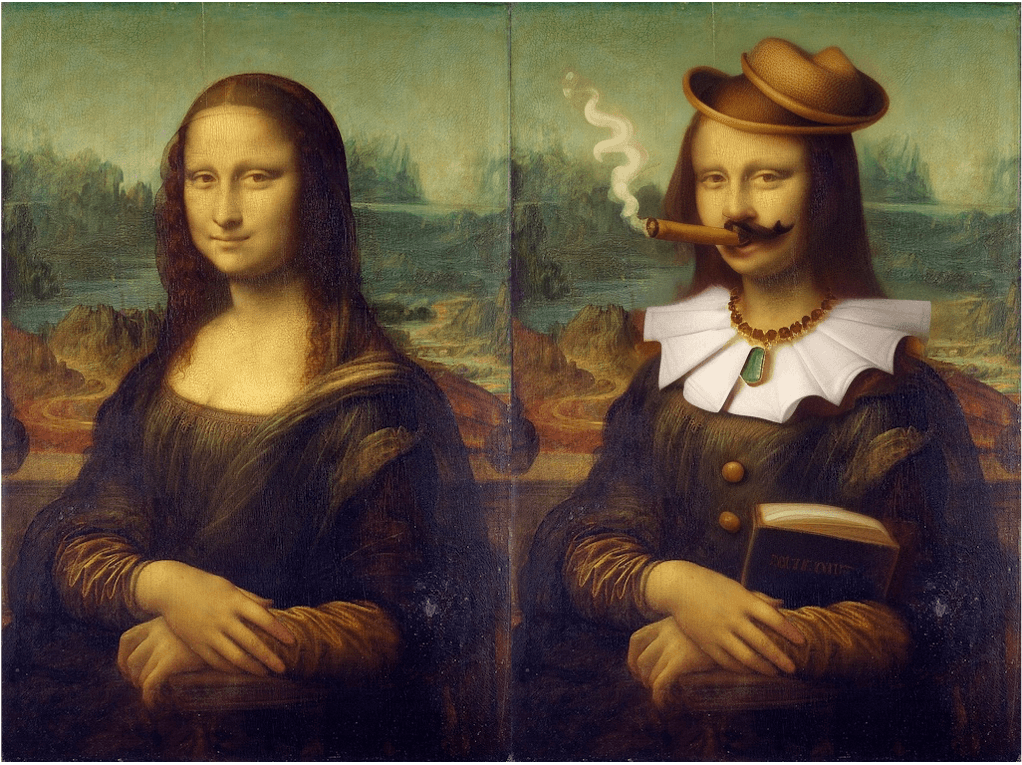}
        \end{subfigure}
    \end{minipage}

    \hspace{1cm}

    \begin{minipage}[b]{0.45\linewidth}
        \begin{subfigure}[b]{\linewidth}
            \centering
            \includegraphics[height=3.8cm]{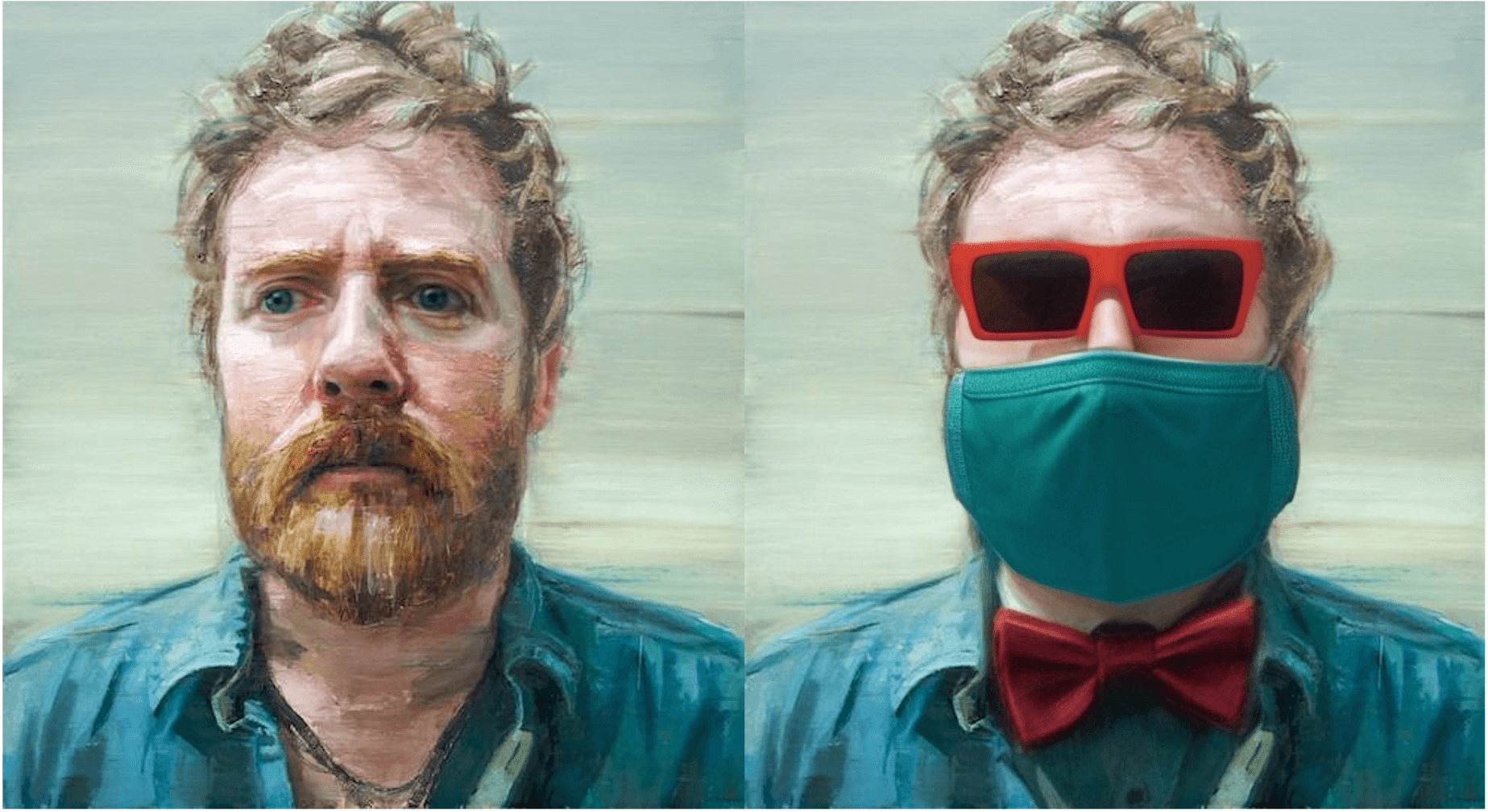}
        \end{subfigure}
    \end{minipage}
    \hspace{0.2cm}
    \begin{minipage}[b]{0.45\linewidth}
        \begin{subfigure}[b]{\linewidth}
            \centering
            \includegraphics[height=3.8cm]{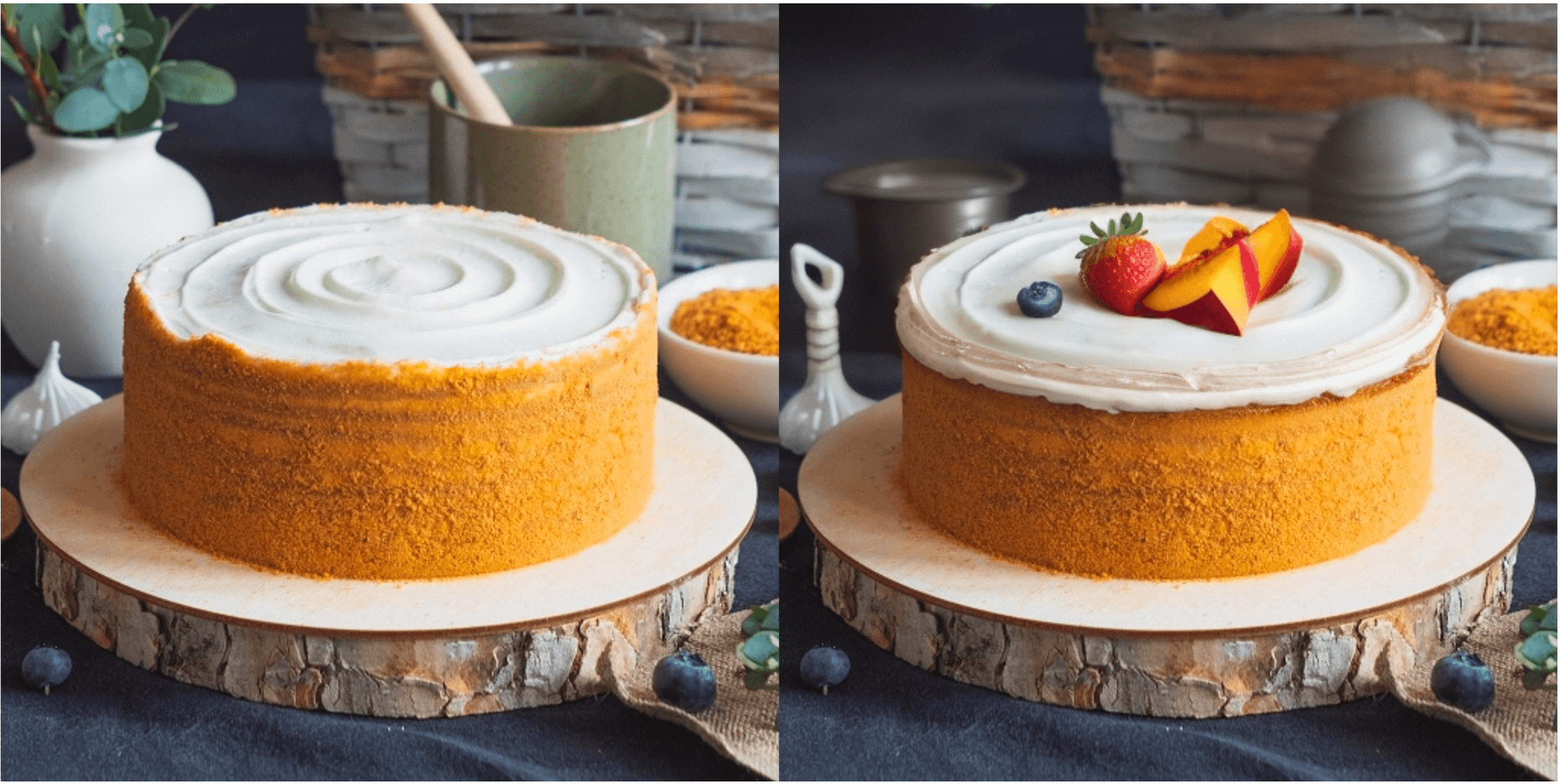}
        \end{subfigure}
    \end{minipage}
    \caption{A gallery of creative image editing achieved by the participants of the user study using \method. Each pair shows the original image and its edited version, demonstrating diverse user-driven modifications.}
    \vspace{-0.2cm}
    \label{fig:user_results}
\end{figure*}

The system's ability to maintain user engagement was evidenced by users voluntarily extending their editing sessions beyond the allocated time. After minimal training, users were able to create compelling edits, demonstrating the system's accessibility and ease of use. A gallery of user-edited images is presented in Fig.~\ref{fig:user_results}.

\end{document}